\documentclass[10pt,journal,compsoc]{IEEEtran}
\usepackage{algorithm}
\usepackage{booktabs}
\usepackage{multirow}
\usepackage{cite}

\usepackage{lipsum}
\usepackage{amsfonts}
\usepackage{graphicx}
\usepackage{epstopdf}
\usepackage{algorithmicx}
\usepackage{algpseudocode}
\usepackage{booktabs}
\usepackage{makecell}
\usepackage{empheq}
\usepackage{multirow}
\usepackage{multicol}
\usepackage{color}
\usepackage{colortbl}
\usepackage{arydshln}

\usepackage{wrapfig}
\usepackage{subcaption}
\usepackage{empheq}
\usepackage{enumitem}
\usepackage{url}
\usepackage{calc}

\definecolor{gray}{gray}{0.85}

\makeatletter
\renewenvironment{IEEEbiographynophoto}[1]{%
  \vspace{0.5\baselineskip}
  \par\noindent\textbf{#1}~%
  \small
}{\vspace{0.5\baselineskip}\par} 
\makeatother

\usepackage{amssymb}
\usepackage{amsthm}
\usepackage{physics}
\allowdisplaybreaks
\usepackage{hyperref}
\usepackage[capitalize,nameinlink]{cleveref}
\crefname{equation}{Eq.}{Eqs.}
\crefname{figure}{Fig.}{Figs.}
\crefname{table}{Tab.}{Tabs.}
\crefname{section}{Sec.}{Secs.}
\crefname{algorithm}{Alg.}{Algs.}
\crefname{theorem}{Thm.}{Thms.}
\crefname{lemma}{Lem.}{Lems.}
\crefname{corollary}{Col.}{Cols.}
\crefname{definition}{Def.}{Defs.}
\crefname{proposition}{Pro.}{Pros.}
\usepackage{aliascnt}
\newtheorem{theorem}{Theorem}[section]

\newaliascnt{lemma}{theorem}
\newtheorem{lemma}[lemma]{Lemma}
\aliascntresetthe{lemma}

\newaliascnt{corollary}{theorem}
\newtheorem{corollary}[corollary]{Corollary}
\aliascntresetthe{corollary}

\newaliascnt{proposition}{theorem}
\newtheorem{proposition}[proposition]{Proposition}
\aliascntresetthe{proposition}

\newaliascnt{definition}{theorem}
\newtheorem{definition}[definition]{Definition}
\aliascntresetthe{definition}

\newaliascnt{conjecture}{theorem}

\aliascntresetthe{conjecture}

\newaliascnt{problem}{theorem}

\aliascntresetthe{problem}

\newaliascnt{claim}{theorem}

\aliascntresetthe{claim}

\newaliascnt{remark}{theorem}

\aliascntresetthe{remark}

\newaliascnt{example}{theorem}

\aliascntresetthe{example}

\newaliascnt{fact}{theorem}
\newtheorem{fact}[fact]{Fact}
\aliascntresetthe{fact}

\crefname{conjecture}{Conjecture}{Conjectures}
\Crefname{conjecture}{Conjecture}{Conjectures}

\crefname{problem}{Problem}{Problems}
\Crefname{problem}{Problem}{Problems}

\crefname{claim}{Claim}{Claims}
\Crefname{claim}{Claim}{Claims}

\crefname{remark}{Remark}{Remarks}
\Crefname{remark}{Remark}{Remarks}

\crefname{example}{Example}{Examples}
\Crefname{example}{Example}{Examples}

\crefname{fact}{Fact}{Facts}
\Crefname{fact}{Fact}{Facts}

\newcommand{\mb}{\boldsymbol}

\newcommand{\mc}{\mathcal}

\newcommand{\bb}{\mathbb}
\newcommand{\set}[1]{\left\{ #1 \right\}}

\newcommand{\paren}{\pqty}
\newcommand{\brac}{\bqty}

\DeclareMathOperator*{\argmin}{arg\,min}

\begin{document}

\title{Exact Reformulation and Optimization for Direct Metric Optimization in Binary Imbalanced Classification}


\author{Le Peng~\thanks{Le Peng and  Yash Travadi contributed equally to this work.}, Yash Travadi, Chuan He, Ying Cui, and Ju Sun~\thanks{Le Peng, Ju Sun are with Computer Science and Engineering, University of Minnesota, MN, 55455, USA (email: peng0347@umn.edu; jusun@umn.edu).}
~\thanks{Chuan He was with the University of Minnesota when this work was performed. He is now with Department of Mathematics, Linköping University, Linköping, Sweden (email: chuan.he@liu.se).}
~\thanks{Yash Travadi is with School of Statistics, University of Minnesota, MN, 55455, USA (email: trava029@umn.edu).}
~\thanks{Ying Cui is with Industrial Engineering and Operations Research, University of California Berkeley, CA, 94704, USA (email: yingcui@berkeley.edu)}
} 

\markboth{Journal of \LaTeX\ Class Files,~Vol.~14, No.~8, August~2021}%
{Shell \MakeLowercase{\textit{et al.}}: A Sample Article Using IEEEtran.cls for IEEE Journals}

\IEEEpubid{0000--0000/00\$00.00~\copyright~2021 IEEE}

\maketitle

\begin{abstract}
For classification with imbalanced class frequencies, i.e., imbalanced classification (IC), standard accuracy is known to be misleading as a performance measure. While most existing methods for IC resort to optimizing balanced accuracy (i.e., the average of class-wise recalls), they fall short in scenarios where the significance of classes varies or certain metrics should reach prescribed levels. In this paper, we study two key classification metrics, precision and recall, under three practical binary IC settings: fix precision optimize recall (FPOR), fix recall optimize precision (FROP), and optimize $F_1$-score (OFOS). Unlike existing methods that rely on smooth approximations to deal with the indicator function involved, \textit{we introduce, for the first time, exact constrained reformulations for these direct metric optimization (DMO) problems}, which can be effectively solved by exact penalty methods. Experiment results on multiple benchmark datasets demonstrate the practical superiority of our approach over the state-of-the-art methods for the three DMO problems. We also expect our exact reformulation and optimization (ERO) framework to be applicable to a wide range of DMO problems for binary IC and beyond. Our code is available at \url{https://github.com/sun-umn/DMO}. 
\end{abstract} 




\section{Introduction}
\label{sec:intro}

Real-world classification problems often exhibit skewed class distributions, i.e., class imbalance, due to intrinsic uneven class frequencies and/or sampling biases. Examples abound in many domains, including disease diagnosis~\cite{codella2018skin,irvin2019chexpert,peng2022imbalanced}, insurance fraud detection~\cite{wei2013effective,herland2018big}, object detection~\cite{lin2017focal,chen2020ap}, image retrieval~\cite{irtaza2018ensemble,khatami2018parallel}, image segmentation~\cite{taghanaki2019combo,li2020analyzing,yeung2022unified}, and text classification~\cite{fernandez2018smote,wei2019eda}. Classification with class imbalance, or \textbf{imbalanced classification} (IC), has been an active research area in machine learning and related fields for decades~\cite{johnson2019survey,yang2022survey,peng2022imbalanced}. In this paper, \emph{we focus on binary IC}, as it covers many applied scenarios and faces several representative technical challenges common to general IC. 

For binary IC, standard performance metrics, such as standard accuracy and balanced accuracy (i.e., mean class-wise recall)~\cite{menon2013statistical,johnson2019survey,yang2022survey,peng2022imbalanced}, are often misaligned with practical goals. In particular, the recalls of the two classes are often not equally important. For example, in medical diagnosis, identifying positive patients is much more crucial than finding negatives; similarly, returning relevant images in image retrieval and detecting true frauds in fraud detection are clear priorities for each case. Thus, for these applications, maximizing the recall for the priority class is much more important than for the other, which can be achieved by a trivial classifier that classifies all inputs into the priority class. Hence, besides recall, it is also necessary to quantify the sharpness of the classifier on the priority class, often measured using the \textbf{precision} metric. 
\IEEEpubidadjcol

The precision-recall tradeoff could be controlled by optimizing the area under the precision-recall curve (AUPRC, or average precision---AUPRC's numerical approximation), which provides a holistic quantification of performance over the whole spectrum of precision-recall tradeoff. However, practical deployment of a binary classifier requires selecting a decision threshold that determines a single operating point on the curve. So, \textbf{in this paper, we focus on IC formulations that directly target the precision-recall tradeoff at single operating points}. Specifically, we consider fixing one metric while optimizing the other~\cite{eban2017scalable}, namely, fix precision optimize recall (\textbf{FPOR}) and fix recall optimize precision (\textbf{FROP}). For example, FROP can be used to maximize precision while ensuring a recall of at least $95\%$, particularly relevant for high-stakes applications such as healthcare and finance. Furthermore, we also consider optimizing the $F_1$ score (\textbf{OFOS}), a standard metric used for precision-recall tradeoff.


The key technical challenge in solving these direct metric optimization (DMO) problems is that all the metrics---precision, recall, and the $F_1$ score---involve \textbf{indicator functions}, which have a zero gradient almost everywhere, precluding gradient-based optimization methods. To address this challenge, most existing methods rely on smooth approximations (e.g., using sigmoid to replace the indicator function)~\cite{eban2017scalable,cotter2019two,benedict2021sigmoidf1}. Although standard for the classic empirical risk minimization (ERM) framework, smooth approximations can be problematic for these DMO problems due to a couple of reasons: (1) for constrained formulations, it is \textbf{critical to find feasible points}---while suboptimality in objective may be tolerated, infeasible points are unacceptable for most practical use cases. When using such approximations, it is challenging to ensure feasibility unless the approximation errors are sufficiently small; (2) since both objectives and constraints involved are \textbf{nonconvex and nonlinear}, using approximations can lead to \textbf{significantly suboptimal solutions}. In this paper, we address these issues with the following contributions. 
\begin{itemize}[leftmargin=*,nosep]
    \item We introduce a novel reformulation of indicator functions (\cref{sec:eq_reform,sec:reform_ineq}), \textbf{which is the first to handle indicator functions exactly}, as opposed to commonly used inexact approximation techniques. Induced reformulations of our three DMO problems \cref{def:FPOR,def:FROP,def:OFBS}, with nontrivial gradients almost everywhere, are amenable to gradient-based (constrained) optimization methods, leading to \textbf{the first computational framework to optimize exact binary IC metrics using gradient-based methods}. 

    \item Under mild conditions, we establish the equivalence of our reformulations with the original problems \cref{def:FPOR,def:FROP,def:OFBS} (\cref{thm:feasibility-retrict,thm:exact-fpor,thm:feasi-weak,thm:global-general}). In particular, we show that one can construct global solutions to the three DMO problems based on global solutions to the respective reformulations and vise versa. 
    
    \item We propose to solve the exact reformulations of \cref{def:FPOR,def:FROP,def:OFBS} using an exact penalty method, and benchmark it on real-world binary IC tasks covering image, text, and structured data. Our method consistently, often substantially, outperforms state-of-the-art (SOTA) methods for solving these binary IC problems.   
\end{itemize}
We have published a brief summary of the current work in~\cite{travadi2023direct}; the current version substantially extends that brief summary with a different optimization approach, theoretical analyses and proofs, comprehensive experiments, and all necessary details.

\section{Background \& related work}

\subsection{Direct metric optimization (DMO) for binary IC}
\label{sec:def_form}
\subsubsection{Three key formulations} 
Consider a binary IC task with a training dataset $\set{(\mb x_i,y_i)}_{i=1}^{N}$ independent and identically distributed (iid) sampled from a data distribution $\mc D_{\mc X \times \mc Y}$, where $\mc X\times\mc Y$ is the input-output data space and $\mc Y=\{0,1\}$. Let $\mc P$ and $\mc N$ denote the indices of positive ($y_i=1$) and negative ($y_i=0$) samples, respectively, and let $N_+\doteq|\mc P|$ and $N_-\doteq|\mc N|$. For any predictive model $f_{\mb \theta}:\mc X\rightarrow[0,1]$ parametrized by $\mb \theta$ and a decision threshold $t \in [0, 1]$, the final binary classifier is $\mb 1\set{f_{\mb \theta} > t}$, where $\mb 1\{\cdot\}$ is the standard indicator function. We are interested in three metrics
\begin{align}\label{def:pre-rec-Fb}
\textbf{Precision:} & \quad p(f_{\mb{\theta}},t) & \doteq & \quad
\frac{\sum\nolimits_{i\in\mc P}\mb{1}\{f_{\mb{\theta}}(\mb{x}_i)> t\}}
     {\sum\nolimits_{i\in\mc P\cup\mc N}\mb{1}\{f_{\mb{\theta}}(\mb{x}_i)> t\}}; \\
\textbf{Recall:} & \quad r(f_{\mb{\theta}},t) & \doteq & \quad
\frac{\sum\nolimits_{i\in\mc P}\mb{1}\{f_{\mb{\theta}}(\mb{x}_i)>t\}}{N_+};\\
\textbf{$F_1$-score:}& \quad F_{1}(f_{\mb{\theta}},t) & \doteq & \quad 
\frac{2\cdot p(f_{\mb{\theta}},t)\cdot r(f_{\mb{\theta}},t)}
     {p(f_{\mb{\theta}},t) + r(f_{\mb{\theta}},t)}; 
\end{align}
and \textbf{focus on three direct metric optimization (DMO) problems for binary IC} defined below. 

\begin{center}
  \fbox{%
    \begin{minipage}{0.95\linewidth}
\begin{subequations}
\noindent\textbf{Fix precision optimize recall (FPOR):}
\begin{equation}
\begin{alignedat}{2}
&\max\nolimits_{\mb\theta,\, t} &&\; r(f_{\mb\theta}, t) \quad \text{s.t.} \quad p(f_{\mb\theta}, t) \ge \alpha
\end{alignedat}
\label{def:FPOR}
\end{equation}

\noindent\textbf{Fix recall optimize precision (FROP):}
\begin{equation}
\begin{alignedat}{2}
&\max\nolimits_{\mb\theta,\, t} &&\; p(f_{\mb\theta}, t) \quad \text{s.t.} \quad r(f_{\mb\theta}, t) \ge \alpha
\end{alignedat}
\label{def:FROP}
\end{equation}

\noindent\textbf{Optimize $F_1$-score (OFOS):}
\begin{equation}
\begin{alignedat}{2}
&\max\nolimits_{\mb\theta,\, t} &&\; F_{1}(f_{\mb\theta}, t)
\end{alignedat}
\label{def:OFBS}
\end{equation}
\end{subequations}
\end{minipage}%
  }
\end{center}
where $\alpha\in[0,1]$ is a user-specified target precision/recall level. These DMO problems are not new: they have been briefly studied in machine learning and information retrieval (e.g., object detection, image retrieval, recommendation systems), where the FPOR/FROP problems are especially rarely touched on~\cite{joachims2005support,nan2012optimizing,puthiya2014optimizing,lipton2014thresholding,eban2017scalable,lee2021surrogate,benedict2021sigmoidf1,werner2022review}. In contrast to AUPRC maximization~\cite{yang2022auc,cakir2019deep,brown2020smooth,qi2021stochastic,wen2022exploring} that optimizes holistic performance over all possible decision thresholds, \cref{def:FPOR,def:FROP,def:OFBS} aims at a single operating point on the precision-recall curve; particularly, the former two put explicit controls on their own prioritized metrics. In computer vision, DMO for other ranking metrics, such as normalized discounted cumulative gain (NDCG) and precision/recall at top-$k$ positions, have also been gaining traction~\cite{kar2015surrogate,patel2022recall,engilberge2019sodeep,yang2022algorithmic}. \textbf{In this paper, we study the three DMO problems in the context of binary IC, but we believe the proposed ideas can be extended to other DMO problems}.

\subsubsection{Optimization challenges} 
Two challenges stand in solving \cref{def:FPOR,def:FROP,def:OFBS}. \textbf{Challenge 1}: The indicator function of the form $\mathbf{1}\{a > 0\}$ is discontinuous at $a = 0$ and has a zero gradient everywhere else. This implies that the objectives and constraint functions involved in \cref{def:FPOR,def:FROP,def:OFBS} have a zero gradient almost everywhere, absent the discontinuous points. So, gradient-based methods are out of the question; \textbf{Challenge 2}: The constraints in \cref{def:FPOR,def:FROP} are often \emph{nonconvex and nonlinear}. Designing numerical methods that can find feasible points for these problems can be a challenging task. However, not finding feasible points defeats the purpose of explicit metric control in the constraints.  

\subsubsection{SOTA methods for addressing the challenges} 
There are mainly three lines of ideas \textbf{to address Challenge 1}:  {\textbf{(A)}} Early work considers structural support vector machines for DMO and effectively optimizes an upper bound of the metric of interest~\cite{joachims2005support,yue2007support,tsochantaridis2005large}. This restricts the choice of classifiers and also induces exponentially many constraints (dealt with by cutting-plane methods) and combinatorial optimization problems (often solvable with quadratic complexity in the training size) per iteration; see also a recent development~\cite{fathony2020ap} that breaks the classifier restriction; \textbf{(B)} Most modern work is based on smooth approximations to the indicator or metric functions~\cite{qin2010general,chapelle2010gradient,benedict2021sigmoidf1,patel2022recall,kar2015surrogate,engilberge2019sodeep,lee2021surrogate,eban2017scalable,rath2022optimizing,sanyal2018optimizing,kumar2021implicit,yang2022algorithmic,cotter2019two,narasimhan2019optimizing,cotter2019training}, so that gradient-based optimization methods can be naturally applied. Although these papers use different forms of approximation in disparate contexts, it is clear that all draw inspiration from surrogate losses commonly used in machine learning, e.g., using the sigmoid function $z \mapsto 1/(1+e^{-z})$ to approximate the indicator function $z \mapsto \mb 1\set{z > 0}$. However, when applying such approximations in solving \cref{def:FPOR,def:FROP,def:OFBS}, there are critical catches including numerical discrepancies and vanishing gradients; see \cref{subsec:smooth}; \textbf{(C)} Moreover, black-box approaches~\cite{rolinek2020optimizing,poganvcic2019differentiation,huang2021metricopt} construct or learn approximations to the metric function or its ``gradient'' based on black-box evaluations of function values. Although these methods are general, they also suffer from numerical discrepancies and vanishing gradients, similar to methods in (B); see \cref{subsec:smooth}. 

\textbf{To tackle Challenge 2}, optimization methods capable of reliably handling nonlinear constraints are needed. \emph{Penalization methods}, including penalty methods, Lagrangian methods, and augmented Lagrangian methods (ALMs), have been popularly used for this purpose~\cite{NoceWrig06}. For example, the TensorFlow-based library TFCO \cite{cotter2019two} has implemented Lagrangian methods, while Python-based GENO \cite{laue2019geno,laue2022optimization} and C++-based Ensmallen \cite{curtin2021ensmallen} have implemented ALMs. Besides penalization methods, \textit{interior-point methods} (IPMs) and \textit{sequential quadratic programming} (SQP) methods are also widely adopted for constrained optimization~\cite{NoceWrig06}, implemented in solvers such as Knitro \cite{byrd2006k}, Ipopt \cite{wachter2006implementation}, and the recent PyGranso~\cite{liang2022ncvx}. In this paper, we develop a unified algorithmic framework for handling the three DMO problems based on \textit{exact penalty methods}; see \cref{subsec:opt}. 

\subsection{Critical issues of approximation-based methods}\label{subsec:smooth}

\begin{wrapfigure}{r}{0.25\textwidth}
    \centering
    \vspace{-0.8em}
    \includegraphics[width=\linewidth]{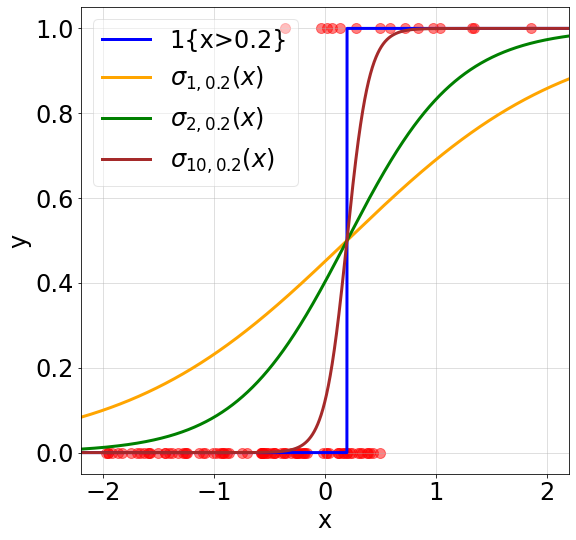}
    \vspace{-1em}
    \caption{Data distribution, classifier, and surrogates for $t=0.2,\ T\in\{1,2,10\}$.}
    \label{fig:data_ind_surrogate}
    \vspace{-1em}
\end{wrapfigure}
Consider a 1D imbalanced dataset and classifier: $\bb P(y=1) = 0.2$, $\bb P(y=0) = 0.8$, $\bb P(x| y=1) \sim \mathrm{Uniform}[-0.5,2]$, $\bb P(x | y=0) \sim \mathrm{Uniform}[-2,0.5]$; $500$ iid points drawn; single-threshold classifier $f_t(x) = \mb{1}\{x>t\}$. Now, suppose that we approximate the indicator function in $f_t(x)$ by a sigmoid with the temperature parameter $T$, i.e., $\sigma_{T,t}(x) = 1/(1+e^{-T(x-t)})$. Note that the larger the $T$, the tighter the approximation. \cref{fig:data_ind_surrogate} visualizes the data, $f_t(x)$, and $\sigma_{T,t}(x)$ with different values of $T$. Next, we highlight a couple of critical issues that approximation-based methods can face when solving \cref{def:FPOR,def:FROP,def:OFBS}. 

\begin{figure*}[!htbp]
    \centering
    \subfloat[Approx. precision\label{fig:ill_sub1}]{\includegraphics[width=0.22\linewidth]{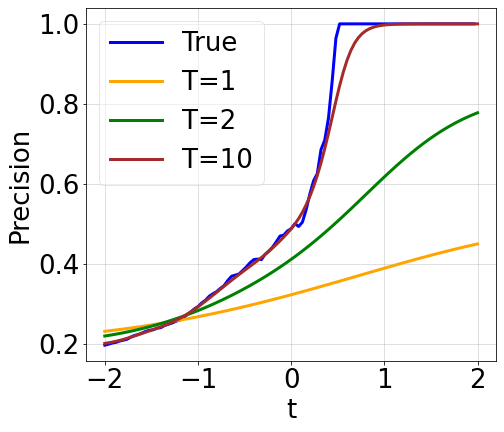}}
    \subfloat[Approx. recall\label{fig:ill_sub2}]{\includegraphics[width=0.22\linewidth]{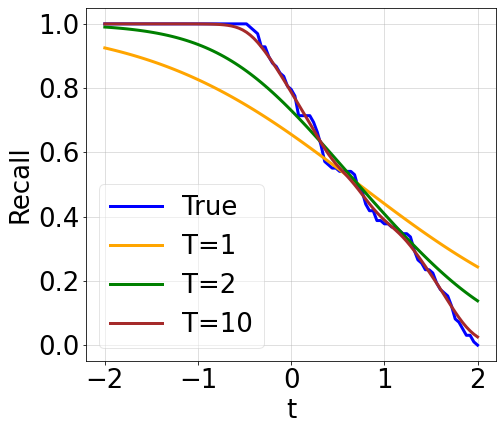}}
    \subfloat[Approx. $F_1$\label{fig:ill_sub3}]{\includegraphics[width=0.22\linewidth]{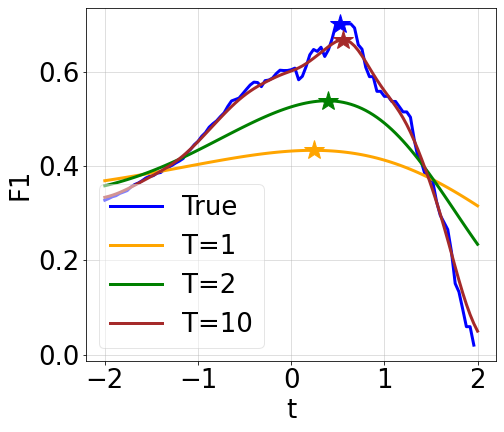}}
    \subfloat[Derivative of $\sigma(x)$\label{fig:ill_sub4}]{\includegraphics[width=0.22\linewidth]{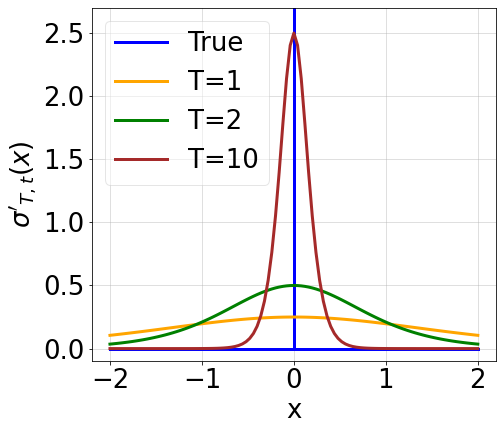}}
    \caption{Illustration of issues with using smooth approximations/surrogates when solving \cref{def:FPOR,def:FROP,def:OFBS}. For our toy 1D dataset and the same binary IC, (a), (b), and (c) show the true and approximate precision/recall/$F_1$-score vs. threshold ($t$), with different temperature parameters ($T$'s), respectively. The $\star$'s in (c) locate the optimal values of the approximated $F_1$'s with different $T$'s. (d) shows the derivative of the parameterized sigmoid function.}
    \label{fig:ill}
    \vspace{-1em}
\end{figure*}

\subsubsection{Issue I: Numerical discrepancies} 
For the same dataset, predictive model, and decision threshold, the approximate value of the precision/recall/$F_1$-score can be very different from the true value; see \cref{fig:ill_sub1,fig:ill_sub2,fig:ill_sub3}. This is problematic when we try to control these metrics in constraints, e.g., as in FPOR and FROP: \textbf{the constraint set may be empty, or the numerical control may become looser or tighter than required}. For example, if we set $\mathrm{precision}\ge 0.9$ in FPOR and take $T=2$ approximation in \cref{fig:ill_sub1}, there is no feasible $t$ as the best achievable precision is less than $0.8$, except for trivial predictive models. Even if we aim for $\mathrm{precision}\ge 0.6$ so that $T=2$ makes the constraint set nonempty, the ranges of feasible $t$ between the true and approximate versions are still vastly different---any feasible $t$ for the approximate version leads to a true precision much higher than the target $0.6$. Moreover, \textbf{for OFOS, the numerical discrepancy can lead to very suboptimal predictive models and decision thresholds}. A simple example is in \cref{fig:ill_sub3}, where $T=1$ or $T=2$ can lead to decision thresholds that are significantly suboptimal in terms of true $F_1$.

\subsubsection{Issue II: Vanishing gradients}
One may wonder why not tighten up the approximation. For example, the numerical gaps we discuss above can be suppressed by setting a larger $T$. Although this is true, the vanishing-gradient issue due to the indicator function resurfaces once we make the approximation reasonably tight, as shown in \cref{fig:ill_sub4}: when $T = 10$, over a large region of $t$, $\sigma'_{T,t}(x)$ is negligibly small, resembling the zero derivative of the indicator function itself. In other words, \textbf{there is a tricky tradeoff between the quality and the numerical well-behavedness of the approximation}. An alternative strategy is to adopt a continuation idea: gradually increase the temperature $T$ during training to smoothly transition from coarse to sharp approximations~\cite{chapelle2010gradient}. However, these methods require careful, and likely problem-specific tuning of the temperature schedule, tricky for general-purpose use. 

\subsection{Other related binary IC problems}
For binary IC, besides the precision-recall tradeoff considered here, another popular direction is the tradeoff between the true positive rate (TPR, i.e., recall) and the false positive rate (FPR). This TPR-FPR tradeoff is usually measured using the receiver operating characteristic curve (ROCC), and its summarizing metric, area under the ROCC (AUROCC). Optimizing the AUROCC has been studied extensively in the literature~\cite{majnik2013roc,yang2022auc}. Another popular formulation targeting these metrics is the Neyman-Pearson classification problem, which aims to maximize TPR (i.e., $1-$type II error) while fixing FPR (i.e, type I error)~\cite{tong2016survey}. However, as argued in \cref{sec:intro}, we focus on the precise-recall tradeoff, which is more informative when there is considerable data imbalance with a priority class~\cite{saito2015precision,williams2021effect}. 

\section{Our methods}\label{sec:methods}
Consider the problem setup in \cref{sec:def_form}, and further assume that the positive class is prioritized so that precision and recall are calculated with respect to it. In this paper, we provide reformulations, computational algorithms, and theoretical guarantees for all three DMO problems (FPOR, FROP, OFOS) defined in \cref{def:FPOR,def:FROP,def:OFBS}. However, \textbf{for clarity, below we focus on FPOR to illustrate the main ideas and results; the complete results for FROP and OFOS can be found in \cref{apx:proofs}}. The FPOR problem is: 
\begin{equation}
\begin{aligned} \label{def:FPOR_expanded}
\max\nolimits_{\mb\theta,\, t \in [0,1]}\ & 
\tfrac{1}{N_+}\sum\nolimits_{i\in\mc P} \mb{1}\{f_{\mb{\theta}}(\mb{x}_i)>t\}\\
\text{s.t.} \quad & 
\frac{\sum_{i\in\mc P} \mb{1}\{f_{\mb{\theta}}(\mb{x}_i)>t\}}
     {\sum_{i\in\mc P\cup\mc N} \mb{1}\{f_{\mb{\theta}}(\mb{x}_i)>t\}} \ge \alpha
\end{aligned}, 
\end{equation}
where $\alpha \in [0,1]$ is the user-specified target precision level. In the corner case $\mb{1}\{f_{\mb{\theta}}(\mb{x}_i)>t\} = 0\; \forall i\in P\cup\mc N$, i.e., no samples are predicted to be positive, we define the precision as $1$ and hence the precision constraint is satisfied trivially. 

\subsection{Equality-constrained reformulation of FPOR}
\label{sec:eq_reform}

The formulation in \cref{def:FPOR_expanded} is not suitable for gradient-based (constrained) optimization methods, as the indicator function has a zero gradient almost everywhere. To combat the challenge, we introduce a continuous lifted reformulation to \cref{def:FPOR_expanded}. The first step is to introduce an auxiliary optimization variables $\mb s =  [s_1,\cdots,s_N]^\top\in [0, 1]^N$ so that 
\begin{align*}
    s_i = \mb{1}\{f_{\mb{\theta}}(\mb{x}_i)>t \} \; \forall \; i \Longleftrightarrow s_i - \mb{1}\{f_{\mb{\theta}}(\mb{x}_i)>t \} = 0 \; \forall \; i,  
\end{align*}
leading to the lifted reformulation of \cref{def:FPOR_expanded}: 
\begin{equation}\label{def:FPOR_eq_lifted}
\begin{aligned}
    & \max\nolimits_{\mb\theta,\mb s \in [0, 1]^N,t \in [0, 1]} \quad \tfrac{1}{N_+}\sum\nolimits_{i\in\mc P}s_i\\
    &\text{s.t.}  \quad \frac{\sum_{i\in\mc P}s_i}{\sum_{i\in\mc P\cup\mc N}s_i} \ge \alpha,\quad s_i - \mb{1}\{f_{\mb{\theta}}(\mb{x}_i)>t\}=0\ \forall i. 
\end{aligned}
\end{equation}
The equivalence of \cref{def:FPOR_expanded} and \cref{def:FPOR_eq_lifted}, in the sense that one can construct a global solution of one from that of the other, is immediate. But this does not make much progress, as indicator functions still appear in the constraints. 

The next step, which is crucial to our reformulation, is to capitalize on the following equivalence---\emph{a main novelty of our paper}: 
\begin{lemma}\label{lemma:indic-equiv}
For any fixed $t \in \bb R$, the following equivalence holds for all $a \ne t$: 
    \begin{equation}
s - \mb{1}\{a>t\} \;\Longleftrightarrow\;
s + [s+a-1-t]_+ - [s+a-t]_+ = 0
\end{equation}
    where $[\cdot]_+\doteq\max(\cdot,0)$. Moreover, $s \in \set{0, 1} \subset [0, 1]$ when either of the two sides holds. 
\end{lemma}
This can be easily verified algebraically; see \cref{sec:proof-indic-equiv}. However, a pictorial interpretation makes it more intuitive. For any fixed $t$, define two functions $G_t(a,s)$ and $H_t(a,s)$ of $\bb R^2 \to \bb R$ as follows: 
\begin{equation}
\begin{aligned}\label{def:g-h}
    G_t(a,s) & \doteq s - \mb{1}\{a>t\}, \\
     H_t(a,s) & \doteq s +  [s+a-1-t]_+ - [s+a-t]_+.
\end{aligned} 
\end{equation}
Note that while $G_t$ is discontinuous at $a = t$, $H_t$ is piecewise linear and continuous everywhere. Recall that for any function $f: \bb R^n \to \bb R$, the $\gamma$-level set of $f$ is defined as 
\begin{align}
    L_{\gamma}(f) \doteq \set{\mb x \in \bb R^n: f(\mb x) = \gamma}. 
\end{align}
\begin{figure}[!htbp] 
      \centering
    \subfloat{%
      \includegraphics[width=0.5\linewidth]{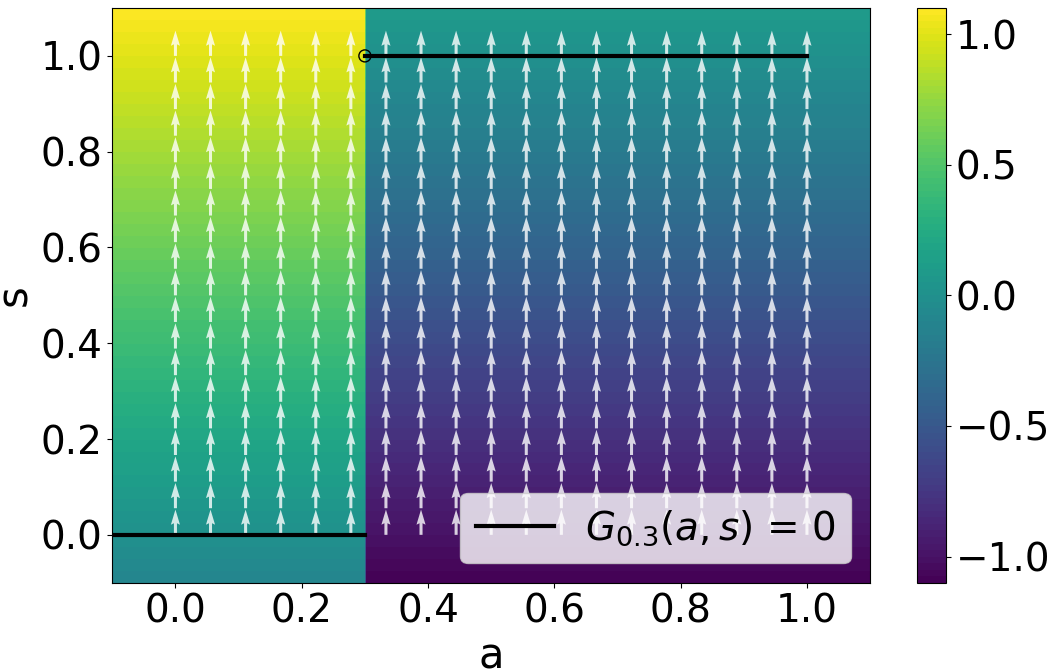}
      \label{fig:indicator}
    }
    \subfloat{%
      \includegraphics[width=0.5\linewidth]{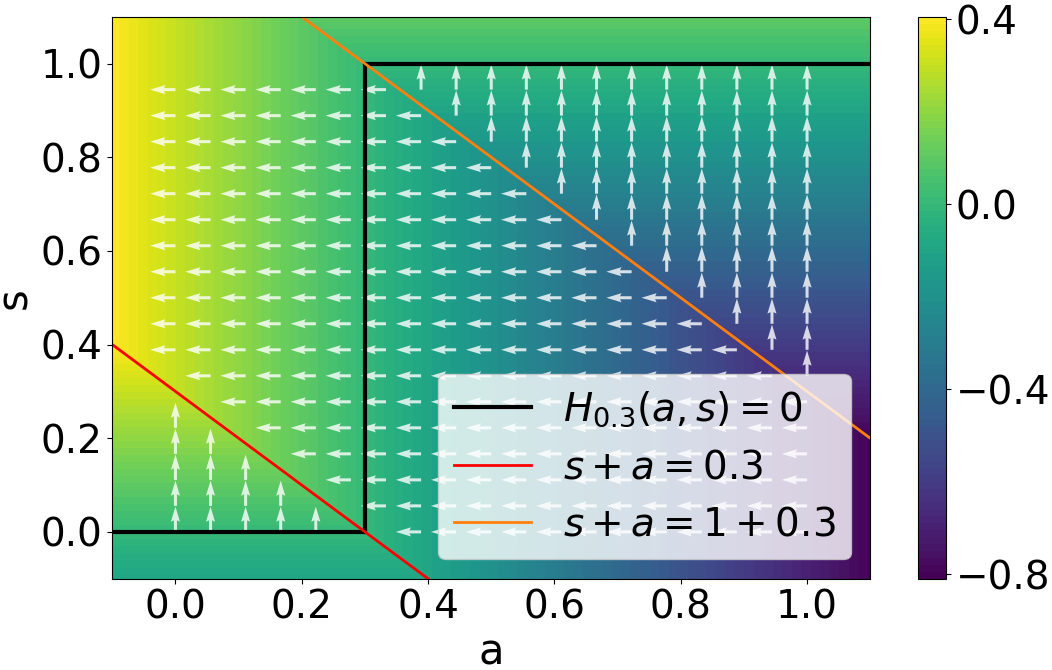}
      \label{fig:cont_appro}
    }
    \caption{Heatmap visualization of $G_t$ and $H_t$ for $t = 0.3$, their $0$-level sets $L_0(G_t)$ and $L_0(f_t)$, as well as their gradient fields. Note that for our purposes, $t \in [0, 1]$ and $a \in [0, 1], s \in [0, 1]$.  }
    \label{fig:contour}
    \vspace{-1em}
\end{figure}
Clearly, 
\begin{align*}
    L_{0}(G_t) & = \set{(a, s): s - \mb{1}\{a>t\}  =  0 }, \\
    L_{0}(H_t) & = \set{(a, s): s + [s+a-1-t]_+ - [s+a-t]_+ = 0 }. 
\end{align*}
The following result can be observed directly from \cref{fig:contour}, and is equivalent to \cref{lemma:indic-equiv}: 
\begin{corollary} \label{obs:coincide}
For any fixed $t \in \bb R$, $L_{0}(G_t) \cap \set{(a, s): a \ne t} = L_{0}(H_t) \cap \set{(a, s): a \ne t}$. 
\end{corollary}
\cref{obs:coincide} suggests that we can replace the $s_i - \mb{1}\{f_{\mb{\theta}}(\mb{x}_i)>t\}=0\ \forall i$ constraints in \cref{def:FPOR_eq_lifted} by $s_i + [s_i+f_{\mb{\theta}}(\mb{x}_i)-t-1]_+ - [s_i+f_{\mb{\theta}}(\mb{x}_i)-t]_+ = 0\ \forall i$, if we can guarantee that $f_{\mb{\theta}}(\mb{x}_i)-t \ne 0 \ \forall i$, leading to 
\begin{center}
  \fbox{%
    \begin{minipage}{0.98\linewidth}
\begin{align}\label{def:FPOR_eq_continuous}
&\max\nolimits_{\mb\theta,\, \mb s \in [0,1]^N,\, t \in [0,1]} \quad 
    \tfrac{1}{N_+}\sum\nolimits_{i\in\mc P}s_i \nonumber \\ 
& \text{s.t.} \quad 
    {\sum\nolimits_{i\in\mc P}s_i}/{\sum\nolimits_{i\in\mc P\cup\mc N}s_i} \ge \alpha, \\[4pt]
&  s_i + [s_i + f_{\mb{\theta}}(\mb{x}_i) - t - 1]_+
       - [s_i + f_{\mb{\theta}}(\mb{x}_i) - t]_+ = 0,\ \forall i.\nonumber
\end{align}
\end{minipage}%
  }
\end{center}
The $f_{\mb{\theta}}(\mb{x}_i)-t \ne 0 \ \forall i$ condition suggests the following definition to rule out pathological points. 
\begin{definition}[Non-singular $(\mb \theta, t)$] 
    A pair $(\mb \theta, t)$ is said to be non-singular over the training set $\set{(\mb x_i, y_i)}_{i=1}^N$ if $f_{\mb \theta}(\mb x_i) \ne t\ \forall i$. 
\end{definition}
Due to the equivalence of \cref{def:FPOR_expanded} and \cref{def:FPOR_eq_lifted}, as well as \cref{obs:coincide} (i.e., \cref{lemma:indic-equiv}), we have the following equivalence. 
\begin{proposition}
     A point $(\mb \theta^\ast, \mb s^\ast, t^\ast)$ with non-singular $(\mb \theta^\ast, t^\ast)$ is a global solution for \cref{def:FPOR_eq_continuous} if and only if it is a global solution for \cref{def:FPOR_eq_lifted}. Moreover, if $(\mb \theta^\ast, \mb s^\ast, t^\ast)$ is a global solution for \cref{def:FPOR_eq_continuous}, $(\mb \theta^\ast, t^\ast)$ is a global solution for \cref{def:FPOR_expanded}; if $(\mb \theta^\ast, t^\ast)$ is a global solution for \cref{def:FPOR_expanded}, $(\mb \theta^\ast, [\mb 1\set{f_{\mb \theta^\ast}(\mb x_i) > t^\ast}]_i, t^\ast)$ is a global solution for \cref{def:FPOR_eq_continuous}, where 
    \begin{align*}
        &[\mb 1\set{f_{\mb \theta^\ast}(\mb x_i) > t^\ast}]_i \doteq \\ 
        & \qquad [\mb 1\set{f_{\mb \theta^\ast}(\mb x_1) > t^\ast}; \dots; \mb 1\set{f_{\mb \theta^\ast}(\mb x_N) > t^\ast}] \in \bb R^{N}. 
    \end{align*} 
\end{proposition} 
To ensure that $f_{\mb \theta}(\mb x_i) \ne t\ \forall i$ in actual computation, we will describe how a simple barrier-style regularization suffices in \cref{sec:regularization}. 

Now, the central question is why reformulation $\cref{def:FPOR_eq_continuous}$ is beneficial. The answer lies in the difference between the (sub)gradient fields of $G_t$ and $H_t$ for $a \ne t$, in $[0, 1] \times [0, 1]$---as $s \in [0, 1]$ and $a = f_{\mb \theta}(\mb x) \in [0, 1]$ in our context: while $ \partial G_t(a, s) = \set{[\begin{smallmatrix} 0\\ 1 \end{smallmatrix}]}$, hence gradient-based optimization methods will make no progress in optimizing $\mb \theta$, 
\begin{align}
    & \partial H_t(a, s) = \nonumber \\
    & \quad \begin{cases} 
    \set{[\begin{smallmatrix} -1\\ 0 \end{smallmatrix}]}   &  t< s+a < 1+t \\
    \set{[\begin{smallmatrix} 0\\ 1 \end{smallmatrix}]}    &  s+a < t \ \text{or} \ s+a > 1+t \\
    \set{[\begin{smallmatrix} \omega - 1\\ \omega \end{smallmatrix}]: \omega \in [0, 1]}  & s+a = t \ \text{or} \ s+a = 1+t
    \end{cases}
\end{align}
where $\partial(\cdot)$ denotes the Clarke subdifferential and we note that piecewise linear functions such as $H_t$ are locally Lipschitz and hence Clarke subdifferentiable~\cite{clarke1990optimization,clason2020introduction}; see \cref{fig:contour} for visualization of the (sub)gradient fields. We observe that: (i) While $\partial_a G_t$ is always $0$ over $[0, 1]^2$, $\partial_a H_t$ is non-zero over $\set{(a ,s): t < s+a < 1+t}$, which takes at least $1- t^2/2 - (1-t)^2/2 \ge 1/2$ measure of $[0, 1]^2$; and (ii) Although $\partial_a H_t$ is zero over $\set{(a ,s): s+a< t \ \text{or} \ s+a > 1+t}$, we have that
\begin{equation}
\begin{aligned}
    s+a & < t \Longrightarrow s < t, a < t \qquad \quad \text{and} \\   
    s+a & > 1+t \Longrightarrow s > t, a > t. 
\end{aligned}
\end{equation}
Since we can gauge the value of $\mb 1\set{a >t}$ from both $a$ and $s$, when $s < t, a < t$ or $s > t, a > t$ we have good confidence in the value of $\mb 1\set{a >t}$---treating $s$ as a ``confidence score''. So, in these cases, $\partial_a G_t = 0$ is fine. In comparison, over $\set{(a ,s): t < s+a < 1+t}$ where $\mb 1\set{a >t}$ is highly uncertain, it is crucial to have non-zero $\partial_a G_t$. 

We note that besides the constraints $s_i + [s_i+f_{\mb{\theta}}(\mb{x}_i)-t-1]_+ - [s_i+f_{\mb{\theta}}(\mb{x}_i)-t]_+ = 0\ \forall i$, the objective and the other constraint in \cref{def:FPOR_eq_continuous} also induce a non-zero gradient for $s$. Moreover, the regularization term described in \cref{sec:regularization} also induces a non-zero gradient for $a$. 

\subsection{Inequality-constrained reformulation of FPOR}\label{sec:reform_ineq}
Our equality-constrained reformulation \cref{def:FPOR_eq_continuous} is grounded on 
\cref{lemma:indic-equiv}, which implies that $\mb s \in \set{0, 1}^N$ for the feasible set of \cref{def:FPOR_eq_continuous}. The empty interior of $\mb s$ can cause computational challenges in practice. In this section, we show that these equality constraints can be relaxed to inequality ones, significantly expanding the feasible set \emph{without affecting the exactness of our reformulation}. The said relaxation hinges on the following technical lemma, which complements \cref{lemma:indic-equiv}; the proof can be found in \cref{sec:proof-relax-key}. 
\begin{lemma} \label{lemma:relax-key}
    For any fixed $t \in \bb R$, the following hold for all $a \ne t$ and all $s \in [0, 1]$: 
    \begin{align*}
        s + [s+a-1-t]_+ - [s+a-t]_+ \le 0 & \Longleftrightarrow s \le \mb 1\set{a > t}, \\
         s + [s+a-1-t]_+ - [s+a-t]_+ \ge 0 & \Longleftrightarrow s \ge \mb 1\set{a > t}. 
    \end{align*}
\end{lemma}
Similar to \cref{lemma:indic-equiv}, there is also a geometric interpretation of \cref{lemma:relax-key}. Recall that for any function $f: \bb R^n \to \bb R$, the $\gamma$-sublevel set and $\gamma$-superlevel set are defined as 
\begin{align}
    L_{\gamma}^-(f) & \doteq \set{\mb x \in \bb R^n: f(\mb x) \le \gamma},  
    \quad \text{and} \nonumber \\ 
    L_{\gamma}^+(f) & \doteq \set{\mb x \in \bb R^n: f(\mb x) \ge \gamma},  
\end{align}
respectively. \cref{lemma:relax-key} states the following equivalence regarding super- and sublevel sets, which complements the geometric result in \cref{obs:coincide} and is clear from \cref{fig:contour}: 
\begin{corollary}  \label{lemma:relax-levelset}
    For any fixed $t \in \bb R$ and $G_t, H_t: \bb R^2 \to \bb R$ as defined in \cref{def:g-h}, we have 
    \begin{align}
        & L_{0}^-(G_t) \cap \set{(a, s): a \ne t, s \in [0, 1]} 
        = \nonumber \\
        & \qquad L_{0}^-(H_t) \cap \set{(a, s): a \ne t, s \in [0, 1]},  \\
        & L_{0}^+(G_t) \cap \set{(a, s): a \ne t,  s \in [0, 1]} 
        = \nonumber \\
        & \qquad L_{0}^+(H_t) \cap \set{(a, s): a \ne t,  s \in [0, 1]}. 
    \end{align}
\end{corollary} 
Now another relaxation of \cref{def:FPOR_eq_continuous}---the final formulation on which we perform the actual computation: 
{\small 
\begin{center}
  \fbox{%
    \begin{minipage}{0.98\linewidth}
\begin{align}\label{eq:FPOR-almost}
    & \max\nolimits_{\mb\theta, \mb{s}\in [0, 1]^N, t \in [0, 1]} \quad  \tfrac{1}{N_+}\sum\nolimits_{i\in\mc P}s_i \quad \nonumber \\
    & \text{s.t.} \quad  {\sum\nolimits_{i\in\mc P}s_i}/{\sum\nolimits_{i\in\mc P\cup\mc N}s_i} \ge \alpha,  \\
    & s_i + [s_i+f_{\mb \theta}(\mb x_i) - t-1]_+ - [s_i+f_{\mb \theta}(\mb x_i) - t]_+\le0\;\forall i\in\mc P,\nonumber\\
    & s_i + [s_i+f_{\mb \theta}(\mb x_i) - t-1]_+ - [s_i+f_{\mb \theta}(\mb x_i) - t]_+\ge0\;\forall i\in\mc N.\nonumber
\end{align}
\end{minipage}%
  }
\end{center}
}
Note that \cref{eq:FPOR-almost} is a relaxation of \cref{def:FPOR_eq_continuous}, as the feasible set of \cref{eq:FPOR-almost} is a superset of that of \cref{def:FPOR_eq_continuous}. More importantly, the feasible set of \cref{eq:FPOR-almost} has a nontrivial interior with respect to $\mb s$ (due to the equivalence in \cref{lemma:relax-key}), making it computationally stable. For analysis, we sometimes also consider the following relaxed form of \cref{def:FPOR_eq_lifted}: 
\begin{align} \label{eq:proof-frop-key-1}
    & \max\nolimits_{\mb\theta, \mb{s}\in [0, 1]^N, t \in [0, 1]} \quad \tfrac{1}{N_+}\sum\nolimits_{i\in\mc P}s_i \nonumber\\
    & \text{s.t.} \quad {\sum\nolimits_{i\in\mc P}s_i}/{\sum\nolimits_{i\in\mc P\cup\mc N}s_i} \ge \alpha,  \\
    & s_i \le \mb 1\set{f_{\mb \theta}(\mb x_i) > t} \;\forall i\in\mc P, \ s_i \ge \mb 1\set{f_{\mb \theta}(\mb x_i) > t} \;\forall i\in\mc N. \nonumber
\end{align}
Despite the apparent relaxation, \cref{eq:FPOR-almost} enjoys a strong \emph{exactness} property. For convenience, below, we write 
\begin{equation}
\begin{aligned}
    \phi_1(\mb s) & \doteq \tfrac{1}{N_+}\sum\nolimits_{i\in\mc P}s_i, \\  
    \phi_2(\mb s) & \doteq {\sum\nolimits_{i\in\mc P}s_i}/{\sum\nolimits_{i\in\mc P\cup\mc N}s_i}. 
\end{aligned}
\end{equation}
The next result establishes the connection between the feasible points of \cref{def:FPOR_expanded} and of \cref{eq:proof-frop-key-1}. 
\begin{lemma}[equivalence in feasibility of \cref{def:FPOR_expanded} and of \cref{eq:proof-frop-key-1}]\label{thm:feasibility-relax}
    A point $(\mb\theta,t)$ is feasible for \cref{def:FPOR_expanded} if and only if $(\mb\theta, [\mb 1\set{f_{\mb \theta}(\mb x_i) > t}]_i, t)$ is feasible for \cref{eq:proof-frop-key-1}. 
\end{lemma}
The next theorem further connects the feasibility sets of \cref{def:FPOR_expanded} and of \cref{eq:FPOR-almost}, which requires an extra non-singularity assumption on the point compared to \cref{thm:feasibility-relax}. 
\begin{theorem}[equivalence in feasibility of \cref{def:FPOR_expanded} and of \cref{eq:FPOR-almost}]\label{thm:feasibility-retrict}
    The following holds
    \begin{enumerate}[label=(\roman*)]
        \item If a non-singular point $(\mb\theta,t)$ is feasible for \cref{def:FPOR_expanded}, $(\mb\theta, \mb s, t)$ is feasible for \cref{eq:FPOR-almost} for a certain $\mb s$; in particular, $(\mb\theta, [\mb 1\set{f_{\mb \theta}(\mb x_i) > t}]_i, t)$ is feasible for \cref{eq:FPOR-almost}. 
        \item If $(\mb\theta, \mb s, t)$ with non-singular $(\mb \theta, t)$ is feasible for \cref{eq:FPOR-almost} for a certain $\mb s$, $(\mb \theta, t)$ is feasible for \cref{def:FPOR_expanded}. 
    \end{enumerate}
\end{theorem}
\begin{proof}
    We need a couple of important facts: 
    \begin{fact} \label{eq:proof-frop-key-2}
        Both $\phi_1(\mb s)$ and $\phi_2(\mb s)$ over $\mb s \in [0, 1]^N$ are coordinate-wise monotonically nondecreasing with respect to $s_i \ \forall i \in \mc P$ and coordinate-wise monotonically nonincreasing with respect to $s_i \ \forall i \in \mc N$. 
    \end{fact}
    This can be easily verified, and, in turn, implies
    \begin{fact}  \label{eq:proof-frop-key-3}
        If a point $(\mb \theta, \mb s, t)$ is feasible for \cref{eq:proof-frop-key-1}, $(\mb \theta, [\mb 1\set{f_{\mb \theta}(\mb x_i) > t}]_i, t)$ is also feasible for \cref{eq:proof-frop-key-1}. Moreover, $\phi_1([\mb 1\set{f_{\mb \theta}(\mb x_i) > t}]_i) \ge \phi_1(\mb s)$. 
    \end{fact}
    To see it, note that for any $(\mb \theta, \mb s, t)$, $(\mb \theta, [\mb 1\set{f_{\mb \theta}(\mb x_i) > t}]_i, t)$ satisfies the constraint $s_i \le \mb 1\set{f_{\mb \theta}(\mb x_i) > t} \;\forall i\in\mc P, \; s_i \ge \mb 1\set{f_{\mb \theta}(\mb x_i) > t} \;\forall i\in\mc N$ trivially, and 
    \begin{align*}
        \phi_1([\mb 1\set{f_{\mb \theta}(\mb x_i) > t}]_i) & \ge \phi_1(\mb s), \nonumber \\
        \phi_2([\mb 1\set{f_{\mb \theta}(\mb x_i) > t}]_i) & \ge \phi_2(\mb s) \ge \alpha 
    \end{align*}
    due to \cref{eq:proof-frop-key-2}. 

    Next, we prove the claimed equivalence: 
    \begin{itemize}[leftmargin=1em]
    \item \textbf{The $\Longrightarrow$ direction:} If a non-singular point $(\mb\theta,t)$ is feasible for \cref{def:FPOR_expanded}, $(\mb\theta, [\mb 1\set{f_{\mb \theta}(\mb x_i) > t}]_i, t)$ is feasible \cref{eq:proof-frop-key-1} by \cref{thm:feasibility-relax}. Due to \cref{lemma:relax-key}, $(\mb\theta, [\mb 1\set{f_{\mb \theta}(\mb x_i) > t}]_i, t)$ is also feasible for \cref{eq:FPOR-almost}; 
    
    \item \textbf{The $\Longleftarrow$ direction:} Suppose a point $(\mb\theta, \mb s, t)$ with $(\mb \theta, t)$ non-singular is feasible for \cref{eq:FPOR-almost}. Due to \cref{lemma:relax-key}, $(\mb\theta, \mb s, t)$ is feasible for \cref{eq:proof-frop-key-1}. Now, by \cref{eq:proof-frop-key-3}, $(\mb\theta, [\mb 1\set{f_{\mb \theta}(\mb x_i) > t}]_i, t)$ is also feasible for \cref{eq:proof-frop-key-1}. Invoking \cref{thm:feasibility-relax}, we conclude that $(\mb \theta, t)$ is feasible for \cref{def:FPOR_expanded}.  
    \end{itemize}
    completing the proof. 
\end{proof}
The next theorem builds the connection between global solutions of \cref{def:FPOR_expanded} and of \cref{eq:FPOR-almost}. 
\begin{theorem}[equivalence in global solution of \cref{def:FPOR_expanded} and of \cref{eq:FPOR-almost}] \label{thm:exact-fpor}
    Any non-singular $(\mb \theta^\ast, t^\ast)$ is a global solution to \cref{def:FPOR_expanded} if and only if $(\mb \theta^\ast, \mb s^\ast, t^\ast)$ is a global solution to \cref{eq:FPOR-almost} for a certain $\mb s^\ast$. 
\end{theorem}
\begin{proof}
    First, due to \cref{lemma:relax-key} (i.e., \cref{lemma:relax-levelset}),   $(\mb \theta^\ast, \mb s^\ast, t^\ast)$ with non-singular $(\mb \theta^\ast, t^\ast)$ is a global solution to \cref{eq:FPOR-almost} if and only if it is a global solution to \cref{eq:proof-frop-key-1}. So, next we establish the connection between \cref{eq:proof-frop-key-1} and \cref{def:FPOR_expanded} in terms of global solutions. 

    Since \cref{thm:feasibility-retrict} already settles the equivalence in feasibility, here we only need to focus on the optimality in the objective value. Note that for any feasible $(\mb \theta, \mb s, t)$ for \cref{eq:proof-frop-key-1}, $(\mb \theta, \mb 1\set{f_{\mb \theta}(\mb x_i) > t}]_i, t)$ is also feasible and $\phi_1(\mb s) \le \phi_1(\mb 1\set{f_{\mb \theta}(\mb x_i) > t}]_i)$ due to \cref{eq:proof-frop-key-3}, implying that there exists a global solution of the form $(\mb \theta, \mb 1\set{f_{\mb \theta}(\mb x_i) > t}]_i, t)$ for \cref{eq:proof-frop-key-1}. So, we have the following chain of equalities: 
        \begin{align*}
            & \max\set{\phi_1(\mb s): (\mb \theta, \mb s, t) \; \text{feasible for}\; \cref{eq:proof-frop-key-1}} \nonumber \\
            & = \; \max \{\phi_1(\mb 1\set{f_{\mb \theta}(\mb x_i) > t}]_i): \nonumber \\
            & \qquad (\mb \theta, \mb 1\set{f_{\mb \theta}(\mb x_i) > t}]_i, t) \; \text{feasible for}\; \cref{eq:proof-frop-key-1} \} \\
            & = \; \max \{\phi_1(\mb 1\set{f_{\mb \theta}(\mb x_i) > t}]_i): \nonumber \\
            & \qquad (\mb \theta, t) \; \text{feasible for}\; \cref{def:FPOR_expanded} \}  \quad (\text{by \cref{thm:feasibility-relax}}),   
        \end{align*}
        i.e., the three optimal values are equal, as claimed. 
\end{proof}
The equivalence results in \cref{thm:feasibility-retrict} and \cref{thm:exact-fpor} are strong in both theory and practice: In theory, we can globally solve the FPOR problem in \cref{def:FPOR_expanded} by globally solving \cref{eq:FPOR-almost}, due to \cref{thm:exact-fpor}. In practice, due to the nice non-zero gradient property of the $H_t$ function used in \cref{eq:FPOR-almost}---as discussed in \cref{sec:eq_reform}, we can develop gradient-based optimization methods. But global optimization of \cref{eq:FPOR-almost} may or may not be possible, \cref{thm:feasibility-retrict} guarantees that any non-singular pair $(\mb \theta, t)$ numerically found is at least feasible for \cref{def:FPOR_expanded}, ensuring effective control on the precision. 

\subsection{Regularization}\label{sec:regularization}
To avoid finding singular points, i.e., $(\mb \theta, \mb s, t)$ so that $f_{\mb \theta}(\mb x_i) = t$ for any $i$, when numerically optimizing \cref{eq:FPOR-almost}, it is sufficient to push all $\abs{f_{\mb \theta}(\mb x_i) - t}$'s away from zero. Among numerous possibilities, we regularize the objective of \cref{eq:FPOR-almost} by
\begin{multline}
     \label{eq:reg-term}
    \psi(\mb \theta, \mb s) \doteq \textstyle \frac{1}{N}\sum\nolimits_{i\in\mc P\cup \mc N} \\ \brac{w_i(s_i \log{f_{\mb\theta}(\mb x_i)} + (1-s_i)\log{(1-f_{\mb\theta}(\mb x_i)})}, 
\end{multline}
where $w_i = 1/N_+$ if $i \in \mc P$ and $1/N_-$ if $i \in \mc N$, i.e., the inverse of the class frequency, to account for the label imbalance. 

\begin{wrapfigure}{r}{0.3\textwidth}
    \centering
    \vspace{-1em}
    \includegraphics[width=1\linewidth]{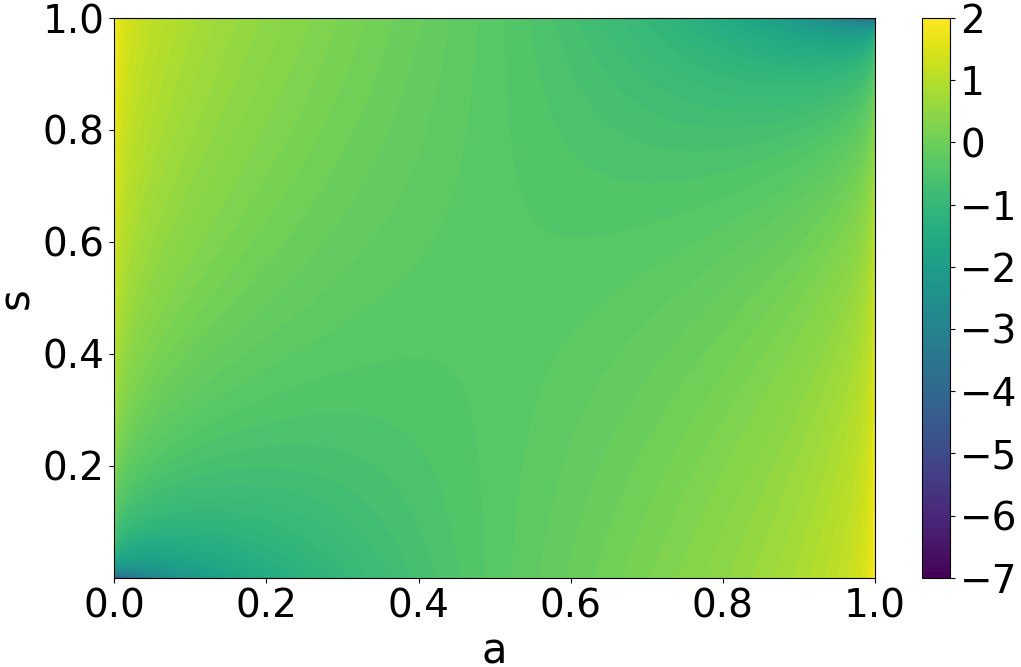}
    \vspace{-2em}
    \caption{Contour plot of $r (a, s) \doteq \log \abs{R(a, s)}$ (value negated and log-scaled for better visualization)}
    \vspace{-1em}
    \label{fig:reg-contour}
\end{wrapfigure}
To see why this works, recall that $\mb s \in [0, 1]^N$ and $f_{\mb \theta}: \mc X \to [0, 1]$. Consider $R(a, s) \allowbreak \doteq \allowbreak a \log s \allowbreak + \allowbreak (1-a) \allowbreak \log (1-s)$ over $[0, 1] \allowbreak \times \allowbreak [0, 1]$. It is maximized when $a = s = 0$ and $a = s =1$; see \cref{fig:reg-contour} for its contour plot. In other words, this regularization encourages both $s$ and $f_{\mb \theta}$ to align with each other and take extreme values together (i.e., from $\set{0, 1}$). This is beneficial because (1) our original lifted reformulation \cref{def:FPOR_eq_lifted} works by introducing $s = \mb 1\set{f_{\mb \theta}(\mb x) > t}$, i.e., $s$ as the predicted label for the given sample. So, ideally, $s$ should have value in $\set{0, 1}$ and $\mb 1\set{f_{\mb \theta}(\mb x) > t}$ should be in agreement with $s$, e.g., achieved when both $s$ and $f_{\mb \theta}(\mb x)$ assume the same extremely value in $\set{0, 1}$, so that we can easily find feasible points; and (2) no matter the value of $t$, driving $f_{\mb \theta}(\mb x_i)$'s to extreme values promotes large decision margins, which can help improve generalization performance, especially when distribution shifts occur in test data. 

\subsection{Optimization by an exact penalty method}\label{subsec:opt}
The inequality-constrained continuous reformulation with regularization $\psi(\mb{\theta},\mb{s})$ for the three DMO problems can be expressed as follows (for details, see \cref{eq:FPOR-almost} on FPOR; \cref{eq:app_dmo_def,def:dmo_original} on the unified form):
\begin{equation}\label{eq:unify}
\begin{aligned}
    \max\nolimits_{\mb\theta,\mb{s}\in [0, 1]^N,t\in[0,1]}\ \phi_{\text{obj}}(\mb{s})+\gamma\psi(\mb{\theta},\mb{s})  \\
    \text{s.t.} \quad \phi_{\text{con}}(\mb{s})\leq 0,\ \eta(\mb\theta,\mb{s},t)\leq \mb{0},
\end{aligned}
\end{equation}
where $\gamma>0$ is the regularization parameter. For example, in FPOR
\begin{align}
\phi_{\text{obj}}(\mb{s}) & = \sum\nolimits_{i\in\mc P}s_i / N_+\\
\phi_{\text{con}}(\mb{s}) &  = \alpha\sum\nolimits_{i\in\mc N} s_i - (1-\alpha)\sum\nolimits_{i\in\mc P} s_i\\
(\eta(\mb\theta,\mb{s},t))_i &= \begin{cases}
    H_t(f_{\mb \theta}(\mb{x}_i),s_i) & i\in\mc P\\
    -H_t(f_{\mb \theta}(\mb{x}_i),s_i) & i\in\mc N
\end{cases}
\end{align}
where $H_t$ is defined in \cref{def:g-h}. Note that, to obtain any meaningful solution to the DMO problem, it is essential to satisfy the constraints $\eta(\mb\theta,\mb{s},t)\leq \mb{0}$. Moreover, to make FPOR and FROP practically useful, we need to enforce the metric constraints $\phi_{\text{con}}(\mb{s})\leq 0$. Therefore, any optimization method used to solve the reformulated problem must be able to reliably find a feasible point in the first place. While the use of the quadratic penalty method is pretty common for constrained optimization problems in practice, the feasibility can only be guaranteed asymptotically by increasing the value of the penalty parameter to infinity~\cite{NoceWrig06}. Lagrangian methods are another popular choice, e.g., in TFCO~\cite{cotter2019two}. But finding feasible point is also not guaranteed in general, unless the Lagrangian multiplier is close to the optimal~\cite{bertsekas2016nonlinear}. We instead choose an \textit{exact} penalty method~\cite{han1979exact} with an $\ell_1$-type penalty function, which ensures that a feasible solution can be obtained for a sufficiently large---but finite---penalty parameter~\cite{di1994exact}. The Augmented Lagrangian Method (ALM) used in our preliminary work~\cite{travadi2023direct} works fine also, but the inclusion of the squared penalty term in the augmented Lagrangian function makes future extensions of our algorithm to the stochastic setting tricky~\cite{alacaoglu2024complexity}.

We now describe an exact penalty method to solve \cref{eq:unify}. The exact penalty function associated with \cref{eq:unify} is defined as 
\begin{multline}
        \mathcal{F}(\mb{\theta},\mb{s},t,\lambda) \doteq -\phi_{\text{obj}}(\mb{s})-\gamma\psi(\mb{\theta},\mb{s}) \ +  \\
        \lambda\left( [\phi_{\text{con}}(\mb{s})]_+ + \sum\nolimits_{i\in\mc P\cup\mc N} [(\eta(\mb\theta,\mb{s},t))_i]_+\right), 
\end{multline}
where $\lambda>0$ is the penalty parameter. For an increasing sequence of penalty parameters $\lambda^{(1)} \le \cdots \le \lambda^{(K)}$, in the $k^{th}$ iteration, the exact penalty method solves an unconstrained optimization problem:
\begin{align}
(\mb{\theta}^{k+1},\mb{s}^{k+1},t^{k+1})& \approx\argmin_{\mb{\theta},\mb{s}\in [0, 1]^N,t \in [0, 1]} \mathcal{F}(\mb{\theta},\mb{s},t,\lambda^{(k)}).
\end{align}
The detailed algorithm is outlined in \cref{alg:alm}. 
\begin{algorithm}[!htbp]
    \caption{An exact penalty method for solving the unified DMO problem \cref{eq:unify}}
    \begin{algorithmic}[1]
        \State\textbf{input:} initial penalty parameter $\lambda^{(0)}$, initial point $(\mb{\theta}^0,\mb{s}^0,t^0)$, penalty multiplier $\rho$, maximum iteration $K$, regularization parameter $\gamma$. Initialize $k=0$.
        \While{$k\le K$}
            \State Apply a solver with initial point $(\mb{\theta}^k,\mb{s}^k,t^k)$ to find an approximate solution $(\mb{\theta}^{k+1},\mb{s}^{k+1},t^{k+1})$ to 
            \[
            \min\nolimits_{\mb{\theta},\mb{s}\in [0, 1]^N,t\in [0,1]}\ \mathcal{F}(\mb{\theta},\mb{s},t,\lambda^{(k)}).
            \]
            \State Set $\lambda^{(k+1)} = \lambda^{(k)} \times \rho$. \Comment{update the penalty parameter}
            \State{Set $k \gets k+1$}.
        \EndWhile
    \end{algorithmic}\label{alg:alm}
\end{algorithm} 
For the subproblem solver, we can choose projected gradient style methods, e.g., ADAM with per-iteration projection onto the simple constraint set $\mb s \in [0, 1]^N, t \in [0, 1]$. 

\section{Experiments}\label{sec:exp}

\subsection{Experimental settings}
\subsubsection{Datasets} 
We evaluate our exact reformulation and optimization (ERO) method for solving the three DMO problems (\textbf{DMO+ERO}) over four datasets, covering image, text, and tabular data: Two image datasets are \textit{Eyepacs}~\cite{diabetic-retinopathy-detection} and MURA~\cite{rajpurkar2017mura}, one text dataset is \textit{ADE-Corpus-V2}~\cite{GURULINGAPPA2012885} from huggingface, and one tabular dataset \textit{creditcard}~\cite{dal2015calibrating} from Kaggle. For all datasets except MURA, train/test splits are generated using a fixed split ratio with stratified sampling to preserve class imbalance. For MURA, we use the official train/test split provided by the dataset. More details can be found in \cref{sec:supp_results}.

\begin{table*}[!htbp]
\def\arraystretch{1.3}
\centering
\caption{Comparison results on \textcolor{red}{FPOR}. Values in (parentheses) are results after TA.}
\vspace{-0.5em}
\resizebox{1\textwidth}{!}{
\begin{tabular}{ll||cc||cc:c}
\Xhline{2\arrayrulewidth}
& &\multicolumn{2}{c||}{\textbf{train}} & \multicolumn{3}{c}{\textbf{test}} \\
dataset & method  & \multicolumn{1}{c}{precision---feasibility} & \multicolumn{1}{c||}{recall---objective} & \multicolumn{1}{c}{precision---feasibility} & \multicolumn{1}{c}{recall---objective} & F1-score  \\
\hline

	& WCE	& 0.190 $\pm$ {\tiny 0.131} (\underline{0.944 $\pm$ {\tiny 0.040}})	& 0.866 $\pm$ {\tiny 0.018} (0.170 $\pm$ {\tiny 0.013})	& 0.194 $\pm$ {\tiny 0.127} (\underline{0.901 $\pm$ {\tiny 0.087}})	& 0.844 $\pm$ {\tiny 0.034} (\textcolor{blue}{0.164 $\pm$ {\tiny 0.021}}) & 0.296 $\pm$ {\tiny 0.130} (0.277 $\pm$ {\tiny 0.032}) \\
	& Focal	& \underline{0.944 $\pm$ {\tiny 0.048}} (\underline{0.924 $\pm$ {\tiny 0.033}})	& 0.278 $\pm$ {\tiny 0.293} (0.259 $\pm$ {\tiny 0.221})	& 0.797 $\pm$ {\tiny 0.289} (0.833 $\pm$ {\tiny 0.152})	& 0.249 $\pm$ {\tiny 0.267} (0.230 $\pm$ {\tiny 0.197}) & 0.312 $\pm$ {\tiny 0.309} (0.323 $\pm$ {\tiny 0.233}) \\
	& LDAM	& 0.102 $\pm$ {\tiny 0.299} (0.312 $\pm$ {\tiny 0.450})	& 0.146 $\pm$ {\tiny 0.190} (0.315 $\pm$ {\tiny 0.448})	& 0.103 $\pm$ {\tiny 0.299} (0.321 $\pm$ {\tiny 0.446})	& 0.142 $\pm$ {\tiny 0.186} (0.316 $\pm$ {\tiny 0.446}) & 0.012 $\pm$ {\tiny 0.021} (0.029 $\pm$ {\tiny 0.030}) \\
	& TERM	& \underline{0.963 $\pm$ {\tiny 0.057}} (\underline{0.938 $\pm$ {\tiny 0.044}})	& 0.226 $\pm$ {\tiny 0.303} (0.095 $\pm$ {\tiny 0.062})	& 0.816 $\pm$ {\tiny 0.293} (0.864 $\pm$ {\tiny 0.118})	& 0.199 $\pm$ {\tiny 0.274} (0.093 $\pm$ {\tiny 0.060}) & 0.241 $\pm$ {\tiny 0.304} (0.163 $\pm$ {\tiny 0.096}) \\
	& ROS	& 0.372 $\pm$ {\tiny 0.095} (\underline{0.915 $\pm$ {\tiny 0.019}})	& 0.832 $\pm$ {\tiny 0.008} (0.208 $\pm$ {\tiny 0.028})	& 0.384 $\pm$ {\tiny 0.102} (0.798 $\pm$ {\tiny 0.048})	& 0.787 $\pm$ {\tiny 0.010} (0.209 $\pm$ {\tiny 0.021}) & 0.508 $\pm$ {\tiny 0.095} (0.331 $\pm$ {\tiny 0.027}) \\
	& RUS	& 0.379 $\pm$ {\tiny 0.179} (\underline{0.918 $\pm$ {\tiny 0.033}})	& 0.835 $\pm$ {\tiny 0.014} (0.203 $\pm$ {\tiny 0.027})	& 0.386 $\pm$ {\tiny 0.170} (0.769 $\pm$ {\tiny 0.035})	& 0.792 $\pm$ {\tiny 0.020} (0.204 $\pm$ {\tiny 0.032}) & 0.497 $\pm$ {\tiny 0.140} (0.321 $\pm$ {\tiny 0.044}) \\
	& DMO+TFCO	& 0.005 $\pm$ {\tiny 0.003} (0.493 $\pm$ {\tiny 0.353})	& 0.909 $\pm$ {\tiny 0.047} (0.196 $\pm$ {\tiny 0.114})	& 0.005 $\pm$ {\tiny 0.003} (0.453 $\pm$ {\tiny 0.283})	& 0.879 $\pm$ {\tiny 0.071} (0.202 $\pm$ {\tiny 0.106}) & 0.010 $\pm$ {\tiny 0.005} (0.236 $\pm$ {\tiny 0.112}) \\
\rowcolor{gray}
\multirow{-8}{*}{\cellcolor{white}creditcard}	& DMO+ERO	& \underline{0.965 $\pm$ {\tiny 0.040}} (\underline{0.968 $\pm$ {\tiny 0.036}})	& \textcolor{red}{1.000 $\pm$ {\tiny 0.000}} (\textcolor{blue}{1.000 $\pm$ {\tiny 0.000}})	& 0.837 $\pm$ {\tiny 0.038} (0.835 $\pm$ {\tiny 0.036})	& 0.705 $\pm$ {\tiny 0.034} (0.705 $\pm$ {\tiny 0.032}) & \textcolor{red}{0.764 $\pm$ {\tiny 0.017}} (\textcolor{blue}{0.763 $\pm$ {\tiny 0.018}}) \\

\hline

	& WCE	& 0.685 $\pm$ {\tiny 0.043} (\underline{0.940 $\pm$ {\tiny 0.041}})	& 0.157 $\pm$ {\tiny 0.059} (0.002 $\pm$ {\tiny 0.002})	& 0.621 $\pm$ {\tiny 0.062} (0.819 $\pm$ {\tiny 0.113})	& 0.169 $\pm$ {\tiny 0.062} (0.002 $\pm$ {\tiny 0.002}) & 0.260 $\pm$ {\tiny 0.092} (0.004 $\pm$ {\tiny 0.004}) \\
	& Focal	& 0.876 $\pm$ {\tiny 0.044} (\underline{0.935 $\pm$ {\tiny 0.043}})	& 0.041 $\pm$ {\tiny 0.034} (0.008 $\pm$ {\tiny 0.007})	& 0.766 $\pm$ {\tiny 0.256} (0.804 $\pm$ {\tiny 0.186})	& 0.040 $\pm$ {\tiny 0.033} (0.008 $\pm$ {\tiny 0.008}) & 0.075 $\pm$ {\tiny 0.056} (0.016 $\pm$ {\tiny 0.015}) \\
	& LDAM	& 0.892 $\pm$ {\tiny 0.088} (\underline{0.954 $\pm$ {\tiny 0.046}})	& 0.033 $\pm$ {\tiny 0.027} (0.005 $\pm$ {\tiny 0.007})	& 0.562 $\pm$ {\tiny 0.330} (0.606 $\pm$ {\tiny 0.359})	& 0.036 $\pm$ {\tiny 0.030} (0.005 $\pm$ {\tiny 0.007}) & 0.068 $\pm$ {\tiny 0.055} (0.011 $\pm$ {\tiny 0.014}) \\
	& TERM	& 0.724 $\pm$ {\tiny 0.011} (\underline{0.941 $\pm$ {\tiny 0.049}})	& 0.150 $\pm$ {\tiny 0.050} (0.005 $\pm$ {\tiny 0.007})	& 0.656 $\pm$ {\tiny 0.124} (0.835 $\pm$ {\tiny 0.173})	& 0.154 $\pm$ {\tiny 0.051} (0.005 $\pm$ {\tiny 0.007}) & 0.247 $\pm$ {\tiny 0.082} (0.009 $\pm$ {\tiny 0.014}) \\
	& ROS	& 0.709 $\pm$ {\tiny 0.098} (\underline{0.945 $\pm$ {\tiny 0.046}})	& 0.155 $\pm$ {\tiny 0.063} (0.002 $\pm$ {\tiny 0.002})	& 0.583 $\pm$ {\tiny 0.195} (0.773 $\pm$ {\tiny 0.265})	& 0.168 $\pm$ {\tiny 0.065} (0.002 $\pm$ {\tiny 0.002}) & 0.259 $\pm$ {\tiny 0.094} (0.004 $\pm$ {\tiny 0.005}) \\
	& RUS	& 0.688 $\pm$ {\tiny 0.049} (\underline{0.913 $\pm$ {\tiny 0.031}})	& 0.160 $\pm$ {\tiny 0.061} (0.003 $\pm$ {\tiny 0.002})	& 0.655 $\pm$ {\tiny 0.034} (0.844 $\pm$ {\tiny 0.052})	& 0.171 $\pm$ {\tiny 0.065} (0.003 $\pm$ {\tiny 0.003}) & \textcolor{red}{0.263 $\pm$ {\tiny 0.094}} (0.006 $\pm$ {\tiny 0.006}) \\
	& DMO+TFCO	& 0.450 $\pm$ {\tiny 0.210} (0.703 $\pm$ {\tiny 0.263})	& 0.301 $\pm$ {\tiny 0.458} (0.103 $\pm$ {\tiny 0.299})	& 0.493 $\pm$ {\tiny 0.279} (0.377 $\pm$ {\tiny 0.290})	& 0.301 $\pm$ {\tiny 0.458} (0.103 $\pm$ {\tiny 0.299}) & 0.126 $\pm$ {\tiny 0.189} (0.047 $\pm$ {\tiny 0.124}) \\
\rowcolor{gray}
\multirow{-8}{*}{\cellcolor{white}Eyepacs}	& DMO+ERO	& \underline{0.901 $\pm$ {\tiny 0.001}} (\underline{0.901 $\pm$ {\tiny 0.001}})	& \textcolor{red}{0.115 $\pm$ {\tiny 0.102}} (\textcolor{blue}{0.115 $\pm$ {\tiny 0.102}})	& 0.794 $\pm$ {\tiny 0.045} (0.799 $\pm$ {\tiny 0.040})	& 0.107 $\pm$ {\tiny 0.094} (0.105 $\pm$ {\tiny 0.093}) & 0.175 $\pm$ {\tiny 0.144} (\textcolor{blue}{0.173 $\pm$ {\tiny 0.144}}) \\

\hline

	& WCE	& 0.709 $\pm$ {\tiny 0.030} (0.887 $\pm$ {\tiny 0.070})	& 0.277 $\pm$ {\tiny 0.238} (0.068 $\pm$ {\tiny 0.173})	& 0.722 $\pm$ {\tiny 0.023} (0.485 $\pm$ {\tiny 0.406})	& 0.273 $\pm$ {\tiny 0.238} (0.062 $\pm$ {\tiny 0.166}) & 0.346 $\pm$ {\tiny 0.213} (0.080 $\pm$ {\tiny 0.203}) \\
	& Focal	& \underline{0.903 $\pm$ {\tiny 0.006}} (\underline{0.901 $\pm$ {\tiny 0.001}})	& 0.190 $\pm$ {\tiny 0.098} (0.220 $\pm$ {\tiny 0.100})	& \underline{0.913 $\pm$ {\tiny 0.021}} (\underline{0.902 $\pm$ {\tiny 0.013}})	& \textcolor{red}{0.181 $\pm$ {\tiny 0.091}} (0.210 $\pm$ {\tiny 0.093}) & 0.292 $\pm$ {\tiny 0.129} (0.329 $\pm$ {\tiny 0.129}) \\
	& LDAM	& 0.808 $\pm$ {\tiny 0.005} (\underline{0.900 $\pm$ {\tiny 0.000}})	& 0.494 $\pm$ {\tiny 0.060} (0.262 $\pm$ {\tiny 0.078})	& 0.800 $\pm$ {\tiny 0.009} (\underline{0.906 $\pm$ {\tiny 0.014}})	& 0.482 $\pm$ {\tiny 0.057} (\textcolor{blue}{0.252 $\pm$ {\tiny 0.073}}) & 0.599 $\pm$ {\tiny 0.044} (0.388 $\pm$ {\tiny 0.095}) \\
	& TERM	& 0.762 $\pm$ {\tiny 0.006} (0.892 $\pm$ {\tiny 0.043})	& 0.374 $\pm$ {\tiny 0.263} (0.149 $\pm$ {\tiny 0.146})	& 0.765 $\pm$ {\tiny 0.014} (0.652 $\pm$ {\tiny 0.365})	& 0.362 $\pm$ {\tiny 0.258} (0.141 $\pm$ {\tiny 0.141}) & 0.433 $\pm$ {\tiny 0.249} (0.214 $\pm$ {\tiny 0.214}) \\
	& ROS	& 0.701 $\pm$ {\tiny 0.033} (\underline{0.915 $\pm$ {\tiny 0.065}})	& 0.491 $\pm$ {\tiny 0.318} (0.185 $\pm$ {\tiny 0.197})	& 0.707 $\pm$ {\tiny 0.033} (0.540 $\pm$ {\tiny 0.441})	& 0.481 $\pm$ {\tiny 0.314} (0.175 $\pm$ {\tiny 0.190}) & 0.499 $\pm$ {\tiny 0.252} (0.248 $\pm$ {\tiny 0.255}) \\
	& RUS	& 0.712 $\pm$ {\tiny 0.032} (0.853 $\pm$ {\tiny 0.084})	& 0.432 $\pm$ {\tiny 0.349} (0.217 $\pm$ {\tiny 0.240})	& 0.710 $\pm$ {\tiny 0.045} (0.640 $\pm$ {\tiny 0.333})	& 0.422 $\pm$ {\tiny 0.347} (0.205 $\pm$ {\tiny 0.229}) & 0.444 $\pm$ {\tiny 0.277} (0.272 $\pm$ {\tiny 0.288}) \\
	& DMO+TFCO	& 0.347 $\pm$ {\tiny 0.060} (0.594 $\pm$ {\tiny 0.139})	& 0.563 $\pm$ {\tiny 0.313} (0.010 $\pm$ {\tiny 0.018})	& 0.348 $\pm$ {\tiny 0.064} (0.446 $\pm$ {\tiny 0.340})	& 0.558 $\pm$ {\tiny 0.312} (0.008 $\pm$ {\tiny 0.015}) & 0.381 $\pm$ {\tiny 0.097} (0.016 $\pm$ {\tiny 0.028}) \\
\rowcolor{gray}
\multirow{-8}{*}{\cellcolor{white}ADE-v2}	& DMO+ERO	& \underline{0.900 $\pm$ {\tiny 0.001}} (\underline{0.900 $\pm$ {\tiny 0.000}})	& \textcolor{red}{0.803 $\pm$ {\tiny 0.010}} (\textcolor{blue}{0.778 $\pm$ {\tiny 0.081}})	& 0.876 $\pm$ {\tiny 0.004} (0.877 $\pm$ {\tiny 0.008})	& 0.769 $\pm$ {\tiny 0.012} (0.746 $\pm$ {\tiny 0.079}) & \textcolor{red}{0.819 $\pm$ {\tiny 0.007}} (\textcolor{blue}{0.803 $\pm$ {\tiny 0.051}}) \\

\hline

	& WCE	& 0.451 $\pm$ {\tiny 0.101} (0.844 $\pm$ {\tiny 0.191})	& 0.118 $\pm$ {\tiny 0.134} (0.005 $\pm$ {\tiny 0.007})	& \underline{0.955 $\pm$ {\tiny 0.047}} (\underline{1.000 $\pm$ {\tiny 0.000}})	& 0.117 $\pm$ {\tiny 0.131} (0.005 $\pm$ {\tiny 0.007}) & 0.182 $\pm$ {\tiny 0.195} (0.010 $\pm$ {\tiny 0.015}) \\
	& Focal	& 0.678 $\pm$ {\tiny 0.132} (0.870 $\pm$ {\tiny 0.165})	& 0.106 $\pm$ {\tiny 0.059} (0.021 $\pm$ {\tiny 0.028})	& \underline{0.974 $\pm$ {\tiny 0.017}} (\underline{0.989 $\pm$ {\tiny 0.013}})	& 0.102 $\pm$ {\tiny 0.053} (0.020 $\pm$ {\tiny 0.028}) & 0.180 $\pm$ {\tiny 0.089} (0.037 $\pm$ {\tiny 0.052}) \\
	& LDAM	& 0.770 $\pm$ {\tiny 0.094} (\underline{0.974 $\pm$ {\tiny 0.062}})	& 0.095 $\pm$ {\tiny 0.051} (0.003 $\pm$ {\tiny 0.005})	& 0.838 $\pm$ {\tiny 0.342} (0.857 $\pm$ {\tiny 0.350})	& 0.083 $\pm$ {\tiny 0.047} (0.002 $\pm$ {\tiny 0.003}) & 0.150 $\pm$ {\tiny 0.080} (0.004 $\pm$ {\tiny 0.006}) \\
	& TERM	& 0.649 $\pm$ {\tiny 0.152} (\underline{0.964 $\pm$ {\tiny 0.087}})	& 0.189 $\pm$ {\tiny 0.120} (0.002 $\pm$ {\tiny 0.001})	& \underline{0.970 $\pm$ {\tiny 0.019}} (0.857 $\pm$ {\tiny 0.350})	& 0.176 $\pm$ {\tiny 0.111} (0.002 $\pm$ {\tiny 0.001}) & 0.281 $\pm$ {\tiny 0.176} (0.003 $\pm$ {\tiny 0.002}) \\
	& ROS	& 0.389 $\pm$ {\tiny 0.054} (\underline{0.957 $\pm$ {\tiny 0.088}})	& 0.113 $\pm$ {\tiny 0.101} (0.002 $\pm$ {\tiny 0.002})	& \underline{0.916 $\pm$ {\tiny 0.033}} (0.600 $\pm$ {\tiny 0.490})	& 0.108 $\pm$ {\tiny 0.096} (0.001 $\pm$ {\tiny 0.002}) & 0.179 $\pm$ {\tiny 0.147} (0.002 $\pm$ {\tiny 0.003}) \\
	& RUS	& 0.397 $\pm$ {\tiny 0.055} (\underline{0.942 $\pm$ {\tiny 0.118}})	& 0.091 $\pm$ {\tiny 0.114} (0.001 $\pm$ {\tiny 0.001})	& \underline{0.951 $\pm$ {\tiny 0.041}} (0.800 $\pm$ {\tiny 0.400})	& 0.092 $\pm$ {\tiny 0.111} (0.001 $\pm$ {\tiny 0.001}) & 0.148 $\pm$ {\tiny 0.162} (0.002 $\pm$ {\tiny 0.001}) \\
	& DMO+TFCO	& 0.074 $\pm$ {\tiny 0.011} (0.313 $\pm$ {\tiny 0.318})	& 0.435 $\pm$ {\tiny 0.334} (0.216 $\pm$ {\tiny 0.358})	& 0.451 $\pm$ {\tiny 0.041} (0.511 $\pm$ {\tiny 0.384})	& 0.379 $\pm$ {\tiny 0.329} (0.197 $\pm$ {\tiny 0.341}) & 0.334 $\pm$ {\tiny 0.226} (0.153 $\pm$ {\tiny 0.246}) \\
\rowcolor{gray}
\multirow{-8}{*}{\cellcolor{white}MURA}	& DMO+ERO	& \underline{0.905 $\pm$ {\tiny 0.005}} (\underline{0.905 $\pm$ {\tiny 0.005}})	& \textcolor{red}{0.401 $\pm$ {\tiny 0.050}} (\textcolor{blue}{0.402 $\pm$ {\tiny 0.050}})	& \underline{0.944 $\pm$ {\tiny 0.007}} (\underline{0.943 $\pm$ {\tiny 0.007}})	& \textcolor{red}{0.252 $\pm$ {\tiny 0.023}} (\textcolor{blue}{0.256 $\pm$ {\tiny 0.020}}) & \textcolor{red}{0.398 $\pm$ {\tiny 0.029}} (\textcolor{blue}{0.403 $\pm$ {\tiny 0.024}}) \\

\hline
\Xhline{3\arrayrulewidth}
\end{tabular}
}
\label{table:FPOR}
\vspace{-1.5em}
\end{table*}

\begin{table*}[!htbp]
\def\arraystretch{1.3}
\centering
\caption{Comparison results on \textcolor{red}{FROP}. Values in (parentheses) are results after TA.}
\vspace{-0.5em}
\resizebox{1\textwidth}{!}{
\begin{tabular}{ll||cc||cc:c}
\Xhline{2\arrayrulewidth}
& &\multicolumn{2}{c||}{\textbf{train}} & \multicolumn{3}{c}{\textbf{test}} \\
dataset & method  & \multicolumn{1}{c}{recall---feasibility} & \multicolumn{1}{c||}{precision---objective} & \multicolumn{1}{c}{recall---feasibility} & \multicolumn{1}{c}{precision---objective} & F1-score  \\
\hline

	& WCE	& \underline{0.980 $\pm$ {\tiny 0.035}} (\underline{0.949 $\pm$ {\tiny 0.039}})	& 0.732 $\pm$ {\tiny 0.320} (0.820 $\pm$ {\tiny 0.353})	& 0.775 $\pm$ {\tiny 0.067} (0.693 $\pm$ {\tiny 0.108})	& 0.564 $\pm$ {\tiny 0.233} (0.759 $\pm$ {\tiny 0.319}) & 0.604 $\pm$ {\tiny 0.213} (0.648 $\pm$ {\tiny 0.219}) \\
	& Focal	& 0.772 $\pm$ {\tiny 0.012} (\underline{0.902 $\pm$ {\tiny 0.000}})	& 0.851 $\pm$ {\tiny 0.006} (0.003 $\pm$ {\tiny 0.000})	& 0.667 $\pm$ {\tiny 0.022} (0.865 $\pm$ {\tiny 0.018})	& 0.802 $\pm$ {\tiny 0.007} (0.003 $\pm$ {\tiny 0.000}) & 0.728 $\pm$ {\tiny 0.014} (0.006 $\pm$ {\tiny 0.001}) \\
	& LDAM	& 0.234 $\pm$ {\tiny 0.217} (\underline{0.983 $\pm$ {\tiny 0.034}})	& 0.003 $\pm$ {\tiny 0.003} (0.002 $\pm$ {\tiny 0.000})	& 0.228 $\pm$ {\tiny 0.217} (\underline{0.975 $\pm$ {\tiny 0.034}})	& 0.003 $\pm$ {\tiny 0.003} (\textcolor{blue}{0.002 $\pm$ {\tiny 0.000}}) & 0.006 $\pm$ {\tiny 0.006} (0.004 $\pm$ {\tiny 0.000}) \\
	& TERM	& 0.797 $\pm$ {\tiny 0.004} (\underline{0.902 $\pm$ {\tiny 0.000}})	& 0.867 $\pm$ {\tiny 0.005} (0.033 $\pm$ {\tiny 0.032})	& 0.699 $\pm$ {\tiny 0.017} (0.881 $\pm$ {\tiny 0.015})	& 0.838 $\pm$ {\tiny 0.012} (0.035 $\pm$ {\tiny 0.032}) & \textcolor{red}{0.762 $\pm$ {\tiny 0.011}} (0.065 $\pm$ {\tiny 0.058}) \\
	& ROS	& \underline{0.961 $\pm$ {\tiny 0.047}} (\underline{0.944 $\pm$ {\tiny 0.039}})	& 0.571 $\pm$ {\tiny 0.409} (0.628 $\pm$ {\tiny 0.453})	& 0.792 $\pm$ {\tiny 0.067} (0.742 $\pm$ {\tiny 0.108})	& 0.420 $\pm$ {\tiny 0.283} (0.570 $\pm$ {\tiny 0.404}) & 0.470 $\pm$ {\tiny 0.272} (0.511 $\pm$ {\tiny 0.302}) \\
	& RUS	& \underline{0.903 $\pm$ {\tiny 0.003}} (\underline{0.902 $\pm$ {\tiny 0.000}})	& 0.062 $\pm$ {\tiny 0.007} (0.065 $\pm$ {\tiny 0.007})	& 0.883 $\pm$ {\tiny 0.013} (0.880 $\pm$ {\tiny 0.013})	& 0.066 $\pm$ {\tiny 0.007} (0.069 $\pm$ {\tiny 0.007}) & 0.124 $\pm$ {\tiny 0.012} (0.128 $\pm$ {\tiny 0.012}) \\
	& DMO+TFCO	& \underline{0.936 $\pm$ {\tiny 0.036}} (\underline{0.902 $\pm$ {\tiny 0.001}})	& 0.012 $\pm$ {\tiny 0.012} (0.018 $\pm$ {\tiny 0.012})	& \underline{0.919 $\pm$ {\tiny 0.048}} (0.875 $\pm$ {\tiny 0.020})	& \textcolor{red}{0.013 $\pm$ {\tiny 0.013}} (0.019 $\pm$ {\tiny 0.013}) & 0.025 $\pm$ {\tiny 0.025} (0.037 $\pm$ {\tiny 0.024}) \\
\rowcolor{gray}
\multirow{-8}{*}{\cellcolor{white}creditcard}	& DMO+ERO	& \underline{1.000 $\pm$ {\tiny 0.000}} (\underline{1.000 $\pm$ {\tiny 0.000}})	& \textcolor{red}{0.999 $\pm$ {\tiny 0.002}} (\textcolor{blue}{0.999 $\pm$ {\tiny 0.002}})	& 0.668 $\pm$ {\tiny 0.022} (0.669 $\pm$ {\tiny 0.022})	& 0.877 $\pm$ {\tiny 0.027} (0.873 $\pm$ {\tiny 0.030}) & 0.758 $\pm$ {\tiny 0.019} (\textcolor{blue}{0.757 $\pm$ {\tiny 0.019}}) \\

\hline

	& WCE	& \underline{0.919 $\pm$ {\tiny 0.010}} (\underline{0.900 $\pm$ {\tiny 0.000}})	& 0.296 $\pm$ {\tiny 0.004} (0.303 $\pm$ {\tiny 0.002})	& \underline{0.921 $\pm$ {\tiny 0.012}} (\underline{0.901 $\pm$ {\tiny 0.011}})	& 0.288 $\pm$ {\tiny 0.005} (0.294 $\pm$ {\tiny 0.004}) & 0.438 $\pm$ {\tiny 0.004} (0.443 $\pm$ {\tiny 0.004}) \\
	& Focal	& \underline{0.922 $\pm$ {\tiny 0.015}} (\underline{0.900 $\pm$ {\tiny 0.000}})	& 0.291 $\pm$ {\tiny 0.006} (0.299 $\pm$ {\tiny 0.003})	& \underline{0.911 $\pm$ {\tiny 0.023}} (0.886 $\pm$ {\tiny 0.015})	& 0.288 $\pm$ {\tiny 0.008} (0.296 $\pm$ {\tiny 0.006}) & 0.438 $\pm$ {\tiny 0.006} (0.443 $\pm$ {\tiny 0.005}) \\
	& LDAM	& \underline{0.970 $\pm$ {\tiny 0.033}} (\underline{0.900 $\pm$ {\tiny 0.001}})	& 0.275 $\pm$ {\tiny 0.011} (0.292 $\pm$ {\tiny 0.004})	& \underline{0.979 $\pm$ {\tiny 0.022}} (\underline{0.930 $\pm$ {\tiny 0.009}})	& 0.267 $\pm$ {\tiny 0.006} (0.279 $\pm$ {\tiny 0.003}) & 0.420 $\pm$ {\tiny 0.005} (0.429 $\pm$ {\tiny 0.003}) \\
	& TERM	& \underline{0.920 $\pm$ {\tiny 0.019}} (\underline{0.900 $\pm$ {\tiny 0.000}})	& 0.292 $\pm$ {\tiny 0.006} (0.298 $\pm$ {\tiny 0.002})	& \underline{0.943 $\pm$ {\tiny 0.014}} (\underline{0.929 $\pm$ {\tiny 0.006}})	& 0.278 $\pm$ {\tiny 0.004} (0.282 $\pm$ {\tiny 0.003}) & 0.429 $\pm$ {\tiny 0.004} (0.433 $\pm$ {\tiny 0.002}) \\
	& ROS	& \underline{0.917 $\pm$ {\tiny 0.013}} (\underline{0.900 $\pm$ {\tiny 0.000}})	& 0.296 $\pm$ {\tiny 0.005} (0.303 $\pm$ {\tiny 0.002})	& \underline{0.922 $\pm$ {\tiny 0.014}} (\underline{0.905 $\pm$ {\tiny 0.011}})	& 0.288 $\pm$ {\tiny 0.005} (0.293 $\pm$ {\tiny 0.005}) & 0.438 $\pm$ {\tiny 0.004} (0.443 $\pm$ {\tiny 0.004}) \\
	& RUS	& \underline{0.914 $\pm$ {\tiny 0.010}} (\underline{0.900 $\pm$ {\tiny 0.000}})	& 0.298 $\pm$ {\tiny 0.003} (0.304 $\pm$ {\tiny 0.002})	& \underline{0.914 $\pm$ {\tiny 0.011}} (\underline{0.900 $\pm$ {\tiny 0.010}})	& 0.291 $\pm$ {\tiny 0.004} (\textcolor{blue}{0.295 $\pm$ {\tiny 0.004}}) & 0.441 $\pm$ {\tiny 0.003} (0.445 $\pm$ {\tiny 0.003}) \\
	& DMO+TFCO	& \underline{0.981 $\pm$ {\tiny 0.026}} (\underline{0.930 $\pm$ {\tiny 0.046}})	& 0.269 $\pm$ {\tiny 0.006} (0.282 $\pm$ {\tiny 0.011})	& \underline{0.983 $\pm$ {\tiny 0.024}} (\underline{0.937 $\pm$ {\tiny 0.043}})	& 0.265 $\pm$ {\tiny 0.005} (0.275 $\pm$ {\tiny 0.009}) & 0.418 $\pm$ {\tiny 0.004} (0.425 $\pm$ {\tiny 0.007}) \\
\rowcolor{gray}
\multirow{-8}{*}{\cellcolor{white}Eyepacs}	& DMO+ERO	& \underline{0.920 $\pm$ {\tiny 0.040}} (\underline{0.900 $\pm$ {\tiny 0.000}})	& \textcolor{red}{0.312 $\pm$ {\tiny 0.038}} (\textcolor{blue}{0.315 $\pm$ {\tiny 0.034}})	& \underline{0.906 $\pm$ {\tiny 0.060}} (0.893 $\pm$ {\tiny 0.043})	& \textcolor{red}{0.301 $\pm$ {\tiny 0.036}} (0.303 $\pm$ {\tiny 0.034}) & \textcolor{red}{0.449 $\pm$ {\tiny 0.033}} (\textcolor{blue}{0.451 $\pm$ {\tiny 0.032}}) \\

\hline

	& WCE	& \underline{0.909 $\pm$ {\tiny 0.010}} (\underline{0.900 $\pm$ {\tiny 0.001}})	& 0.433 $\pm$ {\tiny 0.020} (0.442 $\pm$ {\tiny 0.020})	& 0.896 $\pm$ {\tiny 0.011} (0.887 $\pm$ {\tiny 0.003})	& 0.429 $\pm$ {\tiny 0.020} (0.439 $\pm$ {\tiny 0.020}) & 0.580 $\pm$ {\tiny 0.018} (0.587 $\pm$ {\tiny 0.018}) \\
	& Focal	& \underline{0.915 $\pm$ {\tiny 0.010}} (\underline{0.900 $\pm$ {\tiny 0.000}})	& 0.515 $\pm$ {\tiny 0.015} (0.533 $\pm$ {\tiny 0.009})	& \underline{0.903 $\pm$ {\tiny 0.011}} (0.887 $\pm$ {\tiny 0.003})	& \textcolor{red}{0.511 $\pm$ {\tiny 0.015}} (0.528 $\pm$ {\tiny 0.009}) & 0.653 $\pm$ {\tiny 0.010} (0.662 $\pm$ {\tiny 0.007}) \\
	& LDAM	& 0.770 $\pm$ {\tiny 0.008} (\underline{0.900 $\pm$ {\tiny 0.000}})	& 0.519 $\pm$ {\tiny 0.101} (0.441 $\pm$ {\tiny 0.081})	& 0.757 $\pm$ {\tiny 0.010} (0.888 $\pm$ {\tiny 0.003})	& 0.521 $\pm$ {\tiny 0.095} (0.437 $\pm$ {\tiny 0.077}) & 0.612 $\pm$ {\tiny 0.054} (0.582 $\pm$ {\tiny 0.060}) \\
	& TERM	& \underline{0.955 $\pm$ {\tiny 0.026}} (\underline{0.901 $\pm$ {\tiny 0.001}})	& 0.358 $\pm$ {\tiny 0.036} (0.402 $\pm$ {\tiny 0.017})	& \underline{0.947 $\pm$ {\tiny 0.030}} (0.889 $\pm$ {\tiny 0.003})	& 0.355 $\pm$ {\tiny 0.036} (0.398 $\pm$ {\tiny 0.017}) & 0.515 $\pm$ {\tiny 0.035} (0.549 $\pm$ {\tiny 0.016}) \\
	& ROS	& \underline{0.915 $\pm$ {\tiny 0.023}} (\underline{0.900 $\pm$ {\tiny 0.000}})	& 0.427 $\pm$ {\tiny 0.036} (0.439 $\pm$ {\tiny 0.021})	& \underline{0.902 $\pm$ {\tiny 0.023}} (0.886 $\pm$ {\tiny 0.002})	& 0.424 $\pm$ {\tiny 0.036} (0.436 $\pm$ {\tiny 0.021}) & 0.576 $\pm$ {\tiny 0.029} (0.584 $\pm$ {\tiny 0.019}) \\
	& RUS	& \underline{0.916 $\pm$ {\tiny 0.013}} (\underline{0.901 $\pm$ {\tiny 0.001}})	& 0.455 $\pm$ {\tiny 0.088} (0.467 $\pm$ {\tiny 0.083})	& \underline{0.902 $\pm$ {\tiny 0.014}} (0.886 $\pm$ {\tiny 0.003})	& 0.449 $\pm$ {\tiny 0.084} (0.463 $\pm$ {\tiny 0.080}) & 0.595 $\pm$ {\tiny 0.064} (0.605 $\pm$ {\tiny 0.060}) \\
	& DMO+TFCO	& \underline{0.940 $\pm$ {\tiny 0.027}} (\underline{0.900 $\pm$ {\tiny 0.000}})	& 0.380 $\pm$ {\tiny 0.034} (0.407 $\pm$ {\tiny 0.018})	& \underline{0.936 $\pm$ {\tiny 0.028}} (0.893 $\pm$ {\tiny 0.005})	& 0.378 $\pm$ {\tiny 0.034} (0.404 $\pm$ {\tiny 0.019}) & 0.537 $\pm$ {\tiny 0.030} (0.556 $\pm$ {\tiny 0.018}) \\
\rowcolor{gray}
\multirow{-8}{*}{\cellcolor{white}ADE-v2}	& DMO+ERO	& \underline{0.903 $\pm$ {\tiny 0.004}} (\underline{0.900 $\pm$ {\tiny 0.000}})	& \textcolor{red}{0.697 $\pm$ {\tiny 0.010}} (\textcolor{blue}{0.699 $\pm$ {\tiny 0.005}})	& 0.886 $\pm$ {\tiny 0.004} (0.884 $\pm$ {\tiny 0.002})	& 0.684 $\pm$ {\tiny 0.008} (0.684 $\pm$ {\tiny 0.004}) & \textcolor{red}{0.772 $\pm$ {\tiny 0.004}} (\textcolor{blue}{0.771 $\pm$ {\tiny 0.003}}) \\

\hline

	& WCE	& \underline{0.922 $\pm$ {\tiny 0.009}} (\underline{0.901 $\pm$ {\tiny 0.001}})	& 0.080 $\pm$ {\tiny 0.002} (0.084 $\pm$ {\tiny 0.002})	& \underline{0.907 $\pm$ {\tiny 0.012}} (0.880 $\pm$ {\tiny 0.005})	& 0.541 $\pm$ {\tiny 0.005} (0.554 $\pm$ {\tiny 0.005}) & 0.678 $\pm$ {\tiny 0.003} (0.680 $\pm$ {\tiny 0.003}) \\
	& Focal	& \underline{0.974 $\pm$ {\tiny 0.037}} (\underline{0.930 $\pm$ {\tiny 0.042}})	& 0.066 $\pm$ {\tiny 0.003} (0.068 $\pm$ {\tiny 0.003})	& \underline{0.976 $\pm$ {\tiny 0.030}} (\underline{0.918 $\pm$ {\tiny 0.051}})	& 0.487 $\pm$ {\tiny 0.013} (\textcolor{blue}{0.493 $\pm$ {\tiny 0.016}}) & 0.650 $\pm$ {\tiny 0.005} (0.641 $\pm$ {\tiny 0.016}) \\
	& LDAM	& 0.737 $\pm$ {\tiny 0.153} (\underline{0.901 $\pm$ {\tiny 0.001}})	& 0.096 $\pm$ {\tiny 0.021} (0.072 $\pm$ {\tiny 0.001})	& 0.706 $\pm$ {\tiny 0.167} (0.887 $\pm$ {\tiny 0.005})	& 0.570 $\pm$ {\tiny 0.046} (0.511 $\pm$ {\tiny 0.005}) & 0.613 $\pm$ {\tiny 0.067} (0.649 $\pm$ {\tiny 0.005}) \\
	& TERM	& \underline{0.963 $\pm$ {\tiny 0.025}} (\underline{0.902 $\pm$ {\tiny 0.002}})	& 0.068 $\pm$ {\tiny 0.003} (0.074 $\pm$ {\tiny 0.001})	& \underline{0.955 $\pm$ {\tiny 0.031}} (0.892 $\pm$ {\tiny 0.010})	& 0.493 $\pm$ {\tiny 0.011} (0.517 $\pm$ {\tiny 0.006}) & 0.649 $\pm$ {\tiny 0.003} (0.654 $\pm$ {\tiny 0.005}) \\
	& ROS	& \underline{0.913 $\pm$ {\tiny 0.010}} (\underline{0.901 $\pm$ {\tiny 0.000}})	& 0.081 $\pm$ {\tiny 0.005} (0.084 $\pm$ {\tiny 0.004})	& 0.898 $\pm$ {\tiny 0.017} (0.883 $\pm$ {\tiny 0.009})	& 0.547 $\pm$ {\tiny 0.019} (0.555 $\pm$ {\tiny 0.017}) & 0.679 $\pm$ {\tiny 0.010} (0.681 $\pm$ {\tiny 0.011}) \\
	& RUS	& \underline{0.920 $\pm$ {\tiny 0.008}} (\underline{0.901 $\pm$ {\tiny 0.001}})	& 0.080 $\pm$ {\tiny 0.003} (0.084 $\pm$ {\tiny 0.003})	& \underline{0.908 $\pm$ {\tiny 0.008}} (0.886 $\pm$ {\tiny 0.007})	& \textcolor{red}{0.542 $\pm$ {\tiny 0.011}} (0.556 $\pm$ {\tiny 0.012}) & 0.679 $\pm$ {\tiny 0.007} (0.683 $\pm$ {\tiny 0.008}) \\
	& DMO+TFCO	& 0.693 $\pm$ {\tiny 0.430} (\underline{0.902 $\pm$ {\tiny 0.001}})	& 0.127 $\pm$ {\tiny 0.112} (0.070 $\pm$ {\tiny 0.002})	& 0.690 $\pm$ {\tiny 0.422} (0.891 $\pm$ {\tiny 0.011})	& 0.611 $\pm$ {\tiny 0.198} (0.499 $\pm$ {\tiny 0.005}) & 0.475 $\pm$ {\tiny 0.270} (0.640 $\pm$ {\tiny 0.004}) \\
\rowcolor{gray}
\multirow{-8}{*}{\cellcolor{white}MURA}	& DMO+ERO	& \underline{0.900 $\pm$ {\tiny 0.001}} (\underline{0.901 $\pm$ {\tiny 0.001}})	& \textcolor{red}{0.152 $\pm$ {\tiny 0.054}} (\textcolor{blue}{0.148 $\pm$ {\tiny 0.054}})	& 0.753 $\pm$ {\tiny 0.066} (0.766 $\pm$ {\tiny 0.077})	& 0.651 $\pm$ {\tiny 0.090} (0.646 $\pm$ {\tiny 0.091}) & \textcolor{red}{0.689 $\pm$ {\tiny 0.029}} (\textcolor{blue}{0.691 $\pm$ {\tiny 0.028}}) \\

\hline
\Xhline{3\arrayrulewidth}
\end{tabular}
}
\label{table:FROP}
\vspace{-1.5em}
\end{table*}

\subsubsection{Competing methods}
For optimization of the three DMO problems, we compare our ERO method with \textbf{TensorFlow Constrained Optimization (TFCO)}\footnote{\url{https://github.com/google-research/tensorflow_constrained_optimization}}~\cite{cotter2019two}, the only open-source solver specifically designed for constrained DMO, using a smoothing-based Lagrangian framework---dubbed as \textbf{DMO+TFCO}. For competing binary IC methods, we also include loss-based methods such as \textbf{Weighted Cross-Entropy} (WCE), \textbf{Focal Loss} (Focal)~\cite{lin2017focal}, \textbf{Label-Distribution-Aware Margin Loss} (LDAM)~\cite{cao2019learning}, \textbf{Tilted Empirical Risk Minimization} (TERM)~\cite{li2020tilted}, and \textbf{Sigmoid F1 Loss} (SigmoidF1)~\cite{benedict2021sigmoidf1}, which directly optimizes the F1 score, as well as sampling-based methods including \textbf{Random Oversampling} (ROS) and \textbf{Random Undersampling} (RUS).


\subsubsection{Implementation details} 
\textbf{(1) Classifiers}. For tabular datasets, we use a 3-layer multi-layer perception (MLP) as our predictive model and solve the subproblem in \cref{alg:alm} using the ADAM optimizer (implemented as a deterministic optimizer) with learning rates $10^{-4}$ and $0.1$ for $\mb \theta$ and $s$, respectively. For image and text data, we use pretrained vision foundation model DINO v2~\cite{oquab2023dinov2} and NLP foundation model BERT~\cite{devlin2018bert} respectively for feature extraction and then train a linear model from scratch on these extracted features with a learning rate of $10^{-3}$. We set the decision threshold as $t=0.5$ directly without performing optimization, as we can equivalently adjust the learnable bias term in the last layer of our MLP models or the linear model. 
\textbf{(2) Hyperparameters for competing methods}. For all competing methods, we adopt the hyperparameter ranges recommended in the original papers and perform a grid search to select the best configuration. We take the best model during training for evaluation and report the mean and standard deviation over $10$ random trials. More details are given in \cref{sec:imp-details}. \textbf{(3) Target precision/recall levels}. Unless otherwise specified, all main experiments use a target metric level of $\alpha = 0.9$, corresponding to FPOR (precision $\geq 0.9$) and FROP (recall $\geq 0.9$); we investigate the effect of varying $\alpha$ in \cref{para:ab_exact}. 

\subsubsection{Evaluation metrics} Our evaluation focuses on two aspects: (1) \textbf{Optimization:} how well the optimization problem is solved during training, in terms of feasibility and optimality of the solution found; and (2) \textbf{Generalization:} how well the trained model performs on a held-out test set. Since after training the decision threshold $t$ can be adjusted to potentially make an infeasible solution feasible (for FPOR \& FROP) and/or optimize the objective (for all DMO problems), we also report the model performance after threshold adjustment (TA) on the training set: For FPOR and FROP,  $t$ is chosen to make the solution feasible while achieving the best objective value; For OFOS, $t$ is chosen to maximize the $F_1$ objective.  

\subsection{Main results}

\subsubsection{Results on FPOR and FROP}
\cref{table:FPOR,table:FROP} report the results on FPOR (precision $\geq 0.9$) and FROP (recall $\geq 0.9$), in which feasible solutions ($\text{precision} \ge 0.9$) are \underline{underlined}, and among them, the best objectives are shown in \textcolor{red}{red} (before TA) and \textcolor{blue}{blue} (after TA). For test, due to the difficulty in comparing different precision-recall tradeoffs, we also report the $F_1$ score as a proxy metric and highlight the best $F_1$'s. All underlines and highlights are up to $0.001$ slackness. 

\textbf{Optimization performance.} \textbf{(1) Our DMO+ERO} consistently outperforms all competing methods over all training tasks, before and after TA, returning feasible points that achieve the highest objective values among those returned by any method. We believe the excellent performance stems from our exact reformulations of the metric constraints and judicious choice of the numerical methods to solve the constrained problems; \textbf{(2) Non-DMO baselines} In contrast, without explicit metric controls, non-DMO baselines (WCE, Focal, LDAM, TERM, ROS, RUS) all fail to consistently produce feasible solutions before TA, exhibiting an striking asymmetry between the two problems: on FPOR, only $6$ of $24$ (method, dataset) combinations are feasible before TA ($25\%$) (\emph{none} achieves feasibility on \textit{MURA}), with precision gaps ranging from $0.030$ (LDAM) to $0.411$ (ROS); on FROP, by contrast, $19$ of $24$ combinations are feasible before TA ($79\%$). This asymmetry is intuitive: without an explicit constraint, loss-based and resampling-based methods tend to bias predictions toward the minority (positive) class, easily inflating recall but providing not much improvement on precision; \textbf{(3) DMO+TFCO} fares even worse on FPOR, remaining infeasible on all $4$ FPOR tasks even after TA, with post-TA violations ranging from $0.097$ (\textit{Eyepacs}) to $0.487$ (\textit{MURA}); on FROP it is feasible on $3$ of $4$ tasks before TA and $4$ of $4$ after. We suspect that DMO+TFCO's struggle with feasibility is intrinsic to the Lagrangian methods TFCO uses, which hardly guarantee feasibility for general constrained nonconvex problems. 

\textbf{Generalization performance.}  \textbf{(1) Overall trend}. We find that no single method can reliably ensure feasibility on the test set, but our DMO+ERO still remains the most competitive overall. DMO+ERO yields feasible solutions with the highest objective value on $2$ tasks (FPOR on \textit{MURA} and FROP on \textit{Eyepacs}). Among the remaining $6$ cases: (i) $2$ get no feasible solution from any method (FPOR on \textit{creditcard} and \textit{Eyepacs}), and ERO achieves the smallest feasibility violation; (ii) for the other $4$, some baselines achieve feasibility, but typically with much lower objectives than DMO+ERO. For example, on FPOR fo $\textit{ADE-v2}$, Focal is feasible but the objective is only $0.181$, far below DMO+ERO's $0.769$ with $0.024$ on feasibility violation. We note that DMO+ERO itself does not generalize well on FROP for \textit{creditcard} and \textit{MURA}, potentially for two factors: (i) errors due to finite-sample approximations to population metrics; and (ii) train-test distribution shifts (e.g., imbalance ratio change in MURA).  \textbf{(2) Strategies for improving test-time feasibility}. Previous analysis suggests natural strategies for improving test-time feasibility: (i) \emph{Imposing stricter constraints during training}. The training time constraint can be tightened to absorb the approximation errors, e.g., in FPOR targeting a population-level precision $0.9$, training with a higher sample-level precision, say $0.95$, as the constraint; and (ii) \textit{Calibrating the decision threshold using a validation set} (see \cref{app:ta-validation}): our current post-training TA is performed with respect to the training set as a way to improve optimization performance, but an independent validation set drawn from the test distribution could instead be used to calibrate the threshold for test-time feasibility. \textbf{(3) Precision-recall tradeoff measured by $F_1$}.  Moreover, different methods may have stricken different precision-recall, i.e., objective-constraint, tradeoffs, e.g., ``over''-feasible solutions often come at the price of objectives, and there are cases (FPOR on \textit{creditcard} and FPOR on \textit{Eyepacs}) where none of the methods finds a feasible solution. To quantitatively capture all these aspects, we use the $F_1$ score as a proxy metric. On this, DMO+ERO outperforms all competing methods on all tasks, suggesting that DMO+ERO finds the optimal tradeoffs in general.

\begin{table}[!htbp]
\def\arraystretch{1.1}
\centering
\caption{Comparison results on \textcolor{red}{OFOS}. Values in (parentheses) are results after TA.}
\vspace{-0.5em}
\resizebox{0.48\textwidth}{!}{
\begin{tabular}{ll||c||c}
\Xhline{2\arrayrulewidth}
& &\textbf{train} & \textbf{test} \\
dataset & method & $F_1$-score & $F_1$-score \\
\hline

	& WCE	& 0.293 $\pm$ {\tiny 0.140} (0.835 $\pm$ {\tiny 0.004})	& 0.295 $\pm$ {\tiny 0.128} (0.772 $\pm$ {\tiny 0.010}) \\
	& Focal	& 0.810 $\pm$ {\tiny 0.008} (0.821 $\pm$ {\tiny 0.003})	& 0.727 $\pm$ {\tiny 0.013} (0.740 $\pm$ {\tiny 0.016}) \\
	& LDAM	& 0.091 $\pm$ {\tiny 0.258} (0.229 $\pm$ {\tiny 0.319})	& 0.085 $\pm$ {\tiny 0.240} (0.211 $\pm$ {\tiny 0.290}) \\
	& TERM	& 0.846 $\pm$ {\tiny 0.012} (0.848 $\pm$ {\tiny 0.012})	& \textcolor{red}{0.792 $\pm$ {\tiny 0.012}} (\textcolor{blue}{0.790 $\pm$ {\tiny 0.010}}) \\
	& SigmoidF1	& 0.506 $\pm$ {\tiny 0.190} (0.832 $\pm$ {\tiny 0.010})	& 0.487 $\pm$ {\tiny 0.176} (0.762 $\pm$ {\tiny 0.016}) \\
	& ROS	& 0.507 $\pm$ {\tiny 0.093} (0.657 $\pm$ {\tiny 0.051})	& 0.508 $\pm$ {\tiny 0.095} (0.623 $\pm$ {\tiny 0.054}) \\
	& RUS	& 0.497 $\pm$ {\tiny 0.153} (0.648 $\pm$ {\tiny 0.084})	& 0.497 $\pm$ {\tiny 0.140} (0.621 $\pm$ {\tiny 0.063}) \\
	& DMO+TFCO	& 0.824 $\pm$ {\tiny 0.006} (0.830 $\pm$ {\tiny 0.002})	& 0.771 $\pm$ {\tiny 0.008} (0.762 $\pm$ {\tiny 0.012}) \\
\rowcolor{gray}
\multirow{-9}{*}{\cellcolor{white}creditcard}	& DMO+ERO	& \textcolor{red}{1.000 $\pm$ {\tiny 0.001}} (\textcolor{blue}{1.000 $\pm$ {\tiny 0.001}})	& 0.770 $\pm$ {\tiny 0.016} (0.768 $\pm$ {\tiny 0.016}) \\

\hline

	& WCE	& 0.616 $\pm$ {\tiny 0.005} (0.620 $\pm$ {\tiny 0.002})	& \textcolor{red}{0.588 $\pm$ {\tiny 0.002}} (\textcolor{blue}{0.588 $\pm$ {\tiny 0.002}}) \\
	& Focal	& 0.519 $\pm$ {\tiny 0.005} (0.521 $\pm$ {\tiny 0.005})	& 0.511 $\pm$ {\tiny 0.004} (0.510 $\pm$ {\tiny 0.004}) \\
	& LDAM	& 0.482 $\pm$ {\tiny 0.004} (0.488 $\pm$ {\tiny 0.002})	& 0.482 $\pm$ {\tiny 0.002} (0.481 $\pm$ {\tiny 0.002}) \\
	& TERM	& 0.493 $\pm$ {\tiny 0.002} (0.496 $\pm$ {\tiny 0.002})	& 0.488 $\pm$ {\tiny 0.002} (0.488 $\pm$ {\tiny 0.002}) \\
	& SigmoidF1	& 0.419 $\pm$ {\tiny 0.000} (0.419 $\pm$ {\tiny 0.000})	& 0.415 $\pm$ {\tiny 0.000} (0.415 $\pm$ {\tiny 0.000}) \\
	& ROS	& 0.616 $\pm$ {\tiny 0.008} (0.618 $\pm$ {\tiny 0.006})	& 0.585 $\pm$ {\tiny 0.005} (0.585 $\pm$ {\tiny 0.005}) \\
	& RUS	& 0.616 $\pm$ {\tiny 0.005} (0.618 $\pm$ {\tiny 0.002})	& 0.584 $\pm$ {\tiny 0.002} (0.584 $\pm$ {\tiny 0.002}) \\
	& DMO+TFCO	& 0.608 $\pm$ {\tiny 0.002} (0.609 $\pm$ {\tiny 0.002})	& 0.576 $\pm$ {\tiny 0.002} (0.576 $\pm$ {\tiny 0.003}) \\
\rowcolor{gray}
\multirow{-9}{*}{\cellcolor{white}Eyepacs}	& DMO+ERO	& \textcolor{red}{0.647 $\pm$ {\tiny 0.002}} (\textcolor{blue}{0.651 $\pm$ {\tiny 0.001}})	& 0.586 $\pm$ {\tiny 0.004} (0.585 $\pm$ {\tiny 0.004}) \\

\hline

	& WCE	& 0.804 $\pm$ {\tiny 0.004} (0.810 $\pm$ {\tiny 0.004})	& 0.790 $\pm$ {\tiny 0.004} (0.797 $\pm$ {\tiny 0.002}) \\
	& Focal	& 0.790 $\pm$ {\tiny 0.018} (0.796 $\pm$ {\tiny 0.019})	& 0.775 $\pm$ {\tiny 0.018} (0.783 $\pm$ {\tiny 0.020}) \\
	& LDAM	& 0.811 $\pm$ {\tiny 0.005} (0.816 $\pm$ {\tiny 0.003})	& 0.793 $\pm$ {\tiny 0.004} (0.799 $\pm$ {\tiny 0.002}) \\
	& TERM	& 0.799 $\pm$ {\tiny 0.009} (0.802 $\pm$ {\tiny 0.008})	& 0.785 $\pm$ {\tiny 0.009} (0.789 $\pm$ {\tiny 0.010}) \\
	& SigmoidF1	& 0.553 $\pm$ {\tiny 0.138} (0.564 $\pm$ {\tiny 0.142})	& 0.548 $\pm$ {\tiny 0.134} (0.558 $\pm$ {\tiny 0.139}) \\
	& ROS	& 0.802 $\pm$ {\tiny 0.005} (0.809 $\pm$ {\tiny 0.004})	& 0.788 $\pm$ {\tiny 0.005} (0.795 $\pm$ {\tiny 0.004}) \\
	& RUS	& 0.798 $\pm$ {\tiny 0.004} (0.805 $\pm$ {\tiny 0.005})	& 0.783 $\pm$ {\tiny 0.004} (0.791 $\pm$ {\tiny 0.006}) \\
	& DMO+TFCO	& 0.800 $\pm$ {\tiny 0.004} (0.804 $\pm$ {\tiny 0.003})	& 0.785 $\pm$ {\tiny 0.004} (0.787 $\pm$ {\tiny 0.004}) \\
\rowcolor{gray}
\multirow{-9}{*}{\cellcolor{white}ADE-v2}	& DMO+ERO	& \textcolor{red}{0.848 $\pm$ {\tiny 0.003}} (\textcolor{blue}{0.851 $\pm$ {\tiny 0.002}})	& \textcolor{red}{0.820 $\pm$ {\tiny 0.005}} (\textcolor{blue}{0.823 $\pm$ {\tiny 0.005}}) \\

\hline

	& WCE	& 0.353 $\pm$ {\tiny 0.006} (0.366 $\pm$ {\tiny 0.004})	& 0.526 $\pm$ {\tiny 0.017} (0.483 $\pm$ {\tiny 0.011}) \\
	& Focal	& 0.373 $\pm$ {\tiny 0.010} (0.376 $\pm$ {\tiny 0.009})	& 0.466 $\pm$ {\tiny 0.017} (0.475 $\pm$ {\tiny 0.021}) \\
	& LDAM	& 0.342 $\pm$ {\tiny 0.008} (0.349 $\pm$ {\tiny 0.008})	& 0.423 $\pm$ {\tiny 0.017} (0.441 $\pm$ {\tiny 0.008}) \\
	& TERM	& 0.426 $\pm$ {\tiny 0.054} (0.444 $\pm$ {\tiny 0.063})	& 0.483 $\pm$ {\tiny 0.035} (0.544 $\pm$ {\tiny 0.050}) \\
	& SigmoidF1	& 0.449 $\pm$ {\tiny 0.006} (0.491 $\pm$ {\tiny 0.005})	& \textcolor{red}{0.701 $\pm$ {\tiny 0.001}} (\textcolor{blue}{0.663 $\pm$ {\tiny 0.002}}) \\
	& ROS	& 0.349 $\pm$ {\tiny 0.005} (0.359 $\pm$ {\tiny 0.007})	& 0.518 $\pm$ {\tiny 0.020} (0.474 $\pm$ {\tiny 0.016}) \\
	& RUS	& 0.348 $\pm$ {\tiny 0.010} (0.359 $\pm$ {\tiny 0.010})	& 0.520 $\pm$ {\tiny 0.029} (0.487 $\pm$ {\tiny 0.022}) \\
	& DMO+TFCO	& 0.534 $\pm$ {\tiny 0.004} (0.536 $\pm$ {\tiny 0.004})	& 0.584 $\pm$ {\tiny 0.009} (0.581 $\pm$ {\tiny 0.009}) \\
\rowcolor{gray}
\multirow{-9}{*}{\cellcolor{white}MURA}	& DMO+ERO	& \textcolor{red}{0.569 $\pm$ {\tiny 0.012}} (\textcolor{blue}{0.578 $\pm$ {\tiny 0.013}})	& 0.563 $\pm$ {\tiny 0.019} (0.570 $\pm$ {\tiny 0.017}) \\

\hline
\Xhline{3\arrayrulewidth}
\end{tabular}
}
\label{table:OFBS}
\end{table}

\subsubsection{Results on OFOS}
\cref{table:OFBS} summarizes the results on OFOS, the only problem studied here that optimizes a single target metric ($F_1$) without other metric constraints. \textbf{(1) Optimization performance}. For training (i.e., optimization), DMO+ERO consistently outperforms all competing methods, often with considerable margins (e.g., gaps of $0.024$ on \textit{Eyepacs} and $0.084$ on \textit{ADE-v2} with respect to the second-best). All non-DMO baselines exhibit large performance variations across datasets relative to DMO+ERO---not surprising as they do not target the $F_1$ metric directly. DMO+TFCO is often no better than non-DMO baselines, underscoring the problem with metric smoothing; 
\textbf{(2) Generalization performance}. At test time, DMO+ERO is the best on $1$ (\textit{Eyepacs}) out of the $4$ tasks, and it is always among the top $3$ on all tasks. In contrast, most competing methods have highly unstable performance. For example, SigmoidF1 performs the best on \textit{MURI}, but the worst on \textit{Eyepacs} \& \textit{ADE-v2}; TERM performs the best on \textit{creditcard}, but among the worst three on \textit{Eyepacs}; only WCE has never entered the bottom three (by after-TA $F_1$). This highlights our DMO+ERO's strong test performance. We also note, similar to the FPOR and FROP results above, the generalization gaps for all methods could be reducible by TA with respect to a validation set. 

In \cref{app:further-ablation}, we perform further analysis and ablation study to (1) stress test all methods under more stringent constraints on FPOR \& FROP (\cref{para:ab_alpha_sweep}); (2) understand the performance benefits of three key algorithmic components in our ERO (\cref{para:ab_exact}); and (3) explore alternative post-training threshold finetuning to improve generalization (\cref{app:ta-validation}).    

In sum, our ERO method, combining a novel exact reformulation of the indicator function and an exact penalty method to promote feasibility, is a clear winner in optimization performance for all three DMO problems. Its generalization performance is reasonably good, but improvable via simple strategies. 

\section{Conclusion}\label{sec:conclusion}

In this paper, we introduce a novel \emph{exact} reformulation and optimization (ERO) framework for three (constrained) direct metric optimization (DMO) problems on binary imbalanced classification: fix-precision-optimize-recall (FPOR), fix-recall-optimize-precision (FROP), and optimize-$F_1$-score (OFOS). Our framework is \emph{the first of its kind}, as dominant ideas on DMO in the literature use smooth approximations to replace the indicator function inside these metrics, and hence suffer from such approximation errors. We establish the equivalence of our reformulations to the original DMO problems under mild assumptions, and demonstrate the effectiveness of our ERO framework through experiments on four tasks spanning vision, text and structured datasets.  

Our current work has multiple limitations that warrant future research: \textbf{(1) Extending ERO to cover more DMO problems}. Although we have only dealt with the three metrics, i.e., precision, recall, $F_1$ scores, for binary classification, the ERO technique seems applicable to numerous other metrics in binary classification and information retrieval, e.g., accuracy, balanced accuracy, average precision, precision@k, recall@k, NDCG; see our general results in \cref{thm:general-dmo}. Moreover, since most metrics used in numerous other learning settings, such as multiclass/multilabel classification, selective classification~\cite{liangselective}, conformal prediction~\cite{xie2024boosted}, autolabeling~\cite{vishwakarma2024pearls}, watermark detection~\cite{liang2025baseline}, object detection \& image segmentation, are natural extensions of those used for binary classification, it is likely that we can generalize the ERO technique to these metrics as well; \textbf{(2) Developing stochastic optimization methods for constrained problems}. Typical metrics involve nonlinear composition of finite-sum functions---with number of summands proportional to the dataset size (e.g., precision and average precision), and our reformulation trick induces numerous constraints---number of them scales with the dataset size again. So, our current deterministic exact penalty method cannot scale to large-scale datasets, although it seems plausible and promising to develop stochastic optimization methods to solve the unconstrained subproblem thereof. Overall, the development of scalable stochastic optimization methods to solve constrained optimization problems with stochastic functions and numerous constraints appears to be a nascent area in numerical optimization and machine learning~\cite{liang2022ncvx,liang2023optimization,he2024federated,alacaoglu2024complexity,lu2025first,lu2024variance,cui2025two}; \textbf{(3) Understanding optimization and generalization for constrained deep learning problems}. Overparameterization and algorithmic implicit regularization are known to be critical to the surprisingly favorable optimization and generalization properties associated with first-order methods in unconstrained deep learning~\cite{belkin2021fit,bartlett2021deep}. What are the numerical methods that tend to facilitate global optimization and effective generalization for constrained deep learning problems with overparametrized models? 

\section*{Acknowledgments}
Peng L., He C., Cui Y., and Sun J. are partially supported by the NIH fund R01CA287413. Sun J. is also partially supported by NSF IIS 2435911. The content is solely the responsibility of the authors and does not necessarily represent the official views of the National Institutes of Health. Peng L. and Sun J. are also partially supported by CISCO Research fund 1085646 PO USA000EP390223. This research is part of AI-CLIMATE: ``AI Institute for Climate-Land Interactions, Mitigation, Adaptation, Tradeoffs and Economy,'' and is supported by USDA National Institute of Food and Agriculture (NIFA) and the National Science Foundation (NSF) National AI Research Institutes Competitive Award no. 2023-67021-39829. The authors acknowledge the Minnesota Supercomputing Institute (MSI) at the University of Minnesota for providing resources that contributed to the research results reported in this article.

\bibliographystyle{IEEEtran}
\bibliography{references}

\begin{IEEEbiographynophoto}{Le Peng}
received the Ph.D. degree in Computer Science and Engineering from the University of Minnesota–Twin Cities in 2025. 
He is currently a Research Scientist at Meta. 
His research interests include machine learning, computer vision, and AI for healthcare.
\end{IEEEbiographynophoto}

\vspace{-0.5em}

\begin{IEEEbiographynophoto}{Yash Travadi}
 is a Ph.D. student in the School of Statistics at the University of Minnesota, Twin Cities. He received a B.S.-M.S. dual degree in Mathematics and Scientific Computing from the Indian Institute of Technology, Kanpur, in 2018. His research interests lie in machine learning, optimization, and healthcare.
\end{IEEEbiographynophoto} 

\vspace{-0.5em}

\begin{IEEEbiographynophoto}{Chuan He} is an Assistant Professor in the Department of Mathematics and is affiliated with WASP at Linköping University, Sweden. His research interests center around data science and optimization. His research has been supported by the Wallenberg AI, Autonomous Systems and Software Program (WASP) funded by the Knut and Alice Wallenberg Foundation.
\end{IEEEbiographynophoto}

\vspace{-0.5em}

\begin{IEEEbiographynophoto}{Ying Cui}
is an assistant professor in Industrial Engineering and Operations Research at the University of California, Berkeley. Previously, she was an assistant professor at the University of Minnesota and a postdoctoral research associate in the Daniel J. Epstein Department of Industrial and Systems Engineering at the University of Southern California. She received the Ph.D. degree in Mathematics from the National University of Singapore. Her research focuses on optimization and the mathematical foundations of data science, developing efficient algorithms for large-scale nonlinear problems through nonsmooth analysis. She is a co-author of the monograph ''Modern Nonconvex Nondifferentiable Optimization''.
\end{IEEEbiographynophoto} 

\vspace{-0.5em}

\begin{IEEEbiographynophoto}{Ju Sun (IEEE Member)}
  is an McKnight Land-Grant professor in the Department of Computer Science and Engineering of University of Minnesota. His research interests include machine learning, computer vision, numerical optimization, and AI for science and medicine. He was a postdoc researcher at Stanford University (2016--2019) and PhD student at Columbia University (2011--2016).
\end{IEEEbiographynophoto} 


\onecolumn
\appendices

\section{Proofs of auxiliary lemmas}

\subsection{Proof of \cref{lemma:indic-equiv}}
\label{sec:proof-indic-equiv}
\begin{proof}
First, we have $[s+a-t]_+ - [s+a-1-t]_+ = \min\paren{1, [s+a-t]_+} \in [0, 1]$. When $a \ne t$, we have 

\textbf{The $\Longrightarrow$ direction}: When $a> t$, $s=1$. It is easy to see that $s -  \min\paren{1, [s+a-t]_+}= 1 - 1 = 0$. When $a < t$, $s = 0$, so $s -  \min\paren{1, [s+a-t]_+} = 0 -0 =0$. 

\textbf{The $\Longleftarrow$ direction}: $s -  \min\paren{1, [s+a-t]_+} = 0 \Longrightarrow s =  \min\paren{1, [s+a-t]_+} \in [0, 1]$. When $a > t$, $s = 1$ as $s = [s+a-t]_+ =  s + a - t$ is not possible. So, in this case, $s - \mb{1}\{a>t\} = 1 - 1 = 0$. Similarly, when $a < t$, $s = 0$ as $\min\paren{1, [s+a-t]_+}  = [s+a-t]_+$ and $s = s + a - t$ is not possible. So, in this case, $s - \mb{1}\{a>t\} = 0 - 0 = 0$. 

From the proof, clearly $s \in \set{0, 1}$ always, completing the proof. 
\end{proof}

\subsection{Proof of \cref{lemma:relax-key}}
\label{sec:proof-relax-key}
\begin{proof}
    First, we have $[s+a-t]_+ - [s+a-1-t]_+ = \min\paren{1, [s+a-t]_+} \in [0, 1]$. Also, recall that we assume that $a \ne t$ and $s \in [0, 1]$. Now, 

    \textbf{The $\Longrightarrow$ direction}:  
    \begin{itemize}
        \item When $s \le  \min\paren{1, [s+a-t]_+} \le 1$, \textbf{(i) if $a < t$}, $\min\paren{1, [s+a-t]_+} = [s+a-t]_+ = 0$, as if it were $s+a-t$ we would obtain $s \le s + a -t$, not possible for $a < t$. So, $s \le \mb 1 \set{a > t}$ in this case; \textbf{(ii) if $a > t$}, we have $s \le \mb 1 \set{a > t} = 1$ trivially. 

        \item Similarly, when $s \ge  \min\paren{1, [s+a-t]_+} \ge 0$, \textbf{(i) if $a < t$}, $s \ge \mb 1 \set{a > t} = 0 $ trivially;  \textbf{(i) if $a > t$}, $\min\paren{1, [s+a-t]_+} = 1$, as if it were $ [s+a-t]_+ = s+a - t$ we would obtain $s \ge s + a - t$, not possible for $a > t$. So, $s \ge \mb 1 \set{a > t}$ in this case. 
    \end{itemize}
    
   \textbf{The $\Longleftarrow$ direction: } 
    \begin{itemize}
        \item When $s \le \mb 1\set{a > t}$, \textbf{(i) if $a < t$}, $s = 0$. It is easy to check that $[a - 1 -t]_+ - [a-t]_+ = 0 \le 0$; \textbf{(ii) if $a > t$}, it is easy to check that $s \le \min\paren{1, [s+a-t]_+} = [s+a-t]_+ - [s+a-1-t]_+$. 
        \item When $s \ge \mb 1\set{a > t}$, \textbf{(i) if $a < t$}, it is easy to check that $s \ge [s+a-t]_+ = \min\paren{1, [s+a-t]_+} = [s+a-t]_+ - [s+a-1-t]_+$; \textbf{(ii) if $a > t$}, $s = 1$. It is easy to check that $1 + [a-t]_+ - [1+a-t]_+ = 1 + a-t - (1 + a - t) = 0 \ge 0$.  
    \end{itemize}
    completing the proof. 
\end{proof}

\subsection{Proof of \cref{thm:feasibility-relax}}
\begin{proof}
    Note that any point of the form $(\mb\theta, [\mb 1\set{f_{\mb \theta}(\mb x_i) > t}]_i, t)$ satisfies the constraint $s_i \le \mb 1\set{f_{\mb \theta}(\mb x_i) > t} \;\forall i\in\mc P, \; s_i \ge \mb 1\set{f_{\mb \theta}(\mb x_i) > t} \;\forall i\in\mc N$ trivially. So, 
    \begin{align}
        (\mb \theta, t) \; \text{feasible for}\; \cref{def:FPOR_expanded}  & \Longleftrightarrow \phi_2\paren{[\mb 1\set{f_{\mb \theta}(\mb x_i) > t}]_i} \ge \alpha  \\
        & \Longleftrightarrow (\mb\theta, [\mb 1\set{f_{\mb \theta}(\mb x_i) > t}]_i, t) \; \text{feasible for}\;  \cref{eq:proof-frop-key-1}, 
    \end{align}
    completing the proof. 
\end{proof}

\section{General theoretical results}\label{apx:proofs}
In this section, we treat the three DMO problems, i.e., FPOR, FROP, and OFOS, in a unified manner and consider their inequality-constrained reformulations induced by \cref{lemma:relax-key}. For convenience, define 
\begin{align}\label{def:phi}
    \phi_p(\mb{s}) \doteq \tfrac{\sum_{i\in\mc P}s_i}{\sum_{i\in\mc P\cup\mc N}s_i}, \quad 
    \phi_r(\mb{s}) = \tfrac{\sum_{i\in\mc P}s_i}{N_+},\quad 
    \phi_{F_1}(\mb{s}) = \tfrac{2\,\sum_{i\in\mc P}s_i}{N_+ + \sum_{i\in\mc P\cup\mc N}s_i}.
\end{align}
Then, the three DMO problems can be written compactly as 
\begin{align} \label{eq:app_dmo_def}
    \textbf{(FPOR)} & \quad \max\nolimits_{\mb \theta, t \in [0, 1]} \; \phi_r([\mb 1\set{f_{\mb \theta}(\mb x_i) > t}]_i) \quad \text{s.t.} \; \phi_p([\mb 1\set{f_{\mb \theta}(\mb x_i) > t}]_i) \ge \alpha, \\
    \textbf{(FROP)} & \quad \max\nolimits_{\mb \theta, t \in [0, 1]} \; \phi_p([\mb 1\set{f_{\mb \theta}(\mb x_i) > t}]_i) \quad \text{s.t.} \; \phi_r([\mb 1\set{f_{\mb \theta}(\mb x_i) > t}]_i) \ge \alpha, \\
    \textbf{(OFOS)} & \quad \max\nolimits_{\mb \theta, t \in [0, 1]} \; \phi_{F_{1}}([\mb 1\set{f_{\mb \theta}(\mb x_i) > t}]_i), 
\end{align}
respectively. Note that all three problems can be written in the form 
\begin{align} \label{def:dmo_original}
    \max\nolimits_{\mb \theta, t \in [0, 1]} \; \phi_1([\mb 1\set{f_{\mb \theta}(\mb x_i) > t}]_i) \quad \text{s.t.} \; \phi_2([\mb 1\set{f_{\mb \theta}(\mb x_i) > t}]_i) \ge \alpha, 
\end{align}
where for OFOS, we can define $\phi_2 \equiv 0$ and set $\alpha = 0$. So, below, we study \cref{def:dmo_original} to cover all three problems together. For this, we consider the following inequality-constrained reformulation induced by \cref{lemma:relax-key}: 
\begin{align}\label{def:dmo_reform_linear}
    \begin{split}
    \max\nolimits_{\mb\theta, \mb{s}\in [0, 1]^N, t \in [0, 1]} \; \phi_1(\mb s) & \quad  \text{s.t.} \; \phi_2(\mb s) \ge \alpha,  \\
    &s_i + [s_i+f_{\mb \theta}(\mb x_i) - t-1]_+ - [s_i+f_{\mb \theta}(\mb x_i) - t]_+\le0\;\forall i\in\mc P,\\
    &s_i + [s_i+f_{\mb \theta}(\mb x_i) - t-1]_+ - [s_i+f_{\mb \theta}(\mb x_i) - t]_+\ge0\;\forall i\in\mc N.
    \end{split}
\end{align}
which generalizes \cref{eq:FPOR-almost}, and its cousin that keeps the indicator function
\begin{align} \label{def:dmo_reform_indic}
    \begin{split}
    \max\nolimits_{\mb\theta, \mb{s}\in [0, 1]^N, t \in [0, 1]} \; \phi_1(\mb s) & \quad \text{s.t.} \; \phi_2(\mb s) \ge \alpha,  \\
    & \quad s_i \le \mb 1\set{f_{\mb \theta}(\mb x_i) > t} \;\forall i\in\mc P, \; s_i \ge \mb 1\set{f_{\mb \theta}(\mb x_i) > t} \;\forall i\in\mc N 
    \end{split}
\end{align} 
which generalizes \cref{eq:proof-frop-key-1}. Our development is closely parallel to that in \cref{sec:reform_ineq}. 

The following lemma is a simple generalization of \cref{thm:feasibility-relax} with an identical proof strategy---note that $\phi_1$ and $\phi_2$ in \cref{sec:reform_ineq} are more restrictive. 
\begin{lemma}[equivalence in feasibility of \cref{def:dmo_original} and of \cref{def:dmo_reform_linear}] \label{thm:feasi-strong}
    A point $(\mb\theta,t)$ is feasible for \cref{def:dmo_original} if and only if $(\mb\theta, [\mb 1\set{f_{\mb \theta}(\mb x_i) > t}]_i, t)$ is feasible for \cref{def:dmo_reform_indic}. 
\end{lemma}
\begin{proof}
        Note that any point of the form $(\mb\theta, [\mb 1\set{f_{\mb \theta}(\mb x_i) > t}]_i, t)$ satisfies the constraint $s_i \le \mb 1\set{f_{\mb \theta}(\mb x_i) > t} \;\forall i\in\mc P, \; s_i \ge \mb 1\set{f_{\mb \theta}(\mb x_i) > t} \;\forall i\in\mc N$ trivially. So, 
    \begin{align}
        (\mb \theta, t) \; \text{feasible for}\; \cref{def:FPOR_expanded}  & \Longleftrightarrow \phi_2\paren{[\mb 1\set{f_{\mb \theta}(\mb x_i) > t}]_i} \ge \alpha  \\
        & \Longleftrightarrow (\mb\theta, [\mb 1\set{f_{\mb \theta}(\mb x_i) > t}]_i, t) \; \text{feasible for}\;  \cref{def:dmo_reform_indic}, 
    \end{align}
    completing the proof. 
\end{proof}
The next theorem generalizes \cref{thm:feasibility-retrict}. 
\begin{theorem}[equivalence in feasibility of \cref{def:dmo_original} and of \cref{def:dmo_reform_linear}] \label{thm:feasi-weak}
    \begin{enumerate}[label=(\roman*)]
        \item If a non-singular point $(\mb\theta,t)$ is feasible for \cref{def:dmo_original}, $(\mb\theta, \mb s, t)$ is feasible for \cref{def:dmo_reform_linear} for a certain $\mb s$; in particular, $(\mb\theta, [\mb 1\set{f_{\mb \theta}(\mb x_i) > t}]_i, t)$ is feasible for \cref{def:dmo_reform_linear}. 
        \item If $(\mb\theta, \mb s, t)$ with non-singular $(\mb \theta, t)$ is feasible for \cref{def:dmo_reform_linear} for a certain $\mb s$, $(\mb \theta, t)$ is feasible for \cref{def:dmo_original}. 
    \end{enumerate}
\end{theorem}
\begin{proof}
    We need a couple of important facts: 
    \begin{fact}[generalization of \cref{eq:proof-frop-key-2}] \label{eq:proof-dmo-key-2}
        Both $\phi_1(\mb s)$ and $\phi_2(\mb s)$ over $\mb s \in [0, 1]^N$ are coordinate-wise monotonically nondecreasing with respect to $s_i \ \forall i \in \mc P$ and coordinate-wise monotonically nonincreasing with respect to $s_i \ \forall i \in \mc N$. 
    \end{fact}
    It can be easily verified that $\phi_r(\mb s)$, $\phi_p(\mb s)$, $\phi_{F_{1}}(\mb s)$, and constant-$0$ function are coordinate-wise monotonically nondecreasing with respect to $s_i \ \forall i \in \mc P$ and coordinate-wise monotonically nonincreasing with respect to $s_i \ \forall i \in \mc N$, implying \cref{eq:proof-dmo-key-2}. Moreover, 
    \begin{fact}[generalization of \cref{eq:proof-frop-key-3}]  \label{eq:proof-dmo-key-3}
        If a point $(\mb \theta, \mb s, t)$ is feasible for \cref{def:dmo_reform_indic}, the ``rounded'' point $(\mb \theta, [\mb 1\set{f_{\mb \theta}(\mb x_i) > t}]_i, t)$ is also feasible for \cref{def:dmo_reform_indic}. Moreover, $\phi_1([\mb 1\set{f_{\mb \theta}(\mb x_i) > t}]_i) \ge \phi_1(\mb s)$. 
    \end{fact}
    To see it, note that for any $(\mb \theta, \mb s, t)$, $(\mb \theta, [\mb 1\set{f_{\mb \theta}(\mb x_i) > t}]_i, t)$ satisfies the constraint $s_i \le \mb 1\set{f_{\mb \theta}(\mb x_i) > t} \;\forall i\in\mc P, \; s_i \ge \mb 1\set{f_{\mb \theta}(\mb x_i) > t} \;\forall i\in\mc N$ trivially, and 
    \begin{align}
        \phi_1([\mb 1\set{f_{\mb \theta}(\mb x_i) > t}]_i) \ge \phi_1(\mb s), \quad \phi_2([\mb 1\set{f_{\mb \theta}(\mb x_i) > t}]_i) \ge \phi_2(\mb s) \ge \alpha 
    \end{align}
    due to \cref{eq:proof-dmo-key-2}. 

    Next, we prove the claimed equivalence based on the two facts. 
    \begin{itemize}[leftmargin=1em]
    \item \textbf{The $\Longrightarrow$ direction:} If a non-singular point $(\mb\theta,t)$ is feasible for \cref{def:dmo_original}, $(\mb\theta$, $[\mb 1\set{f_{\mb \theta}(\mb x_i) > t}]_i$, $t)$ is feasible \cref{def:dmo_reform_indic} by \cref{thm:feasi-strong}. Due to \cref{lemma:relax-key}, $(\mb\theta, [\mb 1\set{f_{\mb \theta}(\mb x_i) > t}]_i, t)$ is also feasible for \cref{def:dmo_reform_linear}; 
    
    \item \textbf{The $\Longleftarrow$ direction:} Suppose a point $(\mb\theta, \mb s, t)$ with $(\mb \theta, t)$ non-singular is feasible for \cref{def:dmo_reform_linear}. Due to \cref{lemma:relax-key}, $(\mb\theta, \mb s, t)$ is feasible for \cref{def:dmo_reform_indic}. Now, by \cref{eq:proof-dmo-key-3}, $(\mb\theta, [\mb 1\set{f_{\mb \theta}(\mb x_i) > t}]_i, t)$ is also feasible for \cref{def:dmo_reform_indic}. Invoking \cref{thm:feasi-strong}, we conclude that $(\mb \theta, t)$ is feasible for \cref{def:dmo_original}.  
    \end{itemize}
\end{proof}
The next theorem generalizes \cref{thm:exact-fpor}. 
\begin{theorem}[equivalence in global solution of \cref{def:FPOR_expanded} and of \cref{eq:FPOR-almost}] \label{thm:global-general}
    Any non-singular $(\mb \theta^\ast, t^\ast)$ is a global solution to \cref{def:dmo_original} if and only if $(\mb \theta^\ast, \mb s^\ast, t^\ast)$ is a global solution to \cref{def:dmo_reform_linear} for a certain $\mb s^\ast$. 
\end{theorem}
\begin{proof}
    First, due to \cref{lemma:relax-key},  $(\mb \theta^\ast, \mb s^\ast, t^\ast)$ with non-singular $(\mb \theta^\ast, t^\ast)$ is a global solution to \cref{def:dmo_reform_linear} if and only if it is a global solution to \cref{def:dmo_reform_indic}. So, next we establish the connection between \cref{def:dmo_reform_indic} and \cref{def:dmo_original} in terms of global solutions. 

    Since \cref{thm:feasi-weak} already settles the equivalence in feasibility, here we only need to focus on the optimality in the objective value. Note that for any feasible $(\mb \theta, \mb s, t)$ for \cref{def:dmo_reform_indic}, $(\mb \theta, \mb 1\set{f_{\mb \theta}(\mb x_i) > t}]_i, t)$ is also feasible and $\phi_1(\mb s) \le \phi_1(\mb 1\set{f_{\mb \theta}(\mb x_i) > t}]_i)$ due to \cref{eq:proof-dmo-key-3}, implying that there exists a global solution of the form $(\mb \theta, \mb 1\set{f_{\mb \theta}(\mb x_i) > t}]_i, t)$ for \cref{def:dmo_reform_indic}. So, we have the following chain of equalities: 
        \begin{align}
            & \max\set{\phi_1(\mb s): (\mb \theta, \mb s, t) \; \text{feasible for}\; \cref{def:dmo_reform_indic}} \nonumber \\
            = \; & \max\set{\phi_1(\mb 1\set{f_{\mb \theta}(\mb x_i) > t}]_i): (\mb \theta, \mb 1\set{f_{\mb \theta}(\mb x_i) > t}]_i, t) \; \text{feasible for}\; \cref{def:dmo_reform_indic}} \\
            = \; & \max\set{\phi_1(\mb 1\set{f_{\mb \theta}(\mb x_i) > t}]_i): (\mb \theta, t) \; \text{feasible for}\; \cref{def:dmo_original}}  \quad (\text{by \cref{thm:feasi-strong}}),   
        \end{align}
        i.e., the three optimal values are equal, implying the claimed result. 
\end{proof} 
Note that throughout the above proofs, \cref{lemma:relax-key} and the coordinate-wise monotonicity of $\phi_1$ and $\phi_2$ in \cref{eq:proof-dmo-key-2} are the most crucial results we need. In fact, we have proved the following general result about DMO, beyond the three DMO problems considered in this paper. 
\begin{theorem}[Reformulation of general DMO problems] \label{thm:general-dmo}
    Consider a binary classification problem with a training set $\set{(\mb x_i, y_i)}_{i=1}^N$ over $\mc X \times \set{0, 1}$. Let $\mc P$ and $\mc N$ denote the indices for the positive (``1'') and negative (``0'') classes, respectively. 
    Consider a direct metric optimization (DMO) problem of the form 
    \begin{align} \label{def:dmo_general} 
    \max\nolimits_{\mb \theta, t \in [0, 1]} \; \phi_1([\mb 1\set{f_{\mb \theta}(\mb x_i) > t}]_i) \quad \text{s.t.} \; \phi_2([\mb 1\set{f_{\mb \theta}(\mb x_i) > t}]_i) \ge \alpha, 
    \end{align}
    where we assume the predictive model $f_{\mb \theta}: \mc X \to [0, 1]$, and the following reformulation of the DMO problem: 
    \begin{align}\label{def:general_dmo_reform_linear}
        \begin{split}
        \max\nolimits_{\mb\theta, \mb{s}\in [0, 1]^N, t \in [0, 1]} \; \phi_1(\mb s) & \quad  \text{s.t.} \; \phi_2(\mb s) \ge \alpha,  \\
        &s_i + [s_i+f_{\mb \theta}(\mb x_i) - t-1]_+ - [s_i+f_{\mb \theta}(\mb x_i) - t]_+\le0\;\forall i\in\mc P,\\
        &s_i + [s_i+f_{\mb \theta}(\mb x_i) - t-1]_+ - [s_i+f_{\mb \theta}(\mb x_i) - t]_+\ge0\;\forall i\in\mc N.
        \end{split}
    \end{align}
    If the functions $\phi_1(\mb z)$ and $\phi_2(\mb z)$ are coordinate-wise non-decreasing with respect to $z_i \ \forall i \in \mc P$ and coordinate-wise non-increasing with respect to $z_i \ \forall i \in \mc N$, the following hold: 
    \begin{enumerate}[label=(\roman*)]
        \item If $(\mb \theta, \mb s, t)$ with non-singular $(\mb \theta, t)$ (i.e., $f_{\mb \theta}(\mb x_i) \ne t\ \forall i$) is feasible for \cref{def:general_dmo_reform_linear}, $(\mb \theta, t)$ is feasible for \cref{def:dmo_general}; 
        \item If $(\mb \theta^\ast, \mb s^\ast, t^\ast)$ with non-singular $(\mb \theta^\ast, t^\ast)$ is a global solution to  \cref{def:general_dmo_reform_linear}, $(\mb \theta^\ast, t^\ast)$ is a global solution to \cref{def:dmo_general}. 
    \end{enumerate}
\end{theorem} 
We suspect that the methods and results we develop here can cover and extend to numerous other metrics commonly used in classification and information retrieval, such as accuracy, balanced accuracy, average precision, mean average precision, precision@k, recall@k, NDCG (Normalized Discounted Cumulative Gain), which we leave for future work. 

\subsection{Computational complexity}\label{subsec:complexity}
Here, we briefly discuss the time and memory complexities of the proposed exact reformulation and optimization (ERO) method, and compare them with those of smooth surrogate-based counterparts. Let $N \doteq N_+ + N_-$ denote the size of the training set and $D \doteq |\mb \theta|$ the number of parameters in the model $f_{\mb\theta}$. We write $C_{\mathrm{fwd}}(N)$ and $C_{\mathrm{bwd}}(N)$ for the cost of one full-dataset forward and backward pass through $f_{\mb\theta}$, respectively. These quantities depend on the architecture of $f_{\mb\theta}$ and typically dominate the computational cost for large neural networks.

Consider one iteration of the subproblem optimization in \cref{alg:alm} using a first-order method. ERO requires one forward pass to compute $f_{\mb\theta}(\mb x_i)$ for all samples, followed by the evaluation of the objective term $\phi_{\mathrm{obj}}(\mb s)$, the metric constraint term $\phi_{\mathrm{con}}(\mb s)$, the regularization term $\psi(\mb\theta,\mb s)$, and the reformulation constraints $\eta_i(\mb\theta,\mb s,t)$ for all $i\in[N]$. Apart from the forward and backward passes through $f_{\mb\theta}$, these operations are either separable across samples or involve simple reductions over the dataset, and therefore cost $O(N)$. The projection onto the simple constraint set $\mb s\in[0,1]^N$ and $t\in[0,1]$ also costs $O(N)$.
Thus, the cost of one inner iteration of ERO is
\begin{align}
    C_{\mathrm{fwd}}(N) +
    C_{\mathrm{bwd}}(N) +
    O(N) + O(D).
\end{align}
Here, the $O(D)$ term accounts for updating the model parameters, while the $O(N)$ term accounts for evaluating the metric and reformulation terms, updating the lifted variable $\mb s$, and performing the projection. If each subproblem is solved for $M$ iterations and the outer loop is run for $K$ penalty updates, the overall time complexity of ERO is
\begin{align}
    O\paren{KM\paren{C_{\mathrm{fwd}}(N) + C_{\mathrm{bwd}}(N) + N + D}}.
\end{align}
In terms of memory complexity for full batch gradient updates, in addition to $O(N)$ training data and $O(D)$ model parameters, ERO requires additional $O(N)$ memory for storing the values and gradients for the lifting variables $\mb s$. Hence, when compared to a smooth surrogate version of the DMO solver using a similar optimization strategy, ERO maintains the same asymptotic time and memory complexity, with a modest overhead in the constant factors of the $O(N)$ term.

\section{Additional experimental details and results}\label{sec:supp_results} 

\subsection{Dataset}\label{sec:supp_results_image_text}

This section provides details about the four datasets used in our experiment. An overview of these datasets and their statistics can be found in \cref{table:datasets};  see below for a list of detailed descriptions. 

\begin{table}
\centering
\caption{Statistics of the binary imbalanced classification datasets. For each split, we report the total number of samples ($N$), the number of positive (minority) and negative (majority) samples, and the imbalance ratio (neg:pos).}\label{tab:dataset_stats}
\resizebox{0.7\textwidth}{!}{
\begin{tabular}{lllrrrrr}
\toprule
Dataset & Modality & Split & $N$ & Feature Dimension & Pos & Neg & Imbalance \\
\midrule
\multirow{2}{*}{ADE-v2} & \multirow{2}{*}{Text}
& Train & 18{,}812 & 128 tokens & 5{,}439 & 13{,}373 & 2.5:1 \\
&& Test & 4{,}704 & 128 tokens & 1{,}354 & 3{,}350 & 2.5:1 \\
\midrule
\multirow{2}{*}{Creditcard} & \multirow{2}{*}{Tabular}
& Train & 227{,}845 & 28 & 386 & 227{,}459 & 589.3:1 \\
&& Test & 56{,}962 &  28 & 106 & 56{,}856 & 536.4:1 \\
\midrule
\multirow{2}{*}{Eyepacs} & \multirow{2}{*}{Image}
& Train & 35{,}126 & $224\times224\times3$ & 9{,}316 & 25{,}810 & 2.8:1 \\
&& Test & 53{,}576 & $224\times224\times3$ & 14{,}043 & 39{,}533 & 2.8:1 \\
\midrule
\multirow{2}{*}{MURA} & \multirow{2}{*}{Image}
& Train & 23{,}422 & 1536 & 1{,}487 & 21{,}935 & 14.8:1 \\
&& Test & 3{,}197 & 1536 & 1{,}530 & 1{,}667 & 1.1:1 \\
\bottomrule
\end{tabular}\label{table:datasets}
}
\end{table}

\begin{itemize}[leftmargin=*]


    \item  \textit{ADE-Corpora-V2}: ADE-Corpora-V2\footnote{\url{https://huggingface.co/datasets/SetFit/ade_corpus_v2_classification}} is a medical case report dataset that aims to classify if a sentence is related to an adverse drug reaction or not. As no test data are provided, we randomly divide the dataset into training and test sets with a ratio of $8:2$. 

    \item \textit{Creditcard}: the Credit Card Fraud Detection dataset consists of $284,807$ credit card transactions made by European cardholders over a two-day period in September 2013. Among these transactions, 492 are fraudulent, resulting in an extreme class imbalance with positive ratio of $0.172\%$. Each transaction is represented by $30$ numerical features, including $28$ PCA-transformed features, Time, and Amount, and is labeled as either legitimate or fraudulent.
    
    \item \textit{Eyepacs}: The Eyepacs dataset hosted by Kaggle\footnote{\url{https://www.kaggle.com/competitions/diabetic-retinopathy-detection}} is a large collection of high-resolution retina images taken under a variety of imaging conditions for the detection of diabetic retinopathy (DR). Based on clinical ratings, the images are graded into $5$ different severity levels with a ``No DR'' class. Accordingly, we transform it into a binary classification problem to detect the presence of DR. We follow their  official training-test data split and also resize the images to $224\times224$ for computational efficiency. 
    
    \item The \textit{MURA} (Musculoskeletal Radiographs) dataset is a large-scale collection of musculoskeletal X-ray images for abnormality detection. It comprises $40,561$ radiographs from $14,863$ studies, spanning $7$ anatomical regions: elbow, finger, forearm, hand, humerus, shoulder, and wrist. Each study is annotated as normal or abnormal based on assessments by board-certified radiologists.

\end{itemize}


\begin{table*}[!htbp]
\def\arraystretch{1.3}
\centering
\caption{Alpha ablation on \textcolor{red}{FPOR}. Values in (parentheses) are results after TA. Feasible solutions ($\text{precision} \ge \alpha$, with the target $\alpha$ shown per block) are \underline{underlined}, and among them the best objectives are shown in \textcolor{red}{red} (before TA) and \textcolor{blue}{blue} (after TA). For test, we also highlight the best $F_1$ scores. All underlines and highlights are up to $0.001$ slackness.}
\smallskip
\resizebox{1\textwidth}{!}{
\begin{tabular}{ll||cc||cc:c}
\Xhline{2\arrayrulewidth}
& &\multicolumn{2}{c||}{\textbf{train}} & \multicolumn{3}{c}{\textbf{test}} \\
dataset ($\alpha$) & method  & \multicolumn{1}{c}{precision---feasibility} & \multicolumn{1}{c||}{recall---objective} & \multicolumn{1}{c}{precision---feasibility} & \multicolumn{1}{c}{recall---objective} & F1-score  \\
\hline

	& WCE	& 0.709 $\pm$ {\tiny 0.030} (0.919 $\pm$ {\tiny 0.089})	& 0.277 $\pm$ {\tiny 0.238} (0.042 $\pm$ {\tiny 0.103})	& 0.722 $\pm$ {\tiny 0.023} (0.223 $\pm$ {\tiny 0.356})	& 0.273 $\pm$ {\tiny 0.238} (0.037 $\pm$ {\tiny 0.096}) & 0.346 $\pm$ {\tiny 0.213} (0.057 $\pm$ {\tiny 0.144}) \\
	& Focal	& 0.903 $\pm$ {\tiny 0.006} (\underline{0.957 $\pm$ {\tiny 0.014}})	& 0.190 $\pm$ {\tiny 0.098} (0.032 $\pm$ {\tiny 0.030})	& 0.913 $\pm$ {\tiny 0.021} (0.873 $\pm$ {\tiny 0.292})	& 0.181 $\pm$ {\tiny 0.091} (0.031 $\pm$ {\tiny 0.031}) & 0.292 $\pm$ {\tiny 0.129} (0.059 $\pm$ {\tiny 0.055}) \\
	& LDAM	& 0.808 $\pm$ {\tiny 0.005} (\underline{0.963 $\pm$ {\tiny 0.019}})	& 0.494 $\pm$ {\tiny 0.060} (0.023 $\pm$ {\tiny 0.013})	& 0.800 $\pm$ {\tiny 0.009} (\underline{0.981 $\pm$ {\tiny 0.018}})	& 0.482 $\pm$ {\tiny 0.057} (0.021 $\pm$ {\tiny 0.012}) & 0.599 $\pm$ {\tiny 0.044} (0.041 $\pm$ {\tiny 0.024}) \\
	& TERM	& 0.762 $\pm$ {\tiny 0.006} (0.921 $\pm$ {\tiny 0.058})	& 0.374 $\pm$ {\tiny 0.263} (0.011 $\pm$ {\tiny 0.008})	& 0.765 $\pm$ {\tiny 0.014} (0.693 $\pm$ {\tiny 0.395})	& 0.362 $\pm$ {\tiny 0.258} (0.009 $\pm$ {\tiny 0.009}) & 0.433 $\pm$ {\tiny 0.249} (0.017 $\pm$ {\tiny 0.018}) \\
	& ROS	& 0.701 $\pm$ {\tiny 0.033} (0.946 $\pm$ {\tiny 0.066})	& 0.491 $\pm$ {\tiny 0.318} (0.043 $\pm$ {\tiny 0.107})	& 0.707 $\pm$ {\tiny 0.033} (0.582 $\pm$ {\tiny 0.476})	& 0.481 $\pm$ {\tiny 0.314} (0.039 $\pm$ {\tiny 0.102}) & 0.499 $\pm$ {\tiny 0.252} (0.060 $\pm$ {\tiny 0.149}) \\
	& RUS	& 0.712 $\pm$ {\tiny 0.032} (0.889 $\pm$ {\tiny 0.108})	& 0.432 $\pm$ {\tiny 0.349} (0.123 $\pm$ {\tiny 0.145})	& 0.710 $\pm$ {\tiny 0.045} (0.698 $\pm$ {\tiny 0.363})	& 0.422 $\pm$ {\tiny 0.347} (0.115 $\pm$ {\tiny 0.138}) & 0.444 $\pm$ {\tiny 0.277} (0.177 $\pm$ {\tiny 0.206}) \\
	& TFCO	& 0.343 $\pm$ {\tiny 0.082} (0.627 $\pm$ {\tiny 0.283})	& 0.707 $\pm$ {\tiny 0.380} (0.139 $\pm$ {\tiny 0.290})	& 0.336 $\pm$ {\tiny 0.077} (0.424 $\pm$ {\tiny 0.289})	& 0.704 $\pm$ {\tiny 0.384} (0.137 $\pm$ {\tiny 0.290}) & 0.366 $\pm$ {\tiny 0.144} (0.101 $\pm$ {\tiny 0.139}) \\
\rowcolor{gray}
\multirow{-8}{*}{\cellcolor{white}ADE-v2 ($\alpha$=0.95)}	& ERO	& 0.946 $\pm$ {\tiny 0.003} (\underline{0.951 $\pm$ {\tiny 0.001}})	& 0.749 $\pm$ {\tiny 0.004} (\textcolor{blue}{0.282 $\pm$ {\tiny 0.027}})	& 0.924 $\pm$ {\tiny 0.006} (\underline{0.955 $\pm$ {\tiny 0.002}})	& 0.702 $\pm$ {\tiny 0.006} (\textcolor{blue}{0.267 $\pm$ {\tiny 0.027}}) & \textcolor{red}{0.798 $\pm$ {\tiny 0.004}} (\textcolor{blue}{0.416 $\pm$ {\tiny 0.033}}) \\

\hline

	& WCE	& 0.451 $\pm$ {\tiny 0.101} (0.844 $\pm$ {\tiny 0.191})	& 0.118 $\pm$ {\tiny 0.134} (0.005 $\pm$ {\tiny 0.007})	& \underline{0.955 $\pm$ {\tiny 0.047}} (\underline{1.000 $\pm$ {\tiny 0.000}})	& 0.117 $\pm$ {\tiny 0.131} (0.005 $\pm$ {\tiny 0.007}) & 0.182 $\pm$ {\tiny 0.195} (0.010 $\pm$ {\tiny 0.015}) \\
	& Focal	& 0.678 $\pm$ {\tiny 0.132} (0.870 $\pm$ {\tiny 0.165})	& 0.106 $\pm$ {\tiny 0.059} (0.021 $\pm$ {\tiny 0.028})	& \underline{0.974 $\pm$ {\tiny 0.017}} (\underline{0.989 $\pm$ {\tiny 0.013}})	& 0.102 $\pm$ {\tiny 0.053} (0.020 $\pm$ {\tiny 0.028}) & 0.180 $\pm$ {\tiny 0.089} (0.037 $\pm$ {\tiny 0.052}) \\
	& LDAM	& 0.770 $\pm$ {\tiny 0.094} (\underline{0.974 $\pm$ {\tiny 0.062}})	& 0.095 $\pm$ {\tiny 0.051} (0.003 $\pm$ {\tiny 0.005})	& 0.838 $\pm$ {\tiny 0.342} (0.857 $\pm$ {\tiny 0.350})	& 0.083 $\pm$ {\tiny 0.047} (0.002 $\pm$ {\tiny 0.003}) & 0.150 $\pm$ {\tiny 0.080} (0.004 $\pm$ {\tiny 0.006}) \\
	& TERM	& 0.649 $\pm$ {\tiny 0.152} (\underline{0.964 $\pm$ {\tiny 0.087}})	& 0.189 $\pm$ {\tiny 0.120} (0.002 $\pm$ {\tiny 0.001})	& \underline{0.970 $\pm$ {\tiny 0.019}} (0.857 $\pm$ {\tiny 0.350})	& 0.176 $\pm$ {\tiny 0.111} (0.002 $\pm$ {\tiny 0.001}) & 0.281 $\pm$ {\tiny 0.176} (0.003 $\pm$ {\tiny 0.002}) \\
	& ROS	& 0.389 $\pm$ {\tiny 0.054} (\underline{0.957 $\pm$ {\tiny 0.088}})	& 0.113 $\pm$ {\tiny 0.101} (0.002 $\pm$ {\tiny 0.002})	& 0.916 $\pm$ {\tiny 0.033} (0.600 $\pm$ {\tiny 0.490})	& 0.108 $\pm$ {\tiny 0.096} (0.001 $\pm$ {\tiny 0.002}) & 0.179 $\pm$ {\tiny 0.147} (0.002 $\pm$ {\tiny 0.003}) \\
	& RUS	& 0.397 $\pm$ {\tiny 0.055} (0.942 $\pm$ {\tiny 0.118})	& 0.091 $\pm$ {\tiny 0.114} (0.001 $\pm$ {\tiny 0.001})	& \underline{0.951 $\pm$ {\tiny 0.041}} (0.800 $\pm$ {\tiny 0.400})	& 0.092 $\pm$ {\tiny 0.111} (0.001 $\pm$ {\tiny 0.001}) & 0.148 $\pm$ {\tiny 0.162} (0.002 $\pm$ {\tiny 0.001}) \\
	& TFCO	& 0.066 $\pm$ {\tiny 0.005} (0.168 $\pm$ {\tiny 0.151})	& 0.810 $\pm$ {\tiny 0.294} (0.299 $\pm$ {\tiny 0.438})	& 0.473 $\pm$ {\tiny 0.013} (0.265 $\pm$ {\tiny 0.221})	& 0.781 $\pm$ {\tiny 0.305} (0.288 $\pm$ {\tiny 0.431}) & \textcolor{red}{0.550 $\pm$ {\tiny 0.181}} (0.198 $\pm$ {\tiny 0.286}) \\
\rowcolor{gray}
\multirow{-8}{*}{\cellcolor{white}MURA ($\alpha$=0.95)}	& ERO	& \underline{0.950 $\pm$ {\tiny 0.004}} (\underline{0.954 $\pm$ {\tiny 0.006}})	& \textcolor{red}{0.339 $\pm$ {\tiny 0.074}} (\textcolor{blue}{0.216 $\pm$ {\tiny 0.134}})	& \underline{0.954 $\pm$ {\tiny 0.015}} (\underline{0.956 $\pm$ {\tiny 0.014}})	& \textcolor{red}{0.208 $\pm$ {\tiny 0.036}} (\textcolor{blue}{0.138 $\pm$ {\tiny 0.079}}) & 0.340 $\pm$ {\tiny 0.047} (\textcolor{blue}{0.232 $\pm$ {\tiny 0.127}}) \\

\hline

	& WCE	& 0.709 $\pm$ {\tiny 0.030} (0.928 $\pm$ {\tiny 0.094})	& 0.277 $\pm$ {\tiny 0.238} (0.007 $\pm$ {\tiny 0.017})	& 0.722 $\pm$ {\tiny 0.023} (0.229 $\pm$ {\tiny 0.368})	& 0.273 $\pm$ {\tiny 0.238} (0.005 $\pm$ {\tiny 0.014}) & 0.346 $\pm$ {\tiny 0.213} (0.009 $\pm$ {\tiny 0.027}) \\
	& Focal	& 0.903 $\pm$ {\tiny 0.006} (\underline{0.999 $\pm$ {\tiny 0.003}})	& 0.190 $\pm$ {\tiny 0.098} (\textcolor{blue}{0.009 $\pm$ {\tiny 0.006}})	& 0.913 $\pm$ {\tiny 0.021} (0.900 $\pm$ {\tiny 0.300})	& 0.181 $\pm$ {\tiny 0.091} (0.007 $\pm$ {\tiny 0.005}) & 0.292 $\pm$ {\tiny 0.129} (0.014 $\pm$ {\tiny 0.011}) \\
	& LDAM	& 0.808 $\pm$ {\tiny 0.005} (\underline{1.000 $\pm$ {\tiny 0.000}})	& 0.494 $\pm$ {\tiny 0.060} (0.006 $\pm$ {\tiny 0.003})	& 0.800 $\pm$ {\tiny 0.009} (0.800 $\pm$ {\tiny 0.400})	& 0.482 $\pm$ {\tiny 0.057} (0.005 $\pm$ {\tiny 0.003}) & 0.599 $\pm$ {\tiny 0.044} (0.009 $\pm$ {\tiny 0.006}) \\
	& TERM	& 0.762 $\pm$ {\tiny 0.006} (0.942 $\pm$ {\tiny 0.072})	& 0.374 $\pm$ {\tiny 0.263} (0.005 $\pm$ {\tiny 0.003})	& 0.765 $\pm$ {\tiny 0.014} (0.700 $\pm$ {\tiny 0.400})	& 0.362 $\pm$ {\tiny 0.258} (0.003 $\pm$ {\tiny 0.003}) & 0.433 $\pm$ {\tiny 0.249} (0.006 $\pm$ {\tiny 0.005}) \\
	& ROS	& 0.701 $\pm$ {\tiny 0.033} (0.964 $\pm$ {\tiny 0.072})	& 0.491 $\pm$ {\tiny 0.318} (0.005 $\pm$ {\tiny 0.005})	& 0.707 $\pm$ {\tiny 0.033} (0.592 $\pm$ {\tiny 0.484})	& 0.481 $\pm$ {\tiny 0.314} (0.003 $\pm$ {\tiny 0.003}) & 0.499 $\pm$ {\tiny 0.252} (0.006 $\pm$ {\tiny 0.006}) \\
	& RUS	& 0.712 $\pm$ {\tiny 0.032} (0.916 $\pm$ {\tiny 0.125})	& 0.432 $\pm$ {\tiny 0.349} (0.022 $\pm$ {\tiny 0.043})	& 0.710 $\pm$ {\tiny 0.045} (0.642 $\pm$ {\tiny 0.433})	& 0.422 $\pm$ {\tiny 0.347} (0.020 $\pm$ {\tiny 0.042}) & 0.444 $\pm$ {\tiny 0.277} (0.036 $\pm$ {\tiny 0.071}) \\
	& TFCO	& 0.372 $\pm$ {\tiny 0.085} (0.659 $\pm$ {\tiny 0.248})	& 0.663 $\pm$ {\tiny 0.318} (0.067 $\pm$ {\tiny 0.139})	& 0.369 $\pm$ {\tiny 0.084} (0.579 $\pm$ {\tiny 0.385})	& 0.663 $\pm$ {\tiny 0.320} (0.064 $\pm$ {\tiny 0.135}) & 0.410 $\pm$ {\tiny 0.095} (0.068 $\pm$ {\tiny 0.118}) \\
\rowcolor{gray}
\multirow{-8}{*}{\cellcolor{white}ADE-v2 ($\alpha$=0.99)}	& ERO	& 0.957 $\pm$ {\tiny 0.003} (0.989 $\pm$ {\tiny 0.004})	& 0.694 $\pm$ {\tiny 0.008} (0.048 $\pm$ {\tiny 0.014})	& 0.948 $\pm$ {\tiny 0.005} (\underline{0.991 $\pm$ {\tiny 0.011}})	& 0.640 $\pm$ {\tiny 0.010} (\textcolor{blue}{0.047 $\pm$ {\tiny 0.013}}) & \textcolor{red}{0.764 $\pm$ {\tiny 0.006}} (\textcolor{blue}{0.089 $\pm$ {\tiny 0.024}}) \\

\hline

	& WCE	& 0.451 $\pm$ {\tiny 0.101} (0.844 $\pm$ {\tiny 0.191})	& 0.118 $\pm$ {\tiny 0.134} (0.005 $\pm$ {\tiny 0.007})	& 0.955 $\pm$ {\tiny 0.047} (\underline{1.000 $\pm$ {\tiny 0.000}})	& 0.117 $\pm$ {\tiny 0.131} (\textcolor{blue}{0.005 $\pm$ {\tiny 0.007}}) & 0.182 $\pm$ {\tiny 0.195} (0.010 $\pm$ {\tiny 0.015}) \\
	& Focal	& 0.678 $\pm$ {\tiny 0.132} (0.870 $\pm$ {\tiny 0.165})	& 0.106 $\pm$ {\tiny 0.059} (0.021 $\pm$ {\tiny 0.028})	& 0.974 $\pm$ {\tiny 0.017} (0.989 $\pm$ {\tiny 0.013})	& 0.102 $\pm$ {\tiny 0.053} (0.020 $\pm$ {\tiny 0.028}) & 0.180 $\pm$ {\tiny 0.089} (0.037 $\pm$ {\tiny 0.052}) \\
	& LDAM	& 0.770 $\pm$ {\tiny 0.094} (0.974 $\pm$ {\tiny 0.062})	& 0.095 $\pm$ {\tiny 0.051} (0.003 $\pm$ {\tiny 0.005})	& 0.838 $\pm$ {\tiny 0.342} (0.857 $\pm$ {\tiny 0.350})	& 0.083 $\pm$ {\tiny 0.047} (0.002 $\pm$ {\tiny 0.003}) & 0.150 $\pm$ {\tiny 0.080} (0.004 $\pm$ {\tiny 0.006}) \\
	& TERM	& 0.649 $\pm$ {\tiny 0.152} (0.964 $\pm$ {\tiny 0.087})	& 0.189 $\pm$ {\tiny 0.120} (0.002 $\pm$ {\tiny 0.001})	& 0.970 $\pm$ {\tiny 0.019} (0.857 $\pm$ {\tiny 0.350})	& 0.176 $\pm$ {\tiny 0.111} (0.002 $\pm$ {\tiny 0.001}) & 0.281 $\pm$ {\tiny 0.176} (0.003 $\pm$ {\tiny 0.002}) \\
	& ROS	& 0.389 $\pm$ {\tiny 0.054} (0.957 $\pm$ {\tiny 0.088})	& 0.113 $\pm$ {\tiny 0.101} (0.002 $\pm$ {\tiny 0.002})	& 0.916 $\pm$ {\tiny 0.033} (0.600 $\pm$ {\tiny 0.490})	& 0.108 $\pm$ {\tiny 0.096} (0.001 $\pm$ {\tiny 0.002}) & 0.179 $\pm$ {\tiny 0.147} (0.002 $\pm$ {\tiny 0.003}) \\
	& RUS	& 0.397 $\pm$ {\tiny 0.055} (0.942 $\pm$ {\tiny 0.118})	& 0.091 $\pm$ {\tiny 0.114} (0.001 $\pm$ {\tiny 0.001})	& 0.951 $\pm$ {\tiny 0.041} (0.800 $\pm$ {\tiny 0.400})	& 0.092 $\pm$ {\tiny 0.111} (0.001 $\pm$ {\tiny 0.001}) & 0.148 $\pm$ {\tiny 0.162} (0.002 $\pm$ {\tiny 0.001}) \\
	& TFCO	& 0.065 $\pm$ {\tiny 0.003} (0.137 $\pm$ {\tiny 0.114})	& 0.801 $\pm$ {\tiny 0.299} (0.467 $\pm$ {\tiny 0.468})	& 0.460 $\pm$ {\tiny 0.037} (0.302 $\pm$ {\tiny 0.209})	& 0.769 $\pm$ {\tiny 0.318} (0.451 $\pm$ {\tiny 0.455}) & \textcolor{red}{0.542 $\pm$ {\tiny 0.185}} (\textcolor{blue}{0.311 $\pm$ {\tiny 0.310}}) \\
\rowcolor{gray}
\multirow{-8}{*}{\cellcolor{white}MURA ($\alpha$=0.99)}	& ERO	& \underline{0.991 $\pm$ {\tiny 0.001}} (\underline{0.993 $\pm$ {\tiny 0.004}})	& \textcolor{red}{0.308 $\pm$ {\tiny 0.064}} (\textcolor{blue}{0.231 $\pm$ {\tiny 0.118}})	& 0.960 $\pm$ {\tiny 0.013} (0.972 $\pm$ {\tiny 0.017})	& 0.184 $\pm$ {\tiny 0.028} (0.141 $\pm$ {\tiny 0.068}) & 0.308 $\pm$ {\tiny 0.038} (0.239 $\pm$ {\tiny 0.110}) \\

\hline
\Xhline{3\arrayrulewidth}
\end{tabular}
}
\label{table:FPOR_ab_alpha}
\end{table*}

\begin{table*}[!htbp]
\def\arraystretch{1.3}
\centering
\caption{Alpha ablation on \textcolor{red}{FROP}. Values in (parentheses) are results after TA. Feasible solutions ($\text{recall} \ge \alpha$, with the target $\alpha$ shown per block) are \underline{underlined}, and among them the best objectives are shown in \textcolor{red}{red} (before TA) and \textcolor{blue}{blue} (after TA). For test, we also highlight the best $F_1$ scores. All underlines and highlights are up to $0.001$ slackness.}
\smallskip
\resizebox{1\textwidth}{!}{
\begin{tabular}{ll||cc||cc:c}
\Xhline{2\arrayrulewidth}
& &\multicolumn{2}{c||}{\textbf{train}} & \multicolumn{3}{c}{\textbf{test}} \\
dataset ($\alpha$) & method  & \multicolumn{1}{c}{recall---feasibility} & \multicolumn{1}{c||}{precision---objective} & \multicolumn{1}{c}{recall---feasibility} & \multicolumn{1}{c}{precision---objective} & F1-score  \\
\hline

	& WCE	& 0.909 $\pm$ {\tiny 0.010} (\underline{0.951 $\pm$ {\tiny 0.001}})	& 0.433 $\pm$ {\tiny 0.020} (0.394 $\pm$ {\tiny 0.016})	& 0.896 $\pm$ {\tiny 0.011} (0.941 $\pm$ {\tiny 0.002})	& 0.429 $\pm$ {\tiny 0.020} (0.391 $\pm$ {\tiny 0.015}) & 0.580 $\pm$ {\tiny 0.018} (0.552 $\pm$ {\tiny 0.015}) \\
	& Focal	& 0.915 $\pm$ {\tiny 0.010} (\underline{0.950 $\pm$ {\tiny 0.000}})	& 0.515 $\pm$ {\tiny 0.015} (0.468 $\pm$ {\tiny 0.006})	& 0.903 $\pm$ {\tiny 0.011} (0.945 $\pm$ {\tiny 0.003})	& 0.511 $\pm$ {\tiny 0.015} (0.465 $\pm$ {\tiny 0.006}) & 0.653 $\pm$ {\tiny 0.010} (0.624 $\pm$ {\tiny 0.006}) \\
	& LDAM	& 0.770 $\pm$ {\tiny 0.008} (\underline{0.951 $\pm$ {\tiny 0.001}})	& 0.519 $\pm$ {\tiny 0.101} (0.398 $\pm$ {\tiny 0.064})	& 0.757 $\pm$ {\tiny 0.010} (0.940 $\pm$ {\tiny 0.002})	& 0.521 $\pm$ {\tiny 0.095} (0.395 $\pm$ {\tiny 0.064}) & 0.612 $\pm$ {\tiny 0.054} (0.554 $\pm$ {\tiny 0.057}) \\
	& TERM	& \underline{0.955 $\pm$ {\tiny 0.026}} (\underline{0.950 $\pm$ {\tiny 0.001}})	& 0.358 $\pm$ {\tiny 0.036} (0.365 $\pm$ {\tiny 0.013})	& 0.947 $\pm$ {\tiny 0.030} (0.943 $\pm$ {\tiny 0.004})	& 0.355 $\pm$ {\tiny 0.036} (0.362 $\pm$ {\tiny 0.012}) & 0.515 $\pm$ {\tiny 0.035} (0.523 $\pm$ {\tiny 0.012}) \\
	& ROS	& 0.915 $\pm$ {\tiny 0.023} (\underline{0.950 $\pm$ {\tiny 0.000}})	& 0.427 $\pm$ {\tiny 0.036} (0.394 $\pm$ {\tiny 0.016})	& 0.902 $\pm$ {\tiny 0.023} (0.940 $\pm$ {\tiny 0.002})	& 0.424 $\pm$ {\tiny 0.036} (0.392 $\pm$ {\tiny 0.016}) & 0.576 $\pm$ {\tiny 0.029} (0.553 $\pm$ {\tiny 0.016}) \\
	& RUS	& 0.916 $\pm$ {\tiny 0.013} (\underline{0.950 $\pm$ {\tiny 0.000}})	& 0.455 $\pm$ {\tiny 0.088} (0.420 $\pm$ {\tiny 0.073})	& 0.902 $\pm$ {\tiny 0.014} (0.938 $\pm$ {\tiny 0.003})	& 0.449 $\pm$ {\tiny 0.084} (0.416 $\pm$ {\tiny 0.071}) & 0.595 $\pm$ {\tiny 0.064} (0.573 $\pm$ {\tiny 0.060}) \\
	& TFCO	& \underline{0.976 $\pm$ {\tiny 0.017}} (\underline{0.950 $\pm$ {\tiny 0.000}})	& 0.348 $\pm$ {\tiny 0.032} (0.375 $\pm$ {\tiny 0.019})	& \underline{0.973 $\pm$ {\tiny 0.019}} (0.945 $\pm$ {\tiny 0.004})	& \textcolor{red}{0.346 $\pm$ {\tiny 0.032}} (0.372 $\pm$ {\tiny 0.020}) & 0.510 $\pm$ {\tiny 0.031} (0.534 $\pm$ {\tiny 0.020}) \\
\rowcolor{gray}
\multirow{-8}{*}{\cellcolor{white}ADE-v2 ($\alpha$=0.95)}	& ERO	& \underline{0.950 $\pm$ {\tiny 0.000}} (\underline{0.950 $\pm$ {\tiny 0.000}})	& \textcolor{red}{0.671 $\pm$ {\tiny 0.003}} (\textcolor{blue}{0.657 $\pm$ {\tiny 0.010}})	& 0.937 $\pm$ {\tiny 0.002} (0.939 $\pm$ {\tiny 0.002})	& 0.658 $\pm$ {\tiny 0.003} (0.643 $\pm$ {\tiny 0.009}) & \textcolor{red}{0.773 $\pm$ {\tiny 0.002}} (\textcolor{blue}{0.763 $\pm$ {\tiny 0.006}}) \\

\hline

	& WCE	& 0.922 $\pm$ {\tiny 0.009} (\underline{0.951 $\pm$ {\tiny 0.000}})	& 0.080 $\pm$ {\tiny 0.002} (0.073 $\pm$ {\tiny 0.001})	& 0.907 $\pm$ {\tiny 0.012} (0.944 $\pm$ {\tiny 0.003})	& 0.541 $\pm$ {\tiny 0.005} (0.520 $\pm$ {\tiny 0.005}) & 0.678 $\pm$ {\tiny 0.003} (0.671 $\pm$ {\tiny 0.004}) \\
	& Focal	& \underline{0.974 $\pm$ {\tiny 0.037}} (\underline{0.964 $\pm$ {\tiny 0.020}})	& 0.066 $\pm$ {\tiny 0.003} (0.065 $\pm$ {\tiny 0.001})	& \underline{0.976 $\pm$ {\tiny 0.030}} (\underline{0.960 $\pm$ {\tiny 0.022}})	& 0.487 $\pm$ {\tiny 0.013} (\textcolor{blue}{0.486 $\pm$ {\tiny 0.007}}) & 0.650 $\pm$ {\tiny 0.005} (0.645 $\pm$ {\tiny 0.006}) \\
	& LDAM	& 0.737 $\pm$ {\tiny 0.153} (\underline{0.951 $\pm$ {\tiny 0.001}})	& 0.096 $\pm$ {\tiny 0.021} (0.068 $\pm$ {\tiny 0.001})	& 0.706 $\pm$ {\tiny 0.167} (0.945 $\pm$ {\tiny 0.006})	& 0.570 $\pm$ {\tiny 0.046} (0.495 $\pm$ {\tiny 0.002}) & 0.613 $\pm$ {\tiny 0.067} (0.650 $\pm$ {\tiny 0.002}) \\
	& TERM	& \underline{0.963 $\pm$ {\tiny 0.025}} (\underline{0.951 $\pm$ {\tiny 0.001}})	& 0.068 $\pm$ {\tiny 0.003} (0.069 $\pm$ {\tiny 0.001})	& \underline{0.955 $\pm$ {\tiny 0.031}} (0.942 $\pm$ {\tiny 0.005})	& \textcolor{red}{0.493 $\pm$ {\tiny 0.011}} (0.496 $\pm$ {\tiny 0.004}) & 0.649 $\pm$ {\tiny 0.003} (0.650 $\pm$ {\tiny 0.004}) \\
	& ROS	& 0.913 $\pm$ {\tiny 0.010} (\underline{0.950 $\pm$ {\tiny 0.000}})	& 0.081 $\pm$ {\tiny 0.005} (0.073 $\pm$ {\tiny 0.002})	& 0.898 $\pm$ {\tiny 0.017} (0.944 $\pm$ {\tiny 0.004})	& 0.547 $\pm$ {\tiny 0.019} (0.521 $\pm$ {\tiny 0.009}) & 0.679 $\pm$ {\tiny 0.010} (0.671 $\pm$ {\tiny 0.007}) \\
	& RUS	& 0.920 $\pm$ {\tiny 0.008} (\underline{0.950 $\pm$ {\tiny 0.000}})	& 0.080 $\pm$ {\tiny 0.003} (0.073 $\pm$ {\tiny 0.002})	& 0.908 $\pm$ {\tiny 0.008} (0.945 $\pm$ {\tiny 0.007})	& 0.542 $\pm$ {\tiny 0.011} (0.522 $\pm$ {\tiny 0.008}) & 0.679 $\pm$ {\tiny 0.007} (0.673 $\pm$ {\tiny 0.005}) \\
	& TFCO	& 0.685 $\pm$ {\tiny 0.438} (\underline{0.952 $\pm$ {\tiny 0.003}})	& 0.210 $\pm$ {\tiny 0.227} (0.066 $\pm$ {\tiny 0.001})	& 0.682 $\pm$ {\tiny 0.435} (0.945 $\pm$ {\tiny 0.011})	& 0.527 $\pm$ {\tiny 0.255} (0.489 $\pm$ {\tiny 0.004}) & 0.462 $\pm$ {\tiny 0.277} (0.644 $\pm$ {\tiny 0.004}) \\
\rowcolor{gray}
\multirow{-8}{*}{\cellcolor{white}MURA ($\alpha$=0.95)}	& ERO	& \underline{0.950 $\pm$ {\tiny 0.001}} (\underline{0.950 $\pm$ {\tiny 0.000}})	& \textcolor{red}{0.123 $\pm$ {\tiny 0.033}} (\textcolor{blue}{0.121 $\pm$ {\tiny 0.034}})	& 0.816 $\pm$ {\tiny 0.060} (0.825 $\pm$ {\tiny 0.068})	& 0.613 $\pm$ {\tiny 0.067} (0.610 $\pm$ {\tiny 0.069}) & \textcolor{red}{0.694 $\pm$ {\tiny 0.026}} (\textcolor{blue}{0.695 $\pm$ {\tiny 0.026}}) \\

\hline

	& WCE	& 0.909 $\pm$ {\tiny 0.010} (\underline{0.990 $\pm$ {\tiny 0.000}})	& 0.433 $\pm$ {\tiny 0.020} (0.333 $\pm$ {\tiny 0.006})	& 0.896 $\pm$ {\tiny 0.011} (0.984 $\pm$ {\tiny 0.001})	& 0.429 $\pm$ {\tiny 0.020} (0.330 $\pm$ {\tiny 0.006}) & 0.580 $\pm$ {\tiny 0.018} (0.494 $\pm$ {\tiny 0.007}) \\
	& Focal	& 0.915 $\pm$ {\tiny 0.010} (\underline{0.990 $\pm$ {\tiny 0.000}})	& 0.515 $\pm$ {\tiny 0.015} (0.374 $\pm$ {\tiny 0.004})	& 0.903 $\pm$ {\tiny 0.011} (\underline{0.989 $\pm$ {\tiny 0.001}})	& 0.511 $\pm$ {\tiny 0.015} (\textcolor{blue}{0.373 $\pm$ {\tiny 0.004}}) & 0.653 $\pm$ {\tiny 0.010} (0.541 $\pm$ {\tiny 0.004}) \\
	& LDAM	& 0.770 $\pm$ {\tiny 0.008} (\underline{0.990 $\pm$ {\tiny 0.000}})	& 0.519 $\pm$ {\tiny 0.101} (0.335 $\pm$ {\tiny 0.039})	& 0.757 $\pm$ {\tiny 0.010} (0.984 $\pm$ {\tiny 0.002})	& 0.521 $\pm$ {\tiny 0.095} (0.332 $\pm$ {\tiny 0.040}) & 0.612 $\pm$ {\tiny 0.054} (0.496 $\pm$ {\tiny 0.041}) \\
	& TERM	& 0.955 $\pm$ {\tiny 0.026} (\underline{0.990 $\pm$ {\tiny 0.000}})	& 0.358 $\pm$ {\tiny 0.036} (0.319 $\pm$ {\tiny 0.006})	& 0.947 $\pm$ {\tiny 0.030} (0.985 $\pm$ {\tiny 0.003})	& 0.355 $\pm$ {\tiny 0.036} (0.317 $\pm$ {\tiny 0.006}) & 0.515 $\pm$ {\tiny 0.035} (0.480 $\pm$ {\tiny 0.007}) \\
	& ROS	& 0.915 $\pm$ {\tiny 0.023} (\underline{0.990 $\pm$ {\tiny 0.000}})	& 0.427 $\pm$ {\tiny 0.036} (0.332 $\pm$ {\tiny 0.007})	& 0.902 $\pm$ {\tiny 0.023} (0.984 $\pm$ {\tiny 0.001})	& 0.424 $\pm$ {\tiny 0.036} (0.329 $\pm$ {\tiny 0.007}) & 0.576 $\pm$ {\tiny 0.029} (0.493 $\pm$ {\tiny 0.008}) \\
	& RUS	& 0.916 $\pm$ {\tiny 0.013} (\underline{0.990 $\pm$ {\tiny 0.000}})	& 0.455 $\pm$ {\tiny 0.088} (0.350 $\pm$ {\tiny 0.045})	& 0.902 $\pm$ {\tiny 0.014} (0.984 $\pm$ {\tiny 0.001})	& 0.449 $\pm$ {\tiny 0.084} (0.347 $\pm$ {\tiny 0.045}) & 0.595 $\pm$ {\tiny 0.064} (0.512 $\pm$ {\tiny 0.046}) \\
	& TFCO	& \underline{0.996 $\pm$ {\tiny 0.003}} (\underline{0.990 $\pm$ {\tiny 0.000}})	& 0.316 $\pm$ {\tiny 0.013} (0.328 $\pm$ {\tiny 0.006})	& \underline{0.996 $\pm$ {\tiny 0.004}} (0.989 $\pm$ {\tiny 0.001})	& \textcolor{red}{0.314 $\pm$ {\tiny 0.012}} (0.327 $\pm$ {\tiny 0.006}) & 0.477 $\pm$ {\tiny 0.014} (0.492 $\pm$ {\tiny 0.006}) \\
\rowcolor{gray}
\multirow{-8}{*}{\cellcolor{white}ADE-v2 ($\alpha$=0.99)}	& ERO	& \underline{0.990 $\pm$ {\tiny 0.000}} (\underline{0.990 $\pm$ {\tiny 0.000}})	& \textcolor{red}{0.629 $\pm$ {\tiny 0.006}} (\textcolor{blue}{0.515 $\pm$ {\tiny 0.040}})	& 0.980 $\pm$ {\tiny 0.001} (0.985 $\pm$ {\tiny 0.003})	& 0.614 $\pm$ {\tiny 0.006} (0.512 $\pm$ {\tiny 0.035}) & \textcolor{red}{0.755 $\pm$ {\tiny 0.005}} (\textcolor{blue}{0.673 $\pm$ {\tiny 0.028}}) \\

\hline

	& WCE	& 0.922 $\pm$ {\tiny 0.009} (\underline{0.991 $\pm$ {\tiny 0.000}})	& 0.080 $\pm$ {\tiny 0.002} (0.065 $\pm$ {\tiny 0.000})	& 0.907 $\pm$ {\tiny 0.012} (\underline{0.992 $\pm$ {\tiny 0.001}})	& 0.541 $\pm$ {\tiny 0.005} (\textcolor{blue}{0.489 $\pm$ {\tiny 0.002}}) & 0.678 $\pm$ {\tiny 0.003} (0.655 $\pm$ {\tiny 0.002}) \\
	& Focal	& 0.974 $\pm$ {\tiny 0.037} (\underline{0.993 $\pm$ {\tiny 0.003}})	& 0.066 $\pm$ {\tiny 0.003} (0.064 $\pm$ {\tiny 0.000})	& 0.976 $\pm$ {\tiny 0.030} (0.989 $\pm$ {\tiny 0.004})	& 0.487 $\pm$ {\tiny 0.013} (0.480 $\pm$ {\tiny 0.002}) & 0.650 $\pm$ {\tiny 0.005} (0.646 $\pm$ {\tiny 0.002}) \\
	& LDAM	& 0.737 $\pm$ {\tiny 0.153} (\underline{0.991 $\pm$ {\tiny 0.000}})	& 0.096 $\pm$ {\tiny 0.021} (0.064 $\pm$ {\tiny 0.000})	& 0.706 $\pm$ {\tiny 0.167} (\underline{0.990 $\pm$ {\tiny 0.003}})	& 0.570 $\pm$ {\tiny 0.046} (0.482 $\pm$ {\tiny 0.001}) & 0.613 $\pm$ {\tiny 0.067} (0.649 $\pm$ {\tiny 0.001}) \\
	& TERM	& 0.963 $\pm$ {\tiny 0.025} (\underline{0.991 $\pm$ {\tiny 0.000}})	& 0.068 $\pm$ {\tiny 0.003} (0.065 $\pm$ {\tiny 0.000})	& 0.955 $\pm$ {\tiny 0.031} (0.987 $\pm$ {\tiny 0.005})	& 0.493 $\pm$ {\tiny 0.011} (0.482 $\pm$ {\tiny 0.001}) & 0.649 $\pm$ {\tiny 0.003} (0.648 $\pm$ {\tiny 0.002}) \\
	& ROS	& 0.913 $\pm$ {\tiny 0.010} (\underline{0.991 $\pm$ {\tiny 0.001}})	& 0.081 $\pm$ {\tiny 0.005} (0.065 $\pm$ {\tiny 0.000})	& 0.898 $\pm$ {\tiny 0.017} (\underline{0.992 $\pm$ {\tiny 0.002}})	& 0.547 $\pm$ {\tiny 0.019} (\textcolor{blue}{0.489 $\pm$ {\tiny 0.003}}) & 0.679 $\pm$ {\tiny 0.010} (0.655 $\pm$ {\tiny 0.003}) \\
	& RUS	& 0.920 $\pm$ {\tiny 0.008} (\underline{0.991 $\pm$ {\tiny 0.000}})	& 0.080 $\pm$ {\tiny 0.003} (0.065 $\pm$ {\tiny 0.001})	& 0.908 $\pm$ {\tiny 0.008} (\underline{0.991 $\pm$ {\tiny 0.002}})	& 0.542 $\pm$ {\tiny 0.011} (\textcolor{blue}{0.489 $\pm$ {\tiny 0.003}}) & 0.679 $\pm$ {\tiny 0.007} (0.655 $\pm$ {\tiny 0.002}) \\
	& TFCO	& 0.329 $\pm$ {\tiny 0.440} (\underline{0.992 $\pm$ {\tiny 0.002}})	& 0.333 $\pm$ {\tiny 0.314} (0.064 $\pm$ {\tiny 0.000})	& 0.323 $\pm$ {\tiny 0.437} (0.988 $\pm$ {\tiny 0.005})	& 0.651 $\pm$ {\tiny 0.291} (0.481 $\pm$ {\tiny 0.001}) & 0.239 $\pm$ {\tiny 0.276} (0.647 $\pm$ {\tiny 0.002}) \\
\rowcolor{gray}
\multirow{-8}{*}{\cellcolor{white}MURA ($\alpha$=0.99)}	& ERO	& 0.988 $\pm$ {\tiny 0.006} (\underline{0.991 $\pm$ {\tiny 0.001}})	& 0.100 $\pm$ {\tiny 0.024} (\textcolor{blue}{0.076 $\pm$ {\tiny 0.010}})	& 0.886 $\pm$ {\tiny 0.062} (0.958 $\pm$ {\tiny 0.031})	& 0.570 $\pm$ {\tiny 0.054} (0.521 $\pm$ {\tiny 0.029}) & \textcolor{red}{0.689 $\pm$ {\tiny 0.024}} (\textcolor{blue}{0.673 $\pm$ {\tiny 0.016}}) \\

\hline
\Xhline{3\arrayrulewidth}
\end{tabular}
}
\label{table:FROP_ab_alpha}
\end{table*}

\subsection{Further implementation details} \label{sec:imp-details}

\subsubsection{Details on model training} 
We train the WCE models using \textit{ADAM} with an initial learning rate of $0.001$ and the \textit{CosineAnnealingLR} scheduler. We set a maximum of $30,000$ iterations and terminate the iteration process when the loss does not decrease during the past $10$ iterations. For TFCO, we adopt the training pipeline provided in their official GitHub repository. We fix the maximum number of outer iterations to $1000$ and use the \textit{Adagrad} optimizer. We use sigmoid to approximate the indicator function. We initialize the model weights using \textit{HeNormal} for dense layers, with biases set to zero. We perform a grid search over learning rates $\textit{lr} \in \{1, 0.1, 0.01, 0.001, 0.0001, 0.00001\}$ and dual variable scaling factors $\textit{dual\_scale} \in \{0.1, 1, 10\}$ to select the best model for final evaluation. For SigmoidF1, we follow the training protocol as in the WCE setup. As there are two important hyperparameters, $T$ (temperature scaling factor) and $b$ (horizontal offsets), in their approximation to the $F_1$-score, we perform a grid search, $T\in\{1, 10, 20, 30\}$ and $b\in\{0, 1, 2\}$, to select the best combination of hyperparameters for training. For our ERO, we follow the same optimization setting as in WCE and SigmoidF1 to solve the subproblem in \cref{alg:alm}. We set other hyperparameters in \cref{alg:alm} as follows: we randomly initialize $\mb \theta^0$ and $\mb s^0$, and set $\lambda^{(0)}=1$, $\rho=1.3$, $K=50$, and $\gamma=0.5 * \rho^{k}$ where $k$ is the iteration number (i.e., the regularization parameter is dynamically adjusted to match the rate of growth in the penalty parameter $\lambda$). To prevent numerical instability caused by the growth of the penalty, we set $\lambda_{\max}=10^6$ and $\gamma_{\max}=10$. For OFBS, we set $\rho=1.5$ and $\lambda=1$, while keeping all other hyperparameters unchanged. For all methods, we repeat the experiments $10$ times and report the mean and the standard deviation. All experiments are performed on a system equipped with an NVIDIA A100 GPU and an AMD EPYC 7763 64-core processor.

\subsection{Feature extraction with foundation models}  
For image data, we use DINOv2\footnote{\url{https://github.com/facebookresearch/dinov2}}, a state-of-the-art vision foundation model based on self-supervised learning, as the feature extractor. Specifically, we choose ViT-g/14, the largest pretrained model with $1.1$B weights. We resize the input image to $224\times 224$, and the resulting feature dimension is $1024$. For NLP data, we adopt the bert-base-uncased model\footnote{\url{https://huggingface.co/google-bert/bert-base-uncased}} from huggingface. It has $110$M weights and outputs $768$ features per input.

\section{Further analysis and ablation study}
\label{app:further-ablation}

This section provides (1) a stress test under stricter constraints, and (2) ablation studies on the key components of ERO. Specifically, we examine whether ERO benefits from three algorithmic ingredients: the exact reformulation (\cref{eq:FPOR-almost}, \cref{def:dmo_reform_linear}), the logit regularization (\cref{eq:reg-term}), and the inequality reformulation (\cref{eq:proof-frop-key-1}). We focus these studies on \textit{MURA} and \textit{ADE-v2}, which together span text and image modalities, mild and severe class imbalance, and train–test distribution shifts, offering a representative slice of real-world conditions. 

\begin{figure*}[!htbp]
    \centering
    \includegraphics[width=1\textwidth]{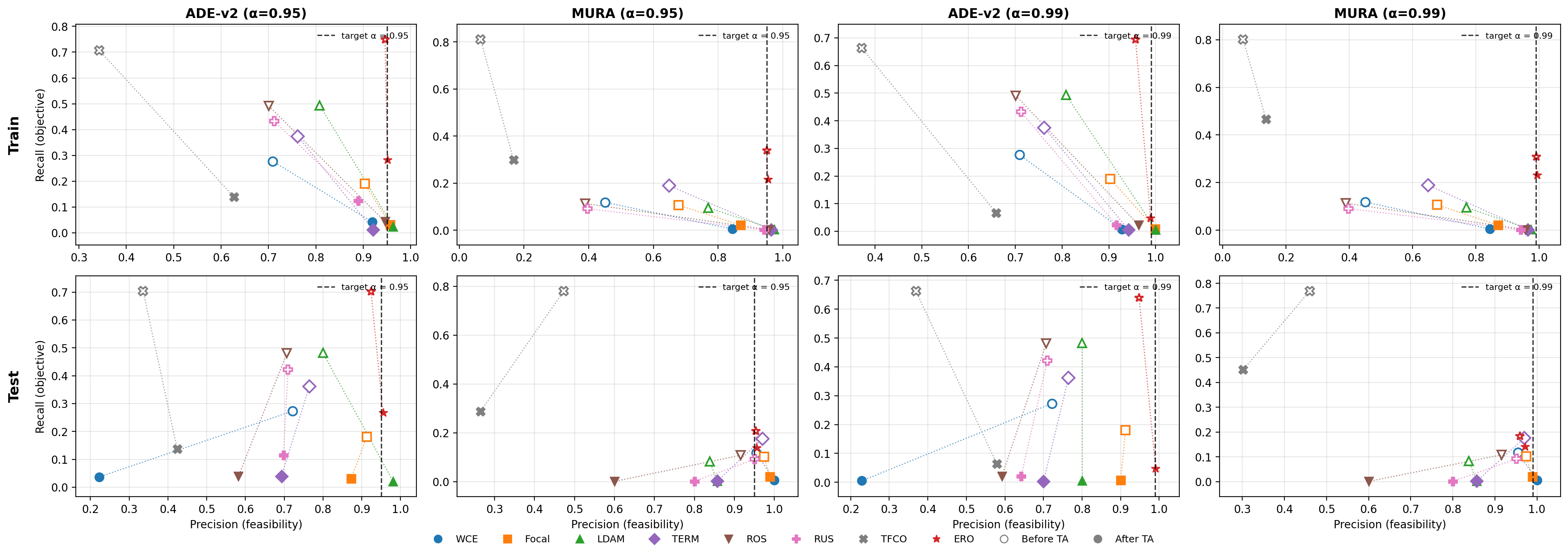}
    \caption{Feasibility stress test for FPOR on \textit{ADE-v2} and \textit{MURA} with $\alpha = 0.95$ and $0.99$.}
    \label{fig:ablation_alpha_fpor}
    \vspace{-1em}
\end{figure*} 

\begin{figure*}[!htbp]
    \centering
    \includegraphics[width=1\textwidth]{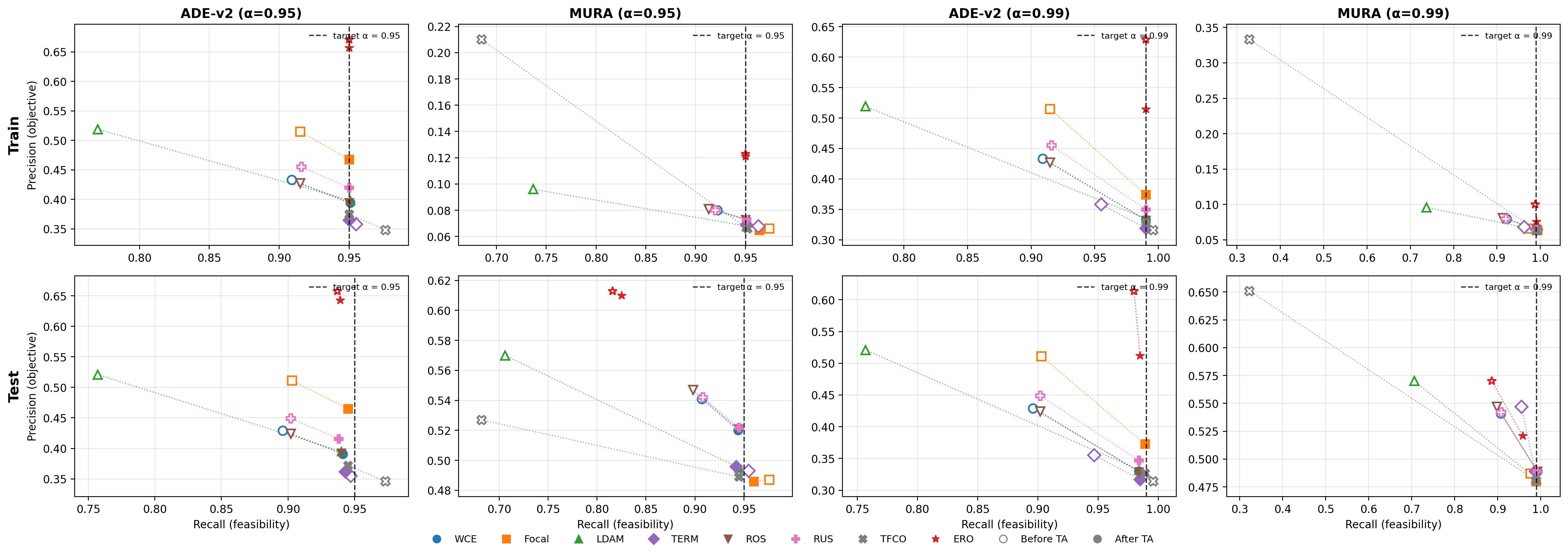}
    \caption{Feasibility stress test for FROP on \textit{ADE-v2} and \textit{MURA} with $\alpha = 0.95$ and $0.99$.}
    \label{fig:ablation_alpha_frop}
    \vspace{-1em}
\end{figure*} 

\subsection{Feasibility stress test}\label{para:ab_alpha_sweep}

To further probe all methods' abilities to achieve feasibility under stress, we repeat the main experiments with substantially tighter targets, $\alpha = 0.95$ and $0.99$, and summarized the results in \cref{fig:ablation_alpha_fpor} and \cref{fig:ablation_alpha_frop}. DMO+ERO remains the strongest performer during training, achieving feasible solutions on $5$ of $8$ tasks before TA and $7$ of $8$ after TA. In  contrast, none of the competing methods produces a feasible solution on any FPOR task before TA, and post-TA feasibility, when achieved, typically comes at the cost of a near-collapsed objective (recall below $5\%$). On FROP, several baselines do satisfy the feasibility bar, but their objective values still trail well behind DMO+ERO's. This advantage largely carries over to test time: while no method, including DMO+ERO, consistently achieves feasibility on the test set, DMO+ERO achieves the best feasibility-objective trade-off in most cases, reflected in its leading test $F_1$-scores both before and after TA. The only exception is FROP on \textit{MURA}, where DMO+ERO exhibits a larger train-test feasibility gap than elsewhere. We attribute this generalization failure to the train-test label distribution shift in \textit{MURA}. Outside this case, DMO+ERO's generalization gap remains consistently smaller than that of all competing baselines. 

\subsection{Ablation studies}\label{para:ab_exact}

\cref{table:ab_combined} isolates the contribution of each of ERO's three key components on FPOR and FROP, evaluated on \textit{MURA} and \textit{ADE-v2}. We compare the full ERO against three ablations: replacing the exact reformulation with a sigmoid smooth approximation (\textbf{SSA}) to the indicator function while keeping the same exact-penalty solver; removing the confidence regularization (\textbf{RCR}); and replacing our inequality reformulation with the equality (\textbf{EQ}) one. 
 
\textbf{Exact reformulation.} SSA remains feasible on $3$ of $4$ tasks, but its objective consistently trails ERO (e.g., FPOR/\textit{MURA} recall $0.233$ vs. $0.401$; test $F_1$ $0.287$ vs. $0.398$), confirming that the smooth surrogate sacrifices objective value even when it achieves feasibility. The one exception is FROP/\textit{ADE-v2}, where SSA slightly exceeds ERO in test $F_1$ ($0.807$ vs. $0.772$), likely because \textit{ADE-v2}'s milder imbalance makes the surrogate a reasonable approximation there.
 
\begin{figure}[!htbp]
    \centering
    \includegraphics[width=0.8\textwidth]{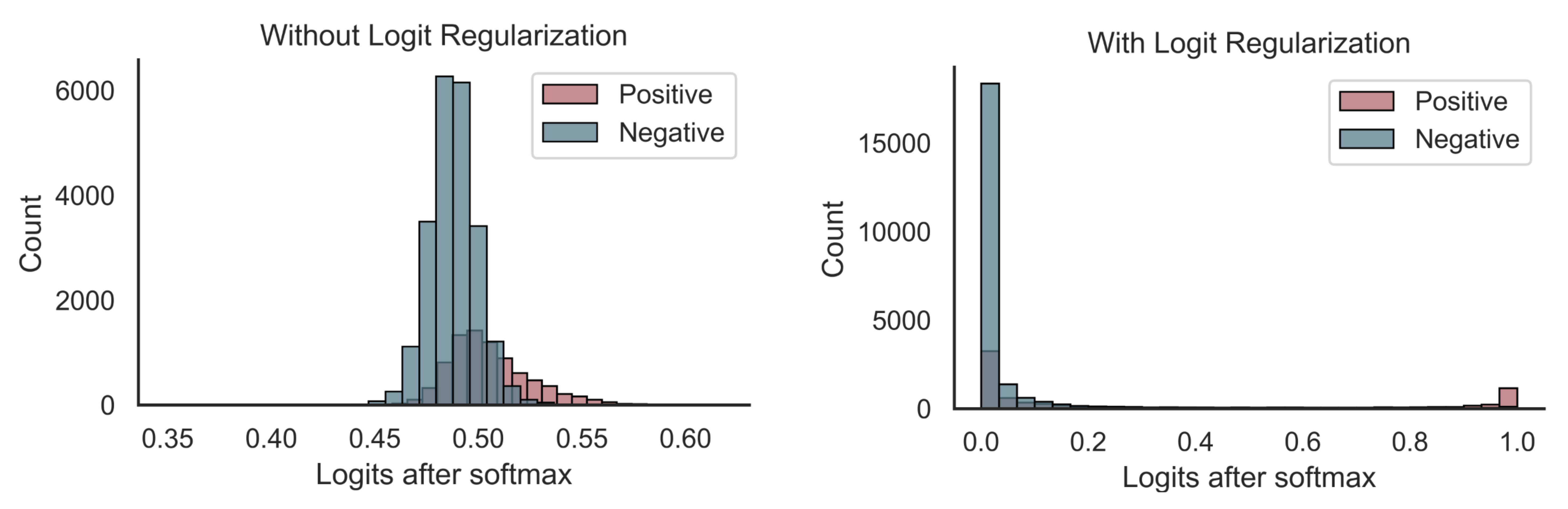}
    \caption{Histograms of the normalized prediction logits with and without the proposed logit regularization in \cref{eq:reg-term}. The task here is FPOR ($\text{recall} \ge 0.9$) on \textit{Eyepacs}.}
    \label{fig:ablation_reg}
    \vspace{-1em}
\end{figure}

\textbf{Confidence regularization.} The logit regularization in \cref{eq:reg-term} has been introduced to avoid the singular case of $f_{\mb \theta}(\mb x_i) = t$ for any $i$. Moreover, our analysis in \cref{sec:regularization} suggests that logit regularization tends to push logits $f_{\mb \theta}(\mb x_i)$ to take extreme values (i.e., $0$ or $1$). Our results unequivocally confirmed it: removing it causes the most severe degradation. RCR loses training feasibility on both FPOR tasks (e.g., precision $0.518$ on \textit{MURA}, with substantially higher variance than ERO). On FROP, RCR is either over-feasible which degrades the objective (e.g., on \textit{MURA}) or infeasible (e.g., on \textit{ADE-V2}). This is because the regularization term pushes the logits away from the decision threshold, thereby preventing optimization collapse (see \cref{fig:ablation_reg}). Collectively, confidence regularization is critical to reliably reaching the feasible region with optimized objectives.
 
\textbf{Inequality reformulation.} As expected (discussed in \cref{sec:eq_reform}), EQ's smaller feasible region makes optimization harder: it is feasible on $3$ of the $4$ tasks, of which $2$ are suboptimal (e.g., FPOR/\textit{MURA} recall $0.371$ vs. ERO's $0.401$) and $1$ marginally exceeds ERO, though within the variance interval; on the remaining task, FROP/\textit{MURA}, EQ is infeasible (train recall $0.836$). 
 
Across all three ablation studies and both FPOR \& FROP, ERO is the only configuration that achieves training feasibility in all four tasks while attaining almost optimal objectives, underscoring that the exact reformulation, the confidence regularization, and our specific inequality formulation provide complementary and non-redundant performance benefits. 

\begin{table*}[!htbp]
\def\arraystretch{1.3}
\centering

\caption{Ablation study on \textbf{FPOR} and \textbf{FROP}. For FPOR, feasible
solutions require $precision \ge 0.9$; for FROP, feasible solutions require
$recall \ge 0.9$. Feasible solutions are \underline{underlined}; best feasible
objective per dataset shown in \textcolor{red}{red}, and best $F_1$ in
\textcolor{red}{red}. Slack = 0.001.}
\label{table:ab_combined}

\smallskip

\resizebox{\textwidth}{!}{
\begin{tabular}{cl|cc|ccc|cc|ccc}
\Xhline{2\arrayrulewidth}

& &
\multicolumn{5}{c|}{\textbf{FPOR}} &
\multicolumn{5}{c}{\textbf{FROP}}
\\
\hline
\textbf{Dataset} & \textbf{Method}
& \multicolumn{2}{c|}{\textbf{Train}}
& \multicolumn{3}{c|}{\textbf{Test}}
& \multicolumn{2}{c|}{\textbf{Train}}
& \multicolumn{3}{c}{\textbf{Test}}
\\

&
& Precision & Recall
& Precision & Recall & $F_1$
& Precision & Recall
& Precision & Recall & $F_1$
\\

\hline

\multirow{4}{*}{MURA}
& SSA
& \underline{0.911 $\pm$ {\tiny 0.004}}
& 0.233 $\pm$ {\tiny 0.025}
& \underline{0.973 $\pm$ {\tiny 0.004}}
& \underline{0.169 $\pm$ {\tiny 0.018}}
& 0.287 $\pm$ {\tiny 0.027}
& \underline{0.945 $\pm$ {\tiny 0.026}}
& 0.093 $\pm$ {\tiny 0.026}
& 0.894 $\pm$ {\tiny 0.060}
& \underline{0.567 $\pm$ {\tiny 0.066}}
& \textcolor{red}{0.688 $\pm$ {\tiny 0.030}}
\\

& RCR
& 0.518 $\pm$ {\tiny 0.220}
& 0.337 $\pm$ {\tiny 0.257}
& 0.873 $\pm$ {\tiny 0.085}
& 0.298 $\pm$ {\tiny 0.204}
& 0.393 $\pm$ {\tiny 0.213}
& \underline{0.917 $\pm$ {\tiny 0.014}}
& 0.146 $\pm$ {\tiny 0.042}
& 0.721 $\pm$ {\tiny 0.062}
& 0.635 $\pm$ {\tiny 0.045}
& 0.671 $\pm$ {\tiny 0.009}
\\

& EQ
& \underline{0.912 $\pm$ {\tiny 0.008}}
& 0.371 $\pm$ {\tiny 0.043}
& \underline{0.951 $\pm$ {\tiny 0.011}}
& \underline{0.236 $\pm$ {\tiny 0.024}}
& 0.377 $\pm$ {\tiny 0.029}
& 0.836 $\pm$ {\tiny 0.052}
& 0.174 $\pm$ {\tiny 0.075}
& 0.711 $\pm$ {\tiny 0.105}
& 0.663 $\pm$ {\tiny 0.105}
& 0.670 $\pm$ {\tiny 0.020}
\\

& \cellcolor{gray}ERO
& \cellcolor{gray}\underline{0.905 $\pm$ {\tiny 0.005}}
& \cellcolor{gray}\textcolor{red}{0.401 $\pm$ {\tiny 0.050}}
& \cellcolor{gray}\underline{0.944 $\pm$ {\tiny 0.007}}
& \cellcolor{gray}\textcolor{red}{\underline{0.252 $\pm$ {\tiny 0.023}}}
& \cellcolor{gray}\textcolor{red}{0.398 $\pm$ {\tiny 0.029}}
& \cellcolor{gray}\underline{0.900 $\pm$ {\tiny 0.001}}
& \cellcolor{gray}\textcolor{red}{0.152 $\pm$ {\tiny 0.054}}
& \cellcolor{gray}0.753 $\pm$ {\tiny 0.066}
& \cellcolor{gray}0.651 $\pm$ {\tiny 0.090}
& \cellcolor{gray}\textcolor{red}{0.689 $\pm$ {\tiny 0.029}}
\\

\cline{1-12}

\multirow{4}{*}{ADE-v2}
& SSA
& \underline{0.908 $\pm$ {\tiny 0.004}}
& 0.772 $\pm$ {\tiny 0.006}
& 0.885 $\pm$ {\tiny 0.006}
& 0.732 $\pm$ {\tiny 0.007}
& 0.801 $\pm$ {\tiny 0.002}
& \underline{0.915 $\pm$ {\tiny 0.004}}
& \textcolor{red}{0.755 $\pm$ {\tiny 0.007}}
& \underline{0.891 $\pm$ {\tiny 0.005}}
& \textcolor{red}{\underline{0.738 $\pm$ {\tiny 0.007}}}
& \textcolor{red}{0.807 $\pm$ {\tiny 0.003}}
\\

& RCR
& 0.745 $\pm$ {\tiny 0.003}
& 0.884 $\pm$ {\tiny 0.014}
& 0.725 $\pm$ {\tiny 0.002}
& 0.868 $\pm$ {\tiny 0.013}
& 0.790 $\pm$ {\tiny 0.004}
& 0.884 $\pm$ {\tiny 0.009}
& 0.744 $\pm$ {\tiny 0.016}
& 0.858 $\pm$ {\tiny 0.008}
& 0.724 $\pm$ {\tiny 0.014}
& 0.785 $\pm$ {\tiny 0.009}
\\

& EQ
& \underline{0.900 $\pm$ {\tiny 0.000}}
& 0.781 $\pm$ {\tiny 0.013}
& 0.873 $\pm$ {\tiny 0.003}
& 0.746 $\pm$ {\tiny 0.016}
& 0.805 $\pm$ {\tiny 0.009}
& \underline{0.900 $\pm$ {\tiny 0.001}}
& 0.710 $\pm$ {\tiny 0.008}
& 0.883 $\pm$ {\tiny 0.002}
& 0.695 $\pm$ {\tiny 0.006}
& 0.777 $\pm$ {\tiny 0.004}
\\

& \cellcolor{gray}ERO
& \cellcolor{gray}\underline{0.900 $\pm$ {\tiny 0.001}}
& \cellcolor{gray}\textcolor{red}{0.803 $\pm$ {\tiny 0.010}}
& \cellcolor{gray}0.876 $\pm$ {\tiny 0.004}
& \cellcolor{gray}0.769 $\pm$ {\tiny 0.012}
& \cellcolor{gray}\textcolor{red}{0.819 $\pm$ {\tiny 0.007}}
& \cellcolor{gray}\underline{0.903 $\pm$ {\tiny 0.004}}
& \cellcolor{gray}0.697 $\pm$ {\tiny 0.010}
& \cellcolor{gray}0.886 $\pm$ {\tiny 0.004}
& \cellcolor{gray}0.684 $\pm$ {\tiny 0.008}
& \cellcolor{gray}0.772 $\pm$ {\tiny 0.004}
\\

\Xhline{2\arrayrulewidth}

\end{tabular}
}

\end{table*}



\subsection{Improving DMO+ERO generalization}
\label{app:ta-validation}
Here, we explore the simple idea of performing post-training finetuning of the decision threshold with respect to a \textbf{validation set} to improve generalization. Specifically, we divide each dataset into three splits---training, validation, and test. The validation set is held out during training and only used to finetune the decision threshold after training. 

The results are summarized in \cref{table:FPOR_ab_alpha} and \cref{table:FROP_ab_alpha}. We find that: 1) threshold tuning using a validation set does not degrade overall performance, and DMO+ERO remains the leading method. Although it does not guarantee training feasibility (particularly for FPOR), when none of the methods are feasible, DMO+ERO consistently finds solutions with the smallest feasibility violations. 2) it effectively improves generalization. DMO+ERO achieves feasible test solutions in 4 out of 12 tasks. Notably, the remaining infeasible cases are typically only marginally below the feasibility threshold. For example, allowing for a $1\%$ feasibility slackness increases the feasible solution to $10$ out of $12$ tasks.  

\begin{table*}[!htbp]
\def\arraystretch{1.3}
\centering
\caption{Three-split generalization study on \textcolor{red}{FPOR}. Values in (parentheses) are results after threshold adjustment (TA) on val. Feasible solutions ($\text{precision} \ge \alpha$) are \underline{underlined}, and among them the best objectives are shown in \textcolor{red}{red} (before TA) and \textcolor{blue}{blue} (after TA). Best $F_1$ scores are also highlighted. All underlines and highlights are up to $0.001$ slackness.}
\smallskip
\resizebox{1\textwidth}{!}{
\begin{tabular}{ll||cc||cc||cc:c}
\Xhline{2\arrayrulewidth}
& & \multicolumn{2}{c||}{\textbf{train}} & \multicolumn{2}{c||}{\textbf{val}} & \multicolumn{3}{c}{\textbf{test}} \\
dataset ($\alpha$) & method  & \multicolumn{1}{c}{precision---feasibility}  & \multicolumn{1}{c||}{recall---objective}  & \multicolumn{1}{c}{precision---feasibility}  & \multicolumn{1}{c||}{recall---objective}  & \multicolumn{1}{c}{precision---feasibility}  & \multicolumn{1}{c}{recall---objective} & F1-score \\
\hline

	& WCE	& 0.712 $\pm$ {\tiny 0.033} (0.751 $\pm$ {\tiny 0.262})	& 0.276 $\pm$ {\tiny 0.241} (0.099 $\pm$ {\tiny 0.165})	& 0.665 $\pm$ {\tiny 0.043} (0.840 $\pm$ {\tiny 0.118})	& 0.286 $\pm$ {\tiny 0.236} (0.110 $\pm$ {\tiny 0.169})	& 0.712 $\pm$ {\tiny 0.030} (0.547 $\pm$ {\tiny 0.362})	& 0.273 $\pm$ {\tiny 0.239} (0.094 $\pm$ {\tiny 0.159})	& 0.344 $\pm$ {\tiny 0.216} (0.136 $\pm$ {\tiny 0.201}) \\
	& Focal	& \underline{0.906 $\pm$ {\tiny 0.007}} (\underline{0.915 $\pm$ {\tiny 0.010}})	& 0.173 $\pm$ {\tiny 0.084} (0.116 $\pm$ {\tiny 0.103})	& 0.876 $\pm$ {\tiny 0.018} (\underline{0.911 $\pm$ {\tiny 0.013}})	& 0.164 $\pm$ {\tiny 0.084} (0.116 $\pm$ {\tiny 0.104})	& \underline{0.910 $\pm$ {\tiny 0.012}} (\underline{0.929 $\pm$ {\tiny 0.031}})	& \textcolor{red}{0.165 $\pm$ {\tiny 0.079}} (0.113 $\pm$ {\tiny 0.096})	& 0.271 $\pm$ {\tiny 0.115} (0.188 $\pm$ {\tiny 0.142}) \\
	& LDAM	& 0.808 $\pm$ {\tiny 0.004} (\underline{0.913 $\pm$ {\tiny 0.014}})	& 0.471 $\pm$ {\tiny 0.117} (\textcolor{blue}{0.221 $\pm$ {\tiny 0.122}})	& 0.799 $\pm$ {\tiny 0.016} (\underline{0.903 $\pm$ {\tiny 0.014}})	& 0.473 $\pm$ {\tiny 0.114} (0.224 $\pm$ {\tiny 0.123})	& 0.799 $\pm$ {\tiny 0.008} (\underline{0.922 $\pm$ {\tiny 0.030}})	& 0.461 $\pm$ {\tiny 0.114} (\textcolor{blue}{0.211 $\pm$ {\tiny 0.118}})	& 0.574 $\pm$ {\tiny 0.110} (0.324 $\pm$ {\tiny 0.173}) \\
	& TERM	& 0.762 $\pm$ {\tiny 0.008} (0.844 $\pm$ {\tiny 0.086})	& 0.330 $\pm$ {\tiny 0.262} (0.175 $\pm$ {\tiny 0.095})	& 0.724 $\pm$ {\tiny 0.027} (0.821 $\pm$ {\tiny 0.104})	& 0.335 $\pm$ {\tiny 0.260} (0.181 $\pm$ {\tiny 0.094})	& 0.760 $\pm$ {\tiny 0.014} (0.747 $\pm$ {\tiny 0.258})	& 0.320 $\pm$ {\tiny 0.257} (0.167 $\pm$ {\tiny 0.089})	& 0.390 $\pm$ {\tiny 0.244} (0.269 $\pm$ {\tiny 0.132}) \\
	& ROS	& 0.719 $\pm$ {\tiny 0.047} (\underline{0.905 $\pm$ {\tiny 0.097}})	& 0.450 $\pm$ {\tiny 0.346} (0.218 $\pm$ {\tiny 0.224})	& 0.689 $\pm$ {\tiny 0.037} (\underline{0.901 $\pm$ {\tiny 0.109}})	& 0.453 $\pm$ {\tiny 0.335} (0.223 $\pm$ {\tiny 0.225})	& 0.720 $\pm$ {\tiny 0.049} (0.707 $\pm$ {\tiny 0.365})	& 0.445 $\pm$ {\tiny 0.342} (0.207 $\pm$ {\tiny 0.214})	& 0.466 $\pm$ {\tiny 0.280} (0.284 $\pm$ {\tiny 0.274}) \\
	& RUS	& 0.692 $\pm$ {\tiny 0.023} (0.878 $\pm$ {\tiny 0.095})	& 0.447 $\pm$ {\tiny 0.304} (0.151 $\pm$ {\tiny 0.149})	& 0.670 $\pm$ {\tiny 0.025} (0.886 $\pm$ {\tiny 0.105})	& 0.456 $\pm$ {\tiny 0.299} (0.155 $\pm$ {\tiny 0.151})	& 0.700 $\pm$ {\tiny 0.025} (0.584 $\pm$ {\tiny 0.392})	& 0.442 $\pm$ {\tiny 0.298} (0.143 $\pm$ {\tiny 0.143})	& 0.476 $\pm$ {\tiny 0.233} (0.218 $\pm$ {\tiny 0.202}) \\
	& DMO+TFCO	& 0.359 $\pm$ {\tiny 0.050} (0.494 $\pm$ {\tiny 0.107})	& 0.543 $\pm$ {\tiny 0.330} (0.036 $\pm$ {\tiny 0.089})	& 0.357 $\pm$ {\tiny 0.060} (0.662 $\pm$ {\tiny 0.167})	& 0.546 $\pm$ {\tiny 0.336} (0.037 $\pm$ {\tiny 0.087})	& 0.337 $\pm$ {\tiny 0.063} (0.436 $\pm$ {\tiny 0.259})	& 0.535 $\pm$ {\tiny 0.330} (0.038 $\pm$ {\tiny 0.097})	& 0.355 $\pm$ {\tiny 0.148} (0.047 $\pm$ {\tiny 0.108}) \\
\rowcolor{gray}
\multirow{-8}{*}{\cellcolor{white}ADE V2 ($\alpha$=0.9)}	& DMO+ERO (ours)	& \underline{0.900 $\pm$ {\tiny 0.000}} (0.897 $\pm$ {\tiny 0.004})	& \textcolor{red}{0.795 $\pm$ {\tiny 0.011}} (0.563 $\pm$ {\tiny 0.036})	& 0.889 $\pm$ {\tiny 0.003} (\underline{0.901 $\pm$ {\tiny 0.001}})	& 0.773 $\pm$ {\tiny 0.007} (\textcolor{blue}{0.582 $\pm$ {\tiny 0.036}})	& 0.875 $\pm$ {\tiny 0.003} (0.890 $\pm$ {\tiny 0.010})	& 0.759 $\pm$ {\tiny 0.009} (0.541 $\pm$ {\tiny 0.034})	& \textcolor{red}{0.813 $\pm$ {\tiny 0.005}} (\textcolor{blue}{0.672 $\pm$ {\tiny 0.024}}) \\

\hline

	& WCE	& 0.416 $\pm$ {\tiny 0.071} (0.581 $\pm$ {\tiny 0.280})	& 0.120 $\pm$ {\tiny 0.133} (0.007 $\pm$ {\tiny 0.009})	& 0.371 $\pm$ {\tiny 0.088} (0.768 $\pm$ {\tiny 0.249})	& 0.117 $\pm$ {\tiny 0.129} (0.008 $\pm$ {\tiny 0.007})	& \underline{0.950 $\pm$ {\tiny 0.044}} (0.784 $\pm$ {\tiny 0.393})	& 0.118 $\pm$ {\tiny 0.127} (0.008 $\pm$ {\tiny 0.009})	& 0.185 $\pm$ {\tiny 0.189} (0.015 $\pm$ {\tiny 0.017}) \\
	& Focal	& 0.634 $\pm$ {\tiny 0.014} (0.722 $\pm$ {\tiny 0.177})	& 0.151 $\pm$ {\tiny 0.033} (0.033 $\pm$ {\tiny 0.043})	& 0.626 $\pm$ {\tiny 0.063} (0.877 $\pm$ {\tiny 0.157})	& 0.133 $\pm$ {\tiny 0.040} (0.033 $\pm$ {\tiny 0.045})	& \underline{0.962 $\pm$ {\tiny 0.013}} (0.888 $\pm$ {\tiny 0.296})	& 0.134 $\pm$ {\tiny 0.027} (0.032 $\pm$ {\tiny 0.046})	& 0.234 $\pm$ {\tiny 0.042} (0.058 $\pm$ {\tiny 0.080}) \\
	& LDAM	& 0.768 $\pm$ {\tiny 0.078} (0.679 $\pm$ {\tiny 0.185})	& 0.094 $\pm$ {\tiny 0.035} (0.016 $\pm$ {\tiny 0.015})	& 0.586 $\pm$ {\tiny 0.198} (0.891 $\pm$ {\tiny 0.093})	& 0.083 $\pm$ {\tiny 0.031} (0.018 $\pm$ {\tiny 0.018})	& \underline{0.980 $\pm$ {\tiny 0.008}} (\underline{0.927 $\pm$ {\tiny 0.144}})	& 0.082 $\pm$ {\tiny 0.031} (0.013 $\pm$ {\tiny 0.012})	& 0.149 $\pm$ {\tiny 0.056} (0.026 $\pm$ {\tiny 0.022}) \\
	& TERM	& 0.653 $\pm$ {\tiny 0.174} (0.552 $\pm$ {\tiny 0.277})	& 0.202 $\pm$ {\tiny 0.133} (0.035 $\pm$ {\tiny 0.044})	& 0.613 $\pm$ {\tiny 0.261} (0.785 $\pm$ {\tiny 0.204})	& 0.184 $\pm$ {\tiny 0.121} (0.040 $\pm$ {\tiny 0.046})	& 0.869 $\pm$ {\tiny 0.290} (\underline{0.979 $\pm$ {\tiny 0.027}})	& 0.185 $\pm$ {\tiny 0.122} (0.031 $\pm$ {\tiny 0.043})	& 0.290 $\pm$ {\tiny 0.189} (0.057 $\pm$ {\tiny 0.075}) \\
	& ROS	& 0.398 $\pm$ {\tiny 0.041} (0.544 $\pm$ {\tiny 0.296})	& 0.086 $\pm$ {\tiny 0.108} (0.025 $\pm$ {\tiny 0.040})	& 0.308 $\pm$ {\tiny 0.113} (0.669 $\pm$ {\tiny 0.253})	& 0.083 $\pm$ {\tiny 0.104} (0.029 $\pm$ {\tiny 0.042})	& \underline{0.918 $\pm$ {\tiny 0.090}} (\underline{0.967 $\pm$ {\tiny 0.046}})	& 0.083 $\pm$ {\tiny 0.103} (0.028 $\pm$ {\tiny 0.044})	& 0.135 $\pm$ {\tiny 0.160} (0.051 $\pm$ {\tiny 0.076}) \\
	& RUS	& 0.483 $\pm$ {\tiny 0.185} (0.650 $\pm$ {\tiny 0.328})	& 0.062 $\pm$ {\tiny 0.091} (0.030 $\pm$ {\tiny 0.045})	& 0.282 $\pm$ {\tiny 0.148} (0.746 $\pm$ {\tiny 0.328})	& 0.059 $\pm$ {\tiny 0.084} (0.033 $\pm$ {\tiny 0.048})	& 0.850 $\pm$ {\tiny 0.286} (\underline{0.955 $\pm$ {\tiny 0.085}})	& 0.062 $\pm$ {\tiny 0.087} (0.036 $\pm$ {\tiny 0.054})	& 0.105 $\pm$ {\tiny 0.129} (0.064 $\pm$ {\tiny 0.095}) \\
	& DMO+TFCO	& 0.066 $\pm$ {\tiny 0.005} (0.056 $\pm$ {\tiny 0.032})	& 0.610 $\pm$ {\tiny 0.439} (0.408 $\pm$ {\tiny 0.407})	& 0.076 $\pm$ {\tiny 0.007} (0.311 $\pm$ {\tiny 0.359})	& 0.611 $\pm$ {\tiny 0.437} (0.423 $\pm$ {\tiny 0.418})	& 0.446 $\pm$ {\tiny 0.041} (0.366 $\pm$ {\tiny 0.189})	& 0.587 $\pm$ {\tiny 0.445} (0.379 $\pm$ {\tiny 0.381})	& \textcolor{red}{0.412 $\pm$ {\tiny 0.274}} (\textcolor{blue}{0.290 $\pm$ {\tiny 0.286}}) \\
\rowcolor{gray}
\multirow{-8}{*}{\cellcolor{white}MURA ($\alpha$=0.9)}	& DMO+ERO (ours)	& 0.897 $\pm$ {\tiny 0.013} (0.824 $\pm$ {\tiny 0.051})	& 0.412 $\pm$ {\tiny 0.079} (0.085 $\pm$ {\tiny 0.036})	& 0.620 $\pm$ {\tiny 0.036} (0.820 $\pm$ {\tiny 0.066})	& 0.244 $\pm$ {\tiny 0.023} (0.080 $\pm$ {\tiny 0.028})	& \underline{0.942 $\pm$ {\tiny 0.010}} (\underline{0.966 $\pm$ {\tiny 0.014}})	& \textcolor{red}{0.251 $\pm$ {\tiny 0.033}} (\textcolor{blue}{0.084 $\pm$ {\tiny 0.031}})	& 0.396 $\pm$ {\tiny 0.041} (0.154 $\pm$ {\tiny 0.054}) \\

\hline

	& WCE	& 0.712 $\pm$ {\tiny 0.033} (0.757 $\pm$ {\tiny 0.266})	& 0.276 $\pm$ {\tiny 0.241} (0.073 $\pm$ {\tiny 0.094})	& 0.665 $\pm$ {\tiny 0.043} (0.845 $\pm$ {\tiny 0.121})	& 0.286 $\pm$ {\tiny 0.236} (0.083 $\pm$ {\tiny 0.100})	& 0.712 $\pm$ {\tiny 0.030} (0.553 $\pm$ {\tiny 0.369})	& 0.273 $\pm$ {\tiny 0.239} (0.067 $\pm$ {\tiny 0.088})	& 0.344 $\pm$ {\tiny 0.216} (0.111 $\pm$ {\tiny 0.139}) \\
	& Focal	& 0.906 $\pm$ {\tiny 0.007} (0.949 $\pm$ {\tiny 0.025})	& 0.173 $\pm$ {\tiny 0.084} (0.022 $\pm$ {\tiny 0.009})	& 0.876 $\pm$ {\tiny 0.018} (\underline{0.974 $\pm$ {\tiny 0.022}})	& 0.164 $\pm$ {\tiny 0.084} (0.028 $\pm$ {\tiny 0.010})	& 0.910 $\pm$ {\tiny 0.012} (0.873 $\pm$ {\tiny 0.293})	& 0.165 $\pm$ {\tiny 0.079} (0.021 $\pm$ {\tiny 0.010})	& 0.271 $\pm$ {\tiny 0.115} (0.042 $\pm$ {\tiny 0.019}) \\
	& LDAM	& 0.808 $\pm$ {\tiny 0.004} (0.948 $\pm$ {\tiny 0.019})	& 0.471 $\pm$ {\tiny 0.117} (0.022 $\pm$ {\tiny 0.010})	& 0.799 $\pm$ {\tiny 0.016} (\underline{0.957 $\pm$ {\tiny 0.030}})	& 0.473 $\pm$ {\tiny 0.114} (0.033 $\pm$ {\tiny 0.014})	& 0.799 $\pm$ {\tiny 0.008} (0.878 $\pm$ {\tiny 0.293})	& 0.461 $\pm$ {\tiny 0.114} (0.022 $\pm$ {\tiny 0.011})	& 0.574 $\pm$ {\tiny 0.110} (0.042 $\pm$ {\tiny 0.022}) \\
	& TERM	& 0.762 $\pm$ {\tiny 0.008} (0.855 $\pm$ {\tiny 0.094})	& 0.330 $\pm$ {\tiny 0.262} (0.071 $\pm$ {\tiny 0.056})	& 0.724 $\pm$ {\tiny 0.027} (0.843 $\pm$ {\tiny 0.123})	& 0.335 $\pm$ {\tiny 0.260} (0.081 $\pm$ {\tiny 0.057})	& 0.760 $\pm$ {\tiny 0.014} (0.771 $\pm$ {\tiny 0.274})	& 0.320 $\pm$ {\tiny 0.257} (0.070 $\pm$ {\tiny 0.055})	& 0.390 $\pm$ {\tiny 0.244} (0.123 $\pm$ {\tiny 0.092}) \\
	& ROS	& 0.719 $\pm$ {\tiny 0.047} (0.922 $\pm$ {\tiny 0.101})	& 0.450 $\pm$ {\tiny 0.346} (0.123 $\pm$ {\tiny 0.129})	& 0.689 $\pm$ {\tiny 0.037} (0.922 $\pm$ {\tiny 0.112})	& 0.453 $\pm$ {\tiny 0.335} (0.133 $\pm$ {\tiny 0.135})	& 0.720 $\pm$ {\tiny 0.049} (0.727 $\pm$ {\tiny 0.376})	& 0.445 $\pm$ {\tiny 0.342} (0.115 $\pm$ {\tiny 0.121})	& 0.466 $\pm$ {\tiny 0.280} (0.182 $\pm$ {\tiny 0.187}) \\
	& RUS	& 0.692 $\pm$ {\tiny 0.023} (0.890 $\pm$ {\tiny 0.101})	& 0.447 $\pm$ {\tiny 0.304} (0.071 $\pm$ {\tiny 0.096})	& 0.670 $\pm$ {\tiny 0.025} (0.908 $\pm$ {\tiny 0.111})	& 0.456 $\pm$ {\tiny 0.299} (0.082 $\pm$ {\tiny 0.104})	& 0.700 $\pm$ {\tiny 0.025} (0.599 $\pm$ {\tiny 0.405})	& 0.442 $\pm$ {\tiny 0.298} (0.069 $\pm$ {\tiny 0.093})	& 0.476 $\pm$ {\tiny 0.233} (0.113 $\pm$ {\tiny 0.145}) \\
	& DMO+TFCO	& 0.369 $\pm$ {\tiny 0.076} (0.506 $\pm$ {\tiny 0.173})	& 0.499 $\pm$ {\tiny 0.437} (0.037 $\pm$ {\tiny 0.059})	& 0.320 $\pm$ {\tiny 0.101} (0.645 $\pm$ {\tiny 0.300})	& 0.500 $\pm$ {\tiny 0.437} (0.035 $\pm$ {\tiny 0.053})	& 0.349 $\pm$ {\tiny 0.091} (0.366 $\pm$ {\tiny 0.212})	& 0.499 $\pm$ {\tiny 0.439} (0.036 $\pm$ {\tiny 0.057})	& 0.286 $\pm$ {\tiny 0.208} (0.056 $\pm$ {\tiny 0.083}) \\
\rowcolor{gray}
\multirow{-8}{*}{\cellcolor{white}ADE V2 ($\alpha$=0.95)}	& DMO+ERO (ours)	& 0.947 $\pm$ {\tiny 0.003} (\underline{0.961 $\pm$ {\tiny 0.005}})	& 0.755 $\pm$ {\tiny 0.007} (\textcolor{blue}{0.231 $\pm$ {\tiny 0.031}})	& 0.915 $\pm$ {\tiny 0.002} (\underline{0.949 $\pm$ {\tiny 0.004}})	& 0.741 $\pm$ {\tiny 0.006} (\textcolor{blue}{0.231 $\pm$ {\tiny 0.037}})	& 0.919 $\pm$ {\tiny 0.004} (\underline{0.965 $\pm$ {\tiny 0.006}})	& 0.709 $\pm$ {\tiny 0.006} (\textcolor{blue}{0.218 $\pm$ {\tiny 0.029}})	& \textcolor{red}{0.800 $\pm$ {\tiny 0.004}} (\textcolor{blue}{0.354 $\pm$ {\tiny 0.038}}) \\

\hline

	& WCE	& 0.416 $\pm$ {\tiny 0.071} (0.581 $\pm$ {\tiny 0.280})	& 0.120 $\pm$ {\tiny 0.133} (0.007 $\pm$ {\tiny 0.009})	& 0.371 $\pm$ {\tiny 0.088} (0.768 $\pm$ {\tiny 0.249})	& 0.117 $\pm$ {\tiny 0.129} (0.008 $\pm$ {\tiny 0.007})	& \underline{0.950 $\pm$ {\tiny 0.044}} (0.784 $\pm$ {\tiny 0.393})	& 0.118 $\pm$ {\tiny 0.127} (0.008 $\pm$ {\tiny 0.009})	& 0.185 $\pm$ {\tiny 0.189} (0.015 $\pm$ {\tiny 0.017}) \\
	& Focal	& 0.634 $\pm$ {\tiny 0.014} (0.722 $\pm$ {\tiny 0.177})	& 0.151 $\pm$ {\tiny 0.033} (0.033 $\pm$ {\tiny 0.043})	& 0.626 $\pm$ {\tiny 0.063} (0.877 $\pm$ {\tiny 0.157})	& 0.133 $\pm$ {\tiny 0.040} (0.033 $\pm$ {\tiny 0.045})	& \underline{0.962 $\pm$ {\tiny 0.013}} (0.888 $\pm$ {\tiny 0.296})	& \textcolor{red}{0.134 $\pm$ {\tiny 0.027}} (0.032 $\pm$ {\tiny 0.046})	& 0.234 $\pm$ {\tiny 0.042} (0.058 $\pm$ {\tiny 0.080}) \\
	& LDAM	& 0.768 $\pm$ {\tiny 0.078} (0.679 $\pm$ {\tiny 0.185})	& 0.094 $\pm$ {\tiny 0.035} (0.016 $\pm$ {\tiny 0.015})	& 0.586 $\pm$ {\tiny 0.198} (0.891 $\pm$ {\tiny 0.093})	& 0.083 $\pm$ {\tiny 0.031} (0.018 $\pm$ {\tiny 0.018})	& \underline{0.980 $\pm$ {\tiny 0.008}} (0.927 $\pm$ {\tiny 0.144})	& 0.082 $\pm$ {\tiny 0.031} (0.013 $\pm$ {\tiny 0.012})	& 0.149 $\pm$ {\tiny 0.056} (0.026 $\pm$ {\tiny 0.022}) \\
	& TERM	& 0.653 $\pm$ {\tiny 0.174} (0.552 $\pm$ {\tiny 0.277})	& 0.202 $\pm$ {\tiny 0.133} (0.035 $\pm$ {\tiny 0.044})	& 0.613 $\pm$ {\tiny 0.261} (0.785 $\pm$ {\tiny 0.204})	& 0.184 $\pm$ {\tiny 0.121} (0.040 $\pm$ {\tiny 0.046})	& 0.869 $\pm$ {\tiny 0.290} (\underline{0.979 $\pm$ {\tiny 0.027}})	& 0.185 $\pm$ {\tiny 0.122} (0.031 $\pm$ {\tiny 0.043})	& 0.290 $\pm$ {\tiny 0.189} (0.057 $\pm$ {\tiny 0.075}) \\
	& ROS	& 0.398 $\pm$ {\tiny 0.041} (0.544 $\pm$ {\tiny 0.296})	& 0.086 $\pm$ {\tiny 0.108} (0.025 $\pm$ {\tiny 0.040})	& 0.308 $\pm$ {\tiny 0.113} (0.669 $\pm$ {\tiny 0.253})	& 0.083 $\pm$ {\tiny 0.104} (0.029 $\pm$ {\tiny 0.042})	& 0.918 $\pm$ {\tiny 0.090} (\underline{0.967 $\pm$ {\tiny 0.046}})	& 0.083 $\pm$ {\tiny 0.103} (0.028 $\pm$ {\tiny 0.044})	& 0.135 $\pm$ {\tiny 0.160} (0.051 $\pm$ {\tiny 0.076}) \\
	& RUS	& 0.483 $\pm$ {\tiny 0.185} (0.650 $\pm$ {\tiny 0.328})	& 0.062 $\pm$ {\tiny 0.091} (0.030 $\pm$ {\tiny 0.045})	& 0.282 $\pm$ {\tiny 0.148} (0.746 $\pm$ {\tiny 0.328})	& 0.059 $\pm$ {\tiny 0.084} (0.033 $\pm$ {\tiny 0.048})	& 0.850 $\pm$ {\tiny 0.286} (\underline{0.955 $\pm$ {\tiny 0.085}})	& 0.062 $\pm$ {\tiny 0.087} (\textcolor{blue}{0.036 $\pm$ {\tiny 0.054}})	& 0.105 $\pm$ {\tiny 0.129} (0.064 $\pm$ {\tiny 0.095}) \\
	& DMO+TFCO	& 0.070 $\pm$ {\tiny 0.011} (0.048 $\pm$ {\tiny 0.024})	& 0.534 $\pm$ {\tiny 0.442} (0.337 $\pm$ {\tiny 0.413})	& 0.083 $\pm$ {\tiny 0.024} (0.238 $\pm$ {\tiny 0.283})	& 0.533 $\pm$ {\tiny 0.443} (0.348 $\pm$ {\tiny 0.422})	& 0.457 $\pm$ {\tiny 0.037} (0.479 $\pm$ {\tiny 0.071})	& 0.509 $\pm$ {\tiny 0.440} (0.317 $\pm$ {\tiny 0.393})	& 0.366 $\pm$ {\tiny 0.283} (\textcolor{blue}{0.237 $\pm$ {\tiny 0.285}}) \\
\rowcolor{gray}
\multirow{-8}{*}{\cellcolor{white}MURA ($\alpha$=0.95)}	& DMO+ERO (ours)	& \underline{0.951 $\pm$ {\tiny 0.003}} (0.909 $\pm$ {\tiny 0.061})	& \textcolor{red}{0.398 $\pm$ {\tiny 0.077}} (0.056 $\pm$ {\tiny 0.040})	& 0.646 $\pm$ {\tiny 0.028} (0.918 $\pm$ {\tiny 0.056})	& 0.224 $\pm$ {\tiny 0.027} (0.046 $\pm$ {\tiny 0.029})	& 0.948 $\pm$ {\tiny 0.009} (0.934 $\pm$ {\tiny 0.061})	& 0.229 $\pm$ {\tiny 0.031} (0.056 $\pm$ {\tiny 0.035})	& \textcolor{red}{0.368 $\pm$ {\tiny 0.041}} (0.103 $\pm$ {\tiny 0.064}) \\

\hline

	& WCE	& 0.712 $\pm$ {\tiny 0.033} (0.758 $\pm$ {\tiny 0.267})	& 0.276 $\pm$ {\tiny 0.241} (0.044 $\pm$ {\tiny 0.054})	& 0.665 $\pm$ {\tiny 0.043} (0.850 $\pm$ {\tiny 0.126})	& 0.286 $\pm$ {\tiny 0.236} (0.054 $\pm$ {\tiny 0.064})	& 0.712 $\pm$ {\tiny 0.030} (0.554 $\pm$ {\tiny 0.370})	& 0.273 $\pm$ {\tiny 0.239} (0.042 $\pm$ {\tiny 0.054})	& 0.344 $\pm$ {\tiny 0.216} (0.073 $\pm$ {\tiny 0.090}) \\
	& Focal	& 0.906 $\pm$ {\tiny 0.007} (0.964 $\pm$ {\tiny 0.018})	& 0.173 $\pm$ {\tiny 0.084} (0.017 $\pm$ {\tiny 0.008})	& 0.876 $\pm$ {\tiny 0.018} (\underline{1.000 $\pm$ {\tiny 0.000}})	& 0.164 $\pm$ {\tiny 0.084} (\textcolor{blue}{0.022 $\pm$ {\tiny 0.009}})	& 0.910 $\pm$ {\tiny 0.012} (0.878 $\pm$ {\tiny 0.294})	& 0.165 $\pm$ {\tiny 0.079} (0.017 $\pm$ {\tiny 0.009})	& 0.271 $\pm$ {\tiny 0.115} (0.033 $\pm$ {\tiny 0.018}) \\
	& LDAM	& 0.808 $\pm$ {\tiny 0.004} (0.964 $\pm$ {\tiny 0.022})	& 0.471 $\pm$ {\tiny 0.117} (0.009 $\pm$ {\tiny 0.005})	& 0.799 $\pm$ {\tiny 0.016} (0.988 $\pm$ {\tiny 0.037})	& 0.473 $\pm$ {\tiny 0.114} (0.016 $\pm$ {\tiny 0.009})	& 0.799 $\pm$ {\tiny 0.008} (0.796 $\pm$ {\tiny 0.398})	& 0.461 $\pm$ {\tiny 0.114} (0.008 $\pm$ {\tiny 0.006})	& 0.574 $\pm$ {\tiny 0.110} (0.017 $\pm$ {\tiny 0.012}) \\
	& TERM	& 0.762 $\pm$ {\tiny 0.008} (0.861 $\pm$ {\tiny 0.099})	& 0.330 $\pm$ {\tiny 0.262} (0.069 $\pm$ {\tiny 0.057})	& 0.724 $\pm$ {\tiny 0.027} (0.860 $\pm$ {\tiny 0.140})	& 0.335 $\pm$ {\tiny 0.260} (0.079 $\pm$ {\tiny 0.059})	& 0.760 $\pm$ {\tiny 0.014} (0.776 $\pm$ {\tiny 0.278})	& 0.320 $\pm$ {\tiny 0.257} (0.069 $\pm$ {\tiny 0.057})	& 0.390 $\pm$ {\tiny 0.244} (0.120 $\pm$ {\tiny 0.095}) \\
	& ROS	& 0.719 $\pm$ {\tiny 0.047} (0.932 $\pm$ {\tiny 0.105})	& 0.450 $\pm$ {\tiny 0.346} (0.036 $\pm$ {\tiny 0.054})	& 0.689 $\pm$ {\tiny 0.037} (0.940 $\pm$ {\tiny 0.119})	& 0.453 $\pm$ {\tiny 0.335} (0.038 $\pm$ {\tiny 0.058})	& 0.720 $\pm$ {\tiny 0.049} (0.735 $\pm$ {\tiny 0.382})	& 0.445 $\pm$ {\tiny 0.342} (0.035 $\pm$ {\tiny 0.051})	& 0.466 $\pm$ {\tiny 0.280} (0.062 $\pm$ {\tiny 0.085}) \\
	& RUS	& 0.692 $\pm$ {\tiny 0.023} (0.891 $\pm$ {\tiny 0.101})	& 0.447 $\pm$ {\tiny 0.304} (0.040 $\pm$ {\tiny 0.065})	& 0.670 $\pm$ {\tiny 0.025} (0.924 $\pm$ {\tiny 0.121})	& 0.456 $\pm$ {\tiny 0.299} (0.046 $\pm$ {\tiny 0.072})	& 0.700 $\pm$ {\tiny 0.025} (0.608 $\pm$ {\tiny 0.413})	& 0.442 $\pm$ {\tiny 0.298} (0.039 $\pm$ {\tiny 0.065})	& 0.476 $\pm$ {\tiny 0.233} (0.065 $\pm$ {\tiny 0.105}) \\
	& DMO+TFCO	& 0.385 $\pm$ {\tiny 0.121} (0.414 $\pm$ {\tiny 0.204})	& 0.471 $\pm$ {\tiny 0.389} (0.064 $\pm$ {\tiny 0.097})	& 0.375 $\pm$ {\tiny 0.106} (0.675 $\pm$ {\tiny 0.258})	& 0.477 $\pm$ {\tiny 0.385} (0.066 $\pm$ {\tiny 0.098})	& 0.389 $\pm$ {\tiny 0.119} (0.336 $\pm$ {\tiny 0.253})	& 0.469 $\pm$ {\tiny 0.389} (0.061 $\pm$ {\tiny 0.095})	& 0.295 $\pm$ {\tiny 0.178} (0.079 $\pm$ {\tiny 0.115}) \\
\rowcolor{gray}
\multirow{-8}{*}{\cellcolor{white}ADE V2 ($\alpha$=0.99)}	& DMO+ERO (ours)	& 0.955 $\pm$ {\tiny 0.004} (0.978 $\pm$ {\tiny 0.006})	& 0.691 $\pm$ {\tiny 0.009} (0.139 $\pm$ {\tiny 0.054})	& 0.924 $\pm$ {\tiny 0.007} (0.979 $\pm$ {\tiny 0.005})	& 0.691 $\pm$ {\tiny 0.008} (0.136 $\pm$ {\tiny 0.055})	& 0.937 $\pm$ {\tiny 0.008} (0.980 $\pm$ {\tiny 0.010})	& 0.643 $\pm$ {\tiny 0.006} (0.132 $\pm$ {\tiny 0.051})	& \textcolor{red}{0.763 $\pm$ {\tiny 0.005}} (\textcolor{blue}{0.230 $\pm$ {\tiny 0.079}}) \\

\hline

	& WCE	& 0.416 $\pm$ {\tiny 0.071} (0.581 $\pm$ {\tiny 0.280})	& 0.120 $\pm$ {\tiny 0.133} (0.007 $\pm$ {\tiny 0.009})	& 0.371 $\pm$ {\tiny 0.088} (0.768 $\pm$ {\tiny 0.249})	& 0.117 $\pm$ {\tiny 0.129} (0.008 $\pm$ {\tiny 0.007})	& 0.950 $\pm$ {\tiny 0.044} (0.784 $\pm$ {\tiny 0.393})	& 0.118 $\pm$ {\tiny 0.127} (0.008 $\pm$ {\tiny 0.009})	& 0.185 $\pm$ {\tiny 0.189} (0.015 $\pm$ {\tiny 0.017}) \\
	& Focal	& 0.634 $\pm$ {\tiny 0.014} (0.722 $\pm$ {\tiny 0.177})	& 0.151 $\pm$ {\tiny 0.033} (0.033 $\pm$ {\tiny 0.043})	& 0.626 $\pm$ {\tiny 0.063} (0.877 $\pm$ {\tiny 0.157})	& 0.133 $\pm$ {\tiny 0.040} (0.033 $\pm$ {\tiny 0.045})	& 0.962 $\pm$ {\tiny 0.013} (0.888 $\pm$ {\tiny 0.296})	& 0.134 $\pm$ {\tiny 0.027} (0.032 $\pm$ {\tiny 0.046})	& 0.234 $\pm$ {\tiny 0.042} (0.058 $\pm$ {\tiny 0.080}) \\
	& LDAM	& 0.768 $\pm$ {\tiny 0.078} (0.679 $\pm$ {\tiny 0.185})	& 0.094 $\pm$ {\tiny 0.035} (0.016 $\pm$ {\tiny 0.015})	& 0.586 $\pm$ {\tiny 0.198} (0.891 $\pm$ {\tiny 0.093})	& 0.083 $\pm$ {\tiny 0.031} (0.018 $\pm$ {\tiny 0.018})	& 0.980 $\pm$ {\tiny 0.008} (0.927 $\pm$ {\tiny 0.144})	& 0.082 $\pm$ {\tiny 0.031} (0.013 $\pm$ {\tiny 0.012})	& 0.149 $\pm$ {\tiny 0.056} (0.026 $\pm$ {\tiny 0.022}) \\
	& TERM	& 0.653 $\pm$ {\tiny 0.174} (0.552 $\pm$ {\tiny 0.277})	& 0.202 $\pm$ {\tiny 0.133} (0.035 $\pm$ {\tiny 0.044})	& 0.613 $\pm$ {\tiny 0.261} (0.785 $\pm$ {\tiny 0.204})	& 0.184 $\pm$ {\tiny 0.121} (0.040 $\pm$ {\tiny 0.046})	& 0.869 $\pm$ {\tiny 0.290} (0.979 $\pm$ {\tiny 0.027})	& 0.185 $\pm$ {\tiny 0.122} (0.031 $\pm$ {\tiny 0.043})	& 0.290 $\pm$ {\tiny 0.189} (0.057 $\pm$ {\tiny 0.075}) \\
	& ROS	& 0.398 $\pm$ {\tiny 0.041} (0.544 $\pm$ {\tiny 0.296})	& 0.086 $\pm$ {\tiny 0.108} (0.025 $\pm$ {\tiny 0.040})	& 0.308 $\pm$ {\tiny 0.113} (0.669 $\pm$ {\tiny 0.253})	& 0.083 $\pm$ {\tiny 0.104} (0.029 $\pm$ {\tiny 0.042})	& 0.918 $\pm$ {\tiny 0.090} (0.967 $\pm$ {\tiny 0.046})	& 0.083 $\pm$ {\tiny 0.103} (0.028 $\pm$ {\tiny 0.044})	& 0.135 $\pm$ {\tiny 0.160} (0.051 $\pm$ {\tiny 0.076}) \\
	& RUS	& 0.483 $\pm$ {\tiny 0.185} (0.650 $\pm$ {\tiny 0.328})	& 0.062 $\pm$ {\tiny 0.091} (0.030 $\pm$ {\tiny 0.045})	& 0.282 $\pm$ {\tiny 0.148} (0.746 $\pm$ {\tiny 0.328})	& 0.059 $\pm$ {\tiny 0.084} (0.033 $\pm$ {\tiny 0.048})	& 0.850 $\pm$ {\tiny 0.286} (0.955 $\pm$ {\tiny 0.085})	& 0.062 $\pm$ {\tiny 0.087} (0.036 $\pm$ {\tiny 0.054})	& 0.105 $\pm$ {\tiny 0.129} (0.064 $\pm$ {\tiny 0.095}) \\
	& DMO+TFCO	& 0.063 $\pm$ {\tiny 0.001} (0.065 $\pm$ {\tiny 0.020})	& 0.762 $\pm$ {\tiny 0.332} (0.417 $\pm$ {\tiny 0.415})	& 0.070 $\pm$ {\tiny 0.001} (0.119 $\pm$ {\tiny 0.079})	& 0.760 $\pm$ {\tiny 0.335} (0.430 $\pm$ {\tiny 0.424})	& 0.466 $\pm$ {\tiny 0.018} (0.477 $\pm$ {\tiny 0.138})	& 0.733 $\pm$ {\tiny 0.346} (0.389 $\pm$ {\tiny 0.394})	& \textcolor{red}{0.522 $\pm$ {\tiny 0.197}} (\textcolor{blue}{0.294 $\pm$ {\tiny 0.286}}) \\
\rowcolor{gray}
\multirow{-8}{*}{\cellcolor{white}MURA ($\alpha$=0.99)}	& DMO+ERO (ours)	& 0.986 $\pm$ {\tiny 0.010} (0.958 $\pm$ {\tiny 0.026})	& 0.356 $\pm$ {\tiny 0.066} (0.049 $\pm$ {\tiny 0.041})	& 0.691 $\pm$ {\tiny 0.032} (0.926 $\pm$ {\tiny 0.067})	& 0.202 $\pm$ {\tiny 0.026} (0.043 $\pm$ {\tiny 0.034})	& 0.956 $\pm$ {\tiny 0.012} (0.961 $\pm$ {\tiny 0.025})	& 0.205 $\pm$ {\tiny 0.032} (0.047 $\pm$ {\tiny 0.036})	& 0.336 $\pm$ {\tiny 0.044} (0.087 $\pm$ {\tiny 0.064}) \\
\hline
\Xhline{3\arrayrulewidth}
\end{tabular}
}
\label{table:three_split_FPOR}
\end{table*}

\begin{table*}[!htbp]
\def\arraystretch{1.3}
\centering
\caption{Three-split generalization study on \textcolor{red}{FROP}. Values in (parentheses) are results after threshold adjustment (TA) on val. Feasible solutions ($\text{recall} \ge \alpha$) are \underline{underlined}, and among them the best objectives are shown in \textcolor{red}{red} (before TA) and \textcolor{blue}{blue} (after TA). Best $F_1$ scores are also highlighted. All underlines and highlights are up to $0.001$ slackness.}
\smallskip
\resizebox{1\textwidth}{!}{
\begin{tabular}{ll||cc||cc||cc:c}
\Xhline{2\arrayrulewidth}
& & \multicolumn{2}{c||}{\textbf{train}} & \multicolumn{2}{c||}{\textbf{val}} & \multicolumn{3}{c}{\textbf{test}} \\
dataset ($\alpha$) & method  & \multicolumn{1}{c}{recall---feasibility}  & \multicolumn{1}{c||}{precision---objective}  & \multicolumn{1}{c}{recall---feasibility}  & \multicolumn{1}{c||}{precision---objective}  & \multicolumn{1}{c}{recall---feasibility}  & \multicolumn{1}{c}{precision---objective} & F1-score \\
\hline

	& WCE	& \underline{0.910 $\pm$ {\tiny 0.010}} (\underline{0.907 $\pm$ {\tiny 0.003}})	& 0.434 $\pm$ {\tiny 0.020} (0.438 $\pm$ {\tiny 0.019})	& \underline{0.904 $\pm$ {\tiny 0.010}} (\underline{0.901 $\pm$ {\tiny 0.001}})	& 0.426 $\pm$ {\tiny 0.021} (0.430 $\pm$ {\tiny 0.019})	& 0.896 $\pm$ {\tiny 0.010} (0.893 $\pm$ {\tiny 0.004})	& 0.429 $\pm$ {\tiny 0.021} (0.433 $\pm$ {\tiny 0.018})	& 0.580 $\pm$ {\tiny 0.018} (0.583 $\pm$ {\tiny 0.016}) \\
	& Focal	& \underline{0.915 $\pm$ {\tiny 0.011}} (\underline{0.901 $\pm$ {\tiny 0.005}})	& 0.515 $\pm$ {\tiny 0.016} (0.533 $\pm$ {\tiny 0.010})	& \underline{0.914 $\pm$ {\tiny 0.008}} (\underline{0.901 $\pm$ {\tiny 0.000}})	& \textcolor{red}{0.513 $\pm$ {\tiny 0.017}} (0.530 $\pm$ {\tiny 0.012})	& \underline{0.904 $\pm$ {\tiny 0.012}} (0.887 $\pm$ {\tiny 0.006})	& \textcolor{red}{0.511 $\pm$ {\tiny 0.016}} (0.528 $\pm$ {\tiny 0.010})	& 0.653 $\pm$ {\tiny 0.011} (0.662 $\pm$ {\tiny 0.008}) \\
	& LDAM	& 0.777 $\pm$ {\tiny 0.010} (\underline{0.908 $\pm$ {\tiny 0.005}})	& 0.522 $\pm$ {\tiny 0.106} (0.440 $\pm$ {\tiny 0.084})	& 0.766 $\pm$ {\tiny 0.012} (\underline{0.901 $\pm$ {\tiny 0.002}})	& 0.507 $\pm$ {\tiny 0.108} (0.431 $\pm$ {\tiny 0.083})	& 0.762 $\pm$ {\tiny 0.010} (0.894 $\pm$ {\tiny 0.005})	& 0.522 $\pm$ {\tiny 0.100} (0.434 $\pm$ {\tiny 0.081})	& 0.615 $\pm$ {\tiny 0.061} (0.581 $\pm$ {\tiny 0.064}) \\
	& TERM	& \underline{0.955 $\pm$ {\tiny 0.026}} (\underline{0.905 $\pm$ {\tiny 0.005}})	& 0.358 $\pm$ {\tiny 0.036} (0.400 $\pm$ {\tiny 0.017})	& \underline{0.955 $\pm$ {\tiny 0.029}} (\underline{0.901 $\pm$ {\tiny 0.001}})	& 0.353 $\pm$ {\tiny 0.035} (0.392 $\pm$ {\tiny 0.016})	& \underline{0.947 $\pm$ {\tiny 0.030}} (0.893 $\pm$ {\tiny 0.005})	& 0.356 $\pm$ {\tiny 0.036} (0.395 $\pm$ {\tiny 0.017})	& 0.515 $\pm$ {\tiny 0.035} (0.548 $\pm$ {\tiny 0.016}) \\
	& ROS	& \underline{0.930 $\pm$ {\tiny 0.028}} (\underline{0.907 $\pm$ {\tiny 0.006}})	& 0.398 $\pm$ {\tiny 0.033} (0.420 $\pm$ {\tiny 0.016})	& \underline{0.926 $\pm$ {\tiny 0.030}} (\underline{0.901 $\pm$ {\tiny 0.001}})	& 0.390 $\pm$ {\tiny 0.032} (0.412 $\pm$ {\tiny 0.017})	& \underline{0.919 $\pm$ {\tiny 0.030}} (0.892 $\pm$ {\tiny 0.005})	& 0.393 $\pm$ {\tiny 0.033} (0.415 $\pm$ {\tiny 0.017})	& 0.549 $\pm$ {\tiny 0.029} (0.566 $\pm$ {\tiny 0.016}) \\
	& RUS	& \underline{0.935 $\pm$ {\tiny 0.032}} (\underline{0.907 $\pm$ {\tiny 0.005}})	& 0.401 $\pm$ {\tiny 0.048} (0.425 $\pm$ {\tiny 0.020})	& \underline{0.930 $\pm$ {\tiny 0.036}} (\underline{0.901 $\pm$ {\tiny 0.000}})	& 0.393 $\pm$ {\tiny 0.046} (0.417 $\pm$ {\tiny 0.020})	& \underline{0.924 $\pm$ {\tiny 0.036}} (0.893 $\pm$ {\tiny 0.005})	& 0.397 $\pm$ {\tiny 0.047} (0.420 $\pm$ {\tiny 0.020})	& 0.553 $\pm$ {\tiny 0.042} (0.571 $\pm$ {\tiny 0.019}) \\
	& DMO+TFCO	& \underline{0.928 $\pm$ {\tiny 0.020}} (\underline{0.904 $\pm$ {\tiny 0.006}})	& 0.389 $\pm$ {\tiny 0.018} (0.408 $\pm$ {\tiny 0.011})	& \underline{0.926 $\pm$ {\tiny 0.019}} (\underline{0.902 $\pm$ {\tiny 0.001}})	& 0.381 $\pm$ {\tiny 0.017} (0.399 $\pm$ {\tiny 0.010})	& \underline{0.921 $\pm$ {\tiny 0.021}} (0.897 $\pm$ {\tiny 0.007})	& 0.386 $\pm$ {\tiny 0.018} (0.405 $\pm$ {\tiny 0.012})	& 0.543 $\pm$ {\tiny 0.015} (0.558 $\pm$ {\tiny 0.012}) \\
\rowcolor{gray}
\multirow{-8}{*}{\cellcolor{white}ADE V2 ($\alpha$=0.9)}	& DMO+ERO (ours)	& \underline{0.900 $\pm$ {\tiny 0.000}} (\underline{0.901 $\pm$ {\tiny 0.001}})	& \textcolor{red}{0.702 $\pm$ {\tiny 0.005}} (\textcolor{blue}{0.691 $\pm$ {\tiny 0.012}})	& 0.896 $\pm$ {\tiny 0.002} (\underline{0.901 $\pm$ {\tiny 0.001}})	& 0.709 $\pm$ {\tiny 0.007} (\textcolor{blue}{0.697 $\pm$ {\tiny 0.012}})	& 0.882 $\pm$ {\tiny 0.002} (0.887 $\pm$ {\tiny 0.003})	& 0.690 $\pm$ {\tiny 0.006} (0.676 $\pm$ {\tiny 0.011})	& \textcolor{red}{0.774 $\pm$ {\tiny 0.004}} (\textcolor{blue}{0.767 $\pm$ {\tiny 0.007}}) \\

\hline

	& WCE	& \underline{0.917 $\pm$ {\tiny 0.010}} (\underline{0.923 $\pm$ {\tiny 0.006}})	& 0.079 $\pm$ {\tiny 0.003} (0.078 $\pm$ {\tiny 0.002})	& 0.894 $\pm$ {\tiny 0.013} (\underline{0.903 $\pm$ {\tiny 0.002}})	& 0.086 $\pm$ {\tiny 0.003} (0.085 $\pm$ {\tiny 0.002})	& 0.899 $\pm$ {\tiny 0.013} (\underline{0.906 $\pm$ {\tiny 0.006}})	& 0.544 $\pm$ {\tiny 0.009} (0.540 $\pm$ {\tiny 0.006})	& 0.677 $\pm$ {\tiny 0.004} (0.677 $\pm$ {\tiny 0.004}) \\
	& Focal	& \underline{0.971 $\pm$ {\tiny 0.033}} (\underline{0.929 $\pm$ {\tiny 0.039}})	& 0.064 $\pm$ {\tiny 0.004} (0.066 $\pm$ {\tiny 0.004})	& \underline{0.963 $\pm$ {\tiny 0.042}} (\underline{0.934 $\pm$ {\tiny 0.037}})	& 0.072 $\pm$ {\tiny 0.003} (0.074 $\pm$ {\tiny 0.004})	& \underline{0.970 $\pm$ {\tiny 0.035}} (\underline{0.914 $\pm$ {\tiny 0.048}})	& 0.487 $\pm$ {\tiny 0.013} (0.491 $\pm$ {\tiny 0.015})	& 0.648 $\pm$ {\tiny 0.004} (0.638 $\pm$ {\tiny 0.015}) \\
	& LDAM	& 0.708 $\pm$ {\tiny 0.109} (\underline{0.916 $\pm$ {\tiny 0.012}})	& 0.101 $\pm$ {\tiny 0.018} (0.070 $\pm$ {\tiny 0.001})	& 0.681 $\pm$ {\tiny 0.109} (\underline{0.911 $\pm$ {\tiny 0.010}})	& 0.109 $\pm$ {\tiny 0.018} (0.078 $\pm$ {\tiny 0.002})	& 0.659 $\pm$ {\tiny 0.131} (\underline{0.908 $\pm$ {\tiny 0.014}})	& 0.582 $\pm$ {\tiny 0.040} (0.506 $\pm$ {\tiny 0.006})	& 0.608 $\pm$ {\tiny 0.041} (0.650 $\pm$ {\tiny 0.004}) \\
	& TERM	& \underline{0.961 $\pm$ {\tiny 0.028}} (\underline{0.916 $\pm$ {\tiny 0.012}})	& 0.066 $\pm$ {\tiny 0.003} (0.071 $\pm$ {\tiny 0.002})	& \underline{0.953 $\pm$ {\tiny 0.032}} (\underline{0.908 $\pm$ {\tiny 0.007}})	& \textcolor{red}{0.074 $\pm$ {\tiny 0.003}} (0.079 $\pm$ {\tiny 0.002})	& \underline{0.953 $\pm$ {\tiny 0.035}} (\underline{0.906 $\pm$ {\tiny 0.015}})	& 0.494 $\pm$ {\tiny 0.011} (0.509 $\pm$ {\tiny 0.006})	& 0.650 $\pm$ {\tiny 0.003} (0.652 $\pm$ {\tiny 0.004}) \\
	& ROS	& \underline{0.918 $\pm$ {\tiny 0.014}} (\underline{0.924 $\pm$ {\tiny 0.011}})	& 0.079 $\pm$ {\tiny 0.004} (0.077 $\pm$ {\tiny 0.002})	& 0.891 $\pm$ {\tiny 0.017} (\underline{0.904 $\pm$ {\tiny 0.003}})	& 0.085 $\pm$ {\tiny 0.004} (0.084 $\pm$ {\tiny 0.003})	& \underline{0.905 $\pm$ {\tiny 0.020}} (\underline{0.910 $\pm$ {\tiny 0.012}})	& \textcolor{red}{0.545 $\pm$ {\tiny 0.014}} (0.540 $\pm$ {\tiny 0.009})	& \textcolor{red}{0.680 $\pm$ {\tiny 0.006}} (0.678 $\pm$ {\tiny 0.006}) \\
	& RUS	& \underline{0.911 $\pm$ {\tiny 0.008}} (\underline{0.923 $\pm$ {\tiny 0.007}})	& 0.081 $\pm$ {\tiny 0.004} (0.078 $\pm$ {\tiny 0.002})	& 0.886 $\pm$ {\tiny 0.011} (\underline{0.902 $\pm$ {\tiny 0.001}})	& 0.087 $\pm$ {\tiny 0.003} (0.085 $\pm$ {\tiny 0.002})	& 0.894 $\pm$ {\tiny 0.011} (\underline{0.908 $\pm$ {\tiny 0.005}})	& 0.550 $\pm$ {\tiny 0.013} (\textcolor{blue}{0.541 $\pm$ {\tiny 0.007}})	& \textcolor{red}{0.681 $\pm$ {\tiny 0.008}} (0.678 $\pm$ {\tiny 0.006}) \\
	& DMO+TFCO	& 0.432 $\pm$ {\tiny 0.443} (\underline{0.908 $\pm$ {\tiny 0.011}})	& 0.155 $\pm$ {\tiny 0.100} (0.068 $\pm$ {\tiny 0.002})	& 0.431 $\pm$ {\tiny 0.439} (\underline{0.910 $\pm$ {\tiny 0.006}})	& 0.215 $\pm$ {\tiny 0.161} (0.076 $\pm$ {\tiny 0.003})	& 0.422 $\pm$ {\tiny 0.435} (\underline{0.899 $\pm$ {\tiny 0.016}})	& 0.680 $\pm$ {\tiny 0.193} (0.498 $\pm$ {\tiny 0.008})	& 0.319 $\pm$ {\tiny 0.298} (0.641 $\pm$ {\tiny 0.008}) \\
\rowcolor{gray}
\multirow{-8}{*}{\cellcolor{white}MURA ($\alpha$=0.9)}	& DMO+ERO (ours)	& \underline{0.900 $\pm$ {\tiny 0.000}} (\underline{0.937 $\pm$ {\tiny 0.017}})	& \textcolor{red}{0.137 $\pm$ {\tiny 0.041}} (\textcolor{blue}{0.085 $\pm$ {\tiny 0.010}})	& 0.771 $\pm$ {\tiny 0.053} (\underline{0.904 $\pm$ {\tiny 0.005}})	& 0.128 $\pm$ {\tiny 0.031} (\textcolor{blue}{0.093 $\pm$ {\tiny 0.010}})	& 0.747 $\pm$ {\tiny 0.063} (0.890 $\pm$ {\tiny 0.017})	& 0.630 $\pm$ {\tiny 0.077} (0.552 $\pm$ {\tiny 0.031})	& 0.676 $\pm$ {\tiny 0.024} (\textcolor{blue}{0.680 $\pm$ {\tiny 0.022}}) \\

\hline

	& WCE	& 0.910 $\pm$ {\tiny 0.010} (\underline{0.954 $\pm$ {\tiny 0.003}})	& 0.434 $\pm$ {\tiny 0.020} (0.392 $\pm$ {\tiny 0.015})	& 0.904 $\pm$ {\tiny 0.010} (\underline{0.952 $\pm$ {\tiny 0.001}})	& 0.426 $\pm$ {\tiny 0.021} (0.384 $\pm$ {\tiny 0.016})	& 0.896 $\pm$ {\tiny 0.010} (0.945 $\pm$ {\tiny 0.004})	& 0.429 $\pm$ {\tiny 0.021} (0.389 $\pm$ {\tiny 0.014})	& 0.580 $\pm$ {\tiny 0.018} (0.550 $\pm$ {\tiny 0.014}) \\
	& Focal	& 0.915 $\pm$ {\tiny 0.011} (\underline{0.954 $\pm$ {\tiny 0.004}})	& 0.515 $\pm$ {\tiny 0.016} (0.461 $\pm$ {\tiny 0.005})	& 0.914 $\pm$ {\tiny 0.008} (\underline{0.951 $\pm$ {\tiny 0.001}})	& 0.513 $\pm$ {\tiny 0.017} (0.459 $\pm$ {\tiny 0.006})	& 0.904 $\pm$ {\tiny 0.012} (0.949 $\pm$ {\tiny 0.005})	& 0.511 $\pm$ {\tiny 0.016} (0.458 $\pm$ {\tiny 0.006})	& 0.653 $\pm$ {\tiny 0.011} (0.618 $\pm$ {\tiny 0.005}) \\
	& LDAM	& 0.777 $\pm$ {\tiny 0.010} (0.949 $\pm$ {\tiny 0.006})	& 0.522 $\pm$ {\tiny 0.106} (0.401 $\pm$ {\tiny 0.063})	& 0.766 $\pm$ {\tiny 0.012} (\underline{0.952 $\pm$ {\tiny 0.002}})	& 0.507 $\pm$ {\tiny 0.108} (0.395 $\pm$ {\tiny 0.064})	& 0.762 $\pm$ {\tiny 0.010} (0.939 $\pm$ {\tiny 0.006})	& 0.522 $\pm$ {\tiny 0.100} (0.397 $\pm$ {\tiny 0.062})	& 0.615 $\pm$ {\tiny 0.061} (0.555 $\pm$ {\tiny 0.056}) \\
	& TERM	& \underline{0.955 $\pm$ {\tiny 0.026}} (\underline{0.950 $\pm$ {\tiny 0.005}})	& 0.358 $\pm$ {\tiny 0.036} (0.366 $\pm$ {\tiny 0.014})	& \underline{0.955 $\pm$ {\tiny 0.029}} (\underline{0.952 $\pm$ {\tiny 0.002}})	& \textcolor{red}{0.353 $\pm$ {\tiny 0.035}} (0.361 $\pm$ {\tiny 0.015})	& 0.947 $\pm$ {\tiny 0.030} (0.943 $\pm$ {\tiny 0.006})	& 0.356 $\pm$ {\tiny 0.036} (0.363 $\pm$ {\tiny 0.015})	& 0.515 $\pm$ {\tiny 0.035} (0.524 $\pm$ {\tiny 0.015}) \\
	& ROS	& 0.930 $\pm$ {\tiny 0.028} (\underline{0.952 $\pm$ {\tiny 0.006}})	& 0.398 $\pm$ {\tiny 0.033} (0.380 $\pm$ {\tiny 0.014})	& 0.926 $\pm$ {\tiny 0.030} (\underline{0.951 $\pm$ {\tiny 0.001}})	& 0.390 $\pm$ {\tiny 0.032} (0.372 $\pm$ {\tiny 0.015})	& 0.919 $\pm$ {\tiny 0.030} (0.942 $\pm$ {\tiny 0.008})	& 0.393 $\pm$ {\tiny 0.033} (0.377 $\pm$ {\tiny 0.013})	& 0.549 $\pm$ {\tiny 0.029} (0.538 $\pm$ {\tiny 0.013}) \\
	& RUS	& 0.935 $\pm$ {\tiny 0.032} (\underline{0.953 $\pm$ {\tiny 0.004}})	& 0.401 $\pm$ {\tiny 0.048} (0.385 $\pm$ {\tiny 0.015})	& 0.930 $\pm$ {\tiny 0.036} (\underline{0.952 $\pm$ {\tiny 0.002}})	& 0.393 $\pm$ {\tiny 0.046} (0.378 $\pm$ {\tiny 0.015})	& 0.924 $\pm$ {\tiny 0.036} (0.942 $\pm$ {\tiny 0.003})	& 0.397 $\pm$ {\tiny 0.047} (0.382 $\pm$ {\tiny 0.015})	& 0.553 $\pm$ {\tiny 0.042} (0.544 $\pm$ {\tiny 0.015}) \\
	& DMO+TFCO	& \underline{0.974 $\pm$ {\tiny 0.012}} (\underline{0.953 $\pm$ {\tiny 0.004}})	& 0.356 $\pm$ {\tiny 0.026} (0.378 $\pm$ {\tiny 0.013})	& \underline{0.974 $\pm$ {\tiny 0.015}} (\underline{0.951 $\pm$ {\tiny 0.001}})	& 0.349 $\pm$ {\tiny 0.024} (0.371 $\pm$ {\tiny 0.012})	& \underline{0.968 $\pm$ {\tiny 0.016}} (0.945 $\pm$ {\tiny 0.005})	& \textcolor{red}{0.353 $\pm$ {\tiny 0.026}} (0.374 $\pm$ {\tiny 0.013})	& 0.516 $\pm$ {\tiny 0.025} (0.536 $\pm$ {\tiny 0.014}) \\
\rowcolor{gray}
\multirow{-8}{*}{\cellcolor{white}ADE V2 ($\alpha$=0.95)}	& DMO+ERO (ours)	& \underline{0.950 $\pm$ {\tiny 0.000}} (\underline{0.957 $\pm$ {\tiny 0.001}})	& \textcolor{red}{0.675 $\pm$ {\tiny 0.006}} (\textcolor{blue}{0.586 $\pm$ {\tiny 0.009}})	& 0.938 $\pm$ {\tiny 0.003} (\underline{0.951 $\pm$ {\tiny 0.000}})	& 0.677 $\pm$ {\tiny 0.006} (\textcolor{blue}{0.587 $\pm$ {\tiny 0.011}})	& 0.938 $\pm$ {\tiny 0.002} (\underline{0.952 $\pm$ {\tiny 0.003}})	& 0.665 $\pm$ {\tiny 0.006} (\textcolor{blue}{0.585 $\pm$ {\tiny 0.009}})	& \textcolor{red}{0.779 $\pm$ {\tiny 0.004}} (\textcolor{blue}{0.725 $\pm$ {\tiny 0.007}}) \\

\hline

	& WCE	& 0.917 $\pm$ {\tiny 0.010} (\underline{0.970 $\pm$ {\tiny 0.004}})	& 0.079 $\pm$ {\tiny 0.003} (0.068 $\pm$ {\tiny 0.001})	& 0.894 $\pm$ {\tiny 0.013} (\underline{0.954 $\pm$ {\tiny 0.004}})	& 0.086 $\pm$ {\tiny 0.003} (0.075 $\pm$ {\tiny 0.001})	& 0.899 $\pm$ {\tiny 0.013} (\underline{0.968 $\pm$ {\tiny 0.006}})	& 0.544 $\pm$ {\tiny 0.009} (0.507 $\pm$ {\tiny 0.004})	& 0.677 $\pm$ {\tiny 0.004} (0.666 $\pm$ {\tiny 0.003}) \\
	& Focal	& \underline{0.971 $\pm$ {\tiny 0.033}} (\underline{0.965 $\pm$ {\tiny 0.018}})	& 0.064 $\pm$ {\tiny 0.004} (0.064 $\pm$ {\tiny 0.002})	& \underline{0.963 $\pm$ {\tiny 0.042}} (\underline{0.970 $\pm$ {\tiny 0.018}})	& 0.072 $\pm$ {\tiny 0.003} (0.072 $\pm$ {\tiny 0.002})	& \underline{0.970 $\pm$ {\tiny 0.035}} (\underline{0.958 $\pm$ {\tiny 0.025}})	& 0.487 $\pm$ {\tiny 0.013} (0.484 $\pm$ {\tiny 0.007})	& 0.648 $\pm$ {\tiny 0.004} (0.643 $\pm$ {\tiny 0.008}) \\
	& LDAM	& 0.708 $\pm$ {\tiny 0.109} (\underline{0.956 $\pm$ {\tiny 0.004}})	& 0.101 $\pm$ {\tiny 0.018} (0.066 $\pm$ {\tiny 0.001})	& 0.681 $\pm$ {\tiny 0.109} (\underline{0.954 $\pm$ {\tiny 0.003}})	& 0.109 $\pm$ {\tiny 0.018} (0.074 $\pm$ {\tiny 0.001})	& 0.659 $\pm$ {\tiny 0.131} (\underline{0.950 $\pm$ {\tiny 0.006}})	& 0.582 $\pm$ {\tiny 0.040} (0.493 $\pm$ {\tiny 0.003})	& 0.608 $\pm$ {\tiny 0.041} (0.649 $\pm$ {\tiny 0.002}) \\
	& TERM	& \underline{0.961 $\pm$ {\tiny 0.028}} (\underline{0.958 $\pm$ {\tiny 0.007}})	& 0.066 $\pm$ {\tiny 0.003} (0.067 $\pm$ {\tiny 0.001})	& \underline{0.953 $\pm$ {\tiny 0.032}} (\underline{0.954 $\pm$ {\tiny 0.005}})	& \textcolor{red}{0.074 $\pm$ {\tiny 0.003}} (0.075 $\pm$ {\tiny 0.001})	& \underline{0.953 $\pm$ {\tiny 0.035}} (\underline{0.950 $\pm$ {\tiny 0.010}})	& \textcolor{red}{0.494 $\pm$ {\tiny 0.011}} (0.494 $\pm$ {\tiny 0.004})	& 0.650 $\pm$ {\tiny 0.003} (0.650 $\pm$ {\tiny 0.003}) \\
	& ROS	& 0.918 $\pm$ {\tiny 0.014} (\underline{0.974 $\pm$ {\tiny 0.004}})	& 0.079 $\pm$ {\tiny 0.004} (0.068 $\pm$ {\tiny 0.001})	& 0.891 $\pm$ {\tiny 0.017} (\underline{0.956 $\pm$ {\tiny 0.007}})	& 0.085 $\pm$ {\tiny 0.004} (0.075 $\pm$ {\tiny 0.001})	& 0.905 $\pm$ {\tiny 0.020} (\underline{0.972 $\pm$ {\tiny 0.003}})	& 0.545 $\pm$ {\tiny 0.014} (0.505 $\pm$ {\tiny 0.003})	& 0.680 $\pm$ {\tiny 0.006} (0.665 $\pm$ {\tiny 0.002}) \\
	& RUS	& 0.911 $\pm$ {\tiny 0.008} (\underline{0.971 $\pm$ {\tiny 0.002}})	& 0.081 $\pm$ {\tiny 0.004} (0.069 $\pm$ {\tiny 0.002})	& 0.886 $\pm$ {\tiny 0.011} (\underline{0.954 $\pm$ {\tiny 0.004}})	& 0.087 $\pm$ {\tiny 0.003} (0.076 $\pm$ {\tiny 0.002})	& 0.894 $\pm$ {\tiny 0.011} (\underline{0.967 $\pm$ {\tiny 0.006}})	& 0.550 $\pm$ {\tiny 0.013} (\textcolor{blue}{0.509 $\pm$ {\tiny 0.007}})	& 0.681 $\pm$ {\tiny 0.008} (0.667 $\pm$ {\tiny 0.005}) \\
	& DMO+TFCO	& 0.280 $\pm$ {\tiny 0.351} (\underline{0.957 $\pm$ {\tiny 0.009}})	& 0.247 $\pm$ {\tiny 0.217} (0.065 $\pm$ {\tiny 0.001})	& 0.273 $\pm$ {\tiny 0.349} (\underline{0.955 $\pm$ {\tiny 0.009}})	& 0.156 $\pm$ {\tiny 0.211} (0.074 $\pm$ {\tiny 0.001})	& 0.263 $\pm$ {\tiny 0.349} (\underline{0.959 $\pm$ {\tiny 0.012}})	& 0.732 $\pm$ {\tiny 0.178} (0.492 $\pm$ {\tiny 0.006})	& 0.246 $\pm$ {\tiny 0.239} (0.650 $\pm$ {\tiny 0.005}) \\
\rowcolor{gray}
\multirow{-8}{*}{\cellcolor{white}MURA ($\alpha$=0.95)}	& DMO+ERO (ours)	& \underline{0.949 $\pm$ {\tiny 0.001}} (\underline{0.973 $\pm$ {\tiny 0.012}})	& \textcolor{red}{0.128 $\pm$ {\tiny 0.036}} (\textcolor{blue}{0.075 $\pm$ {\tiny 0.004}})	& 0.821 $\pm$ {\tiny 0.055} (\underline{0.951 $\pm$ {\tiny 0.001}})	& 0.122 $\pm$ {\tiny 0.027} (\textcolor{blue}{0.083 $\pm$ {\tiny 0.003}})	& 0.797 $\pm$ {\tiny 0.063} (0.941 $\pm$ {\tiny 0.012})	& 0.617 $\pm$ {\tiny 0.069} (0.527 $\pm$ {\tiny 0.016})	& \textcolor{red}{0.689 $\pm$ {\tiny 0.025}} (\textcolor{blue}{0.675 $\pm$ {\tiny 0.015}}) \\

\hline

	& WCE	& 0.910 $\pm$ {\tiny 0.010} (0.987 $\pm$ {\tiny 0.002})	& 0.434 $\pm$ {\tiny 0.020} (0.342 $\pm$ {\tiny 0.009})	& 0.904 $\pm$ {\tiny 0.010} (\underline{0.991 $\pm$ {\tiny 0.000}})	& 0.426 $\pm$ {\tiny 0.021} (0.337 $\pm$ {\tiny 0.008})	& 0.896 $\pm$ {\tiny 0.010} (0.980 $\pm$ {\tiny 0.003})	& 0.429 $\pm$ {\tiny 0.021} (0.338 $\pm$ {\tiny 0.009})	& 0.580 $\pm$ {\tiny 0.018} (0.503 $\pm$ {\tiny 0.010}) \\
	& Focal	& 0.915 $\pm$ {\tiny 0.011} (\underline{0.990 $\pm$ {\tiny 0.001}})	& 0.515 $\pm$ {\tiny 0.016} (0.376 $\pm$ {\tiny 0.007})	& 0.914 $\pm$ {\tiny 0.008} (\underline{0.991 $\pm$ {\tiny 0.000}})	& 0.513 $\pm$ {\tiny 0.017} (0.374 $\pm$ {\tiny 0.008})	& 0.904 $\pm$ {\tiny 0.012} (\underline{0.989 $\pm$ {\tiny 0.002}})	& 0.511 $\pm$ {\tiny 0.016} (\textcolor{blue}{0.374 $\pm$ {\tiny 0.007}})	& 0.653 $\pm$ {\tiny 0.011} (0.543 $\pm$ {\tiny 0.007}) \\
	& LDAM	& 0.777 $\pm$ {\tiny 0.010} (0.986 $\pm$ {\tiny 0.003})	& 0.522 $\pm$ {\tiny 0.106} (0.342 $\pm$ {\tiny 0.034})	& 0.766 $\pm$ {\tiny 0.012} (\underline{0.991 $\pm$ {\tiny 0.000}})	& 0.507 $\pm$ {\tiny 0.108} (0.337 $\pm$ {\tiny 0.034})	& 0.762 $\pm$ {\tiny 0.010} (0.980 $\pm$ {\tiny 0.005})	& 0.522 $\pm$ {\tiny 0.100} (0.339 $\pm$ {\tiny 0.034})	& 0.615 $\pm$ {\tiny 0.061} (0.502 $\pm$ {\tiny 0.036}) \\
	& TERM	& 0.955 $\pm$ {\tiny 0.026} (0.987 $\pm$ {\tiny 0.003})	& 0.358 $\pm$ {\tiny 0.036} (0.323 $\pm$ {\tiny 0.007})	& 0.955 $\pm$ {\tiny 0.029} (\underline{0.991 $\pm$ {\tiny 0.000}})	& 0.353 $\pm$ {\tiny 0.035} (0.320 $\pm$ {\tiny 0.007})	& 0.947 $\pm$ {\tiny 0.030} (0.983 $\pm$ {\tiny 0.006})	& 0.356 $\pm$ {\tiny 0.036} (0.321 $\pm$ {\tiny 0.007})	& 0.515 $\pm$ {\tiny 0.035} (0.484 $\pm$ {\tiny 0.007}) \\
	& ROS	& 0.930 $\pm$ {\tiny 0.028} (0.987 $\pm$ {\tiny 0.003})	& 0.398 $\pm$ {\tiny 0.033} (0.336 $\pm$ {\tiny 0.008})	& 0.926 $\pm$ {\tiny 0.030} (\underline{0.992 $\pm$ {\tiny 0.001}})	& 0.390 $\pm$ {\tiny 0.032} (0.331 $\pm$ {\tiny 0.007})	& 0.919 $\pm$ {\tiny 0.030} (0.981 $\pm$ {\tiny 0.004})	& 0.393 $\pm$ {\tiny 0.033} (0.333 $\pm$ {\tiny 0.007})	& 0.549 $\pm$ {\tiny 0.029} (0.497 $\pm$ {\tiny 0.007}) \\
	& RUS	& 0.935 $\pm$ {\tiny 0.032} (0.987 $\pm$ {\tiny 0.002})	& 0.401 $\pm$ {\tiny 0.048} (0.338 $\pm$ {\tiny 0.011})	& 0.930 $\pm$ {\tiny 0.036} (\underline{0.991 $\pm$ {\tiny 0.001}})	& 0.393 $\pm$ {\tiny 0.046} (0.333 $\pm$ {\tiny 0.011})	& 0.924 $\pm$ {\tiny 0.036} (0.980 $\pm$ {\tiny 0.003})	& 0.397 $\pm$ {\tiny 0.047} (0.334 $\pm$ {\tiny 0.012})	& 0.553 $\pm$ {\tiny 0.042} (0.498 $\pm$ {\tiny 0.013}) \\
	& DMO+TFCO	& \underline{0.993 $\pm$ {\tiny 0.003}} (\underline{0.989 $\pm$ {\tiny 0.001}})	& 0.325 $\pm$ {\tiny 0.008} (0.334 $\pm$ {\tiny 0.003})	& \underline{0.994 $\pm$ {\tiny 0.003}} (\underline{0.991 $\pm$ {\tiny 0.000}})	& \textcolor{red}{0.321 $\pm$ {\tiny 0.008}} (0.329 $\pm$ {\tiny 0.003})	& \underline{0.992 $\pm$ {\tiny 0.005}} (0.986 $\pm$ {\tiny 0.001})	& \textcolor{red}{0.323 $\pm$ {\tiny 0.008}} (0.331 $\pm$ {\tiny 0.003})	& 0.487 $\pm$ {\tiny 0.009} (0.496 $\pm$ {\tiny 0.003}) \\
\rowcolor{gray}
\multirow{-8}{*}{\cellcolor{white}ADE V2 ($\alpha$=0.99)}	& DMO+ERO (ours)	& \underline{0.990 $\pm$ {\tiny 0.000}} (\underline{0.991 $\pm$ {\tiny 0.000}})	& \textcolor{red}{0.628 $\pm$ {\tiny 0.004}} (\textcolor{blue}{0.465 $\pm$ {\tiny 0.011}})	& 0.976 $\pm$ {\tiny 0.002} (\underline{0.991 $\pm$ {\tiny 0.000}})	& 0.621 $\pm$ {\tiny 0.007} (\textcolor{blue}{0.465 $\pm$ {\tiny 0.010}})	& 0.980 $\pm$ {\tiny 0.001} (0.988 $\pm$ {\tiny 0.001})	& 0.617 $\pm$ {\tiny 0.004} (0.461 $\pm$ {\tiny 0.011})	& \textcolor{red}{0.757 $\pm$ {\tiny 0.003}} (\textcolor{blue}{0.629 $\pm$ {\tiny 0.010}}) \\

\hline

	& WCE	& 0.917 $\pm$ {\tiny 0.010} (\underline{0.996 $\pm$ {\tiny 0.001}})	& 0.079 $\pm$ {\tiny 0.003} (0.063 $\pm$ {\tiny 0.000})	& 0.894 $\pm$ {\tiny 0.013} (\underline{0.993 $\pm$ {\tiny 0.002}})	& 0.086 $\pm$ {\tiny 0.003} (0.071 $\pm$ {\tiny 0.000})	& 0.899 $\pm$ {\tiny 0.013} (\underline{0.996 $\pm$ {\tiny 0.002}})	& 0.544 $\pm$ {\tiny 0.009} (0.484 $\pm$ {\tiny 0.002})	& 0.677 $\pm$ {\tiny 0.004} (0.652 $\pm$ {\tiny 0.001}) \\
	& Focal	& 0.971 $\pm$ {\tiny 0.033} (\underline{0.993 $\pm$ {\tiny 0.006}})	& 0.064 $\pm$ {\tiny 0.004} (0.062 $\pm$ {\tiny 0.000})	& 0.963 $\pm$ {\tiny 0.042} (\underline{0.994 $\pm$ {\tiny 0.004}})	& 0.072 $\pm$ {\tiny 0.003} (0.070 $\pm$ {\tiny 0.000})	& 0.970 $\pm$ {\tiny 0.035} (\underline{0.992 $\pm$ {\tiny 0.006}})	& 0.487 $\pm$ {\tiny 0.013} (0.480 $\pm$ {\tiny 0.002})	& 0.648 $\pm$ {\tiny 0.004} (0.647 $\pm$ {\tiny 0.002}) \\
	& LDAM	& 0.708 $\pm$ {\tiny 0.109} (\underline{0.994 $\pm$ {\tiny 0.003}})	& 0.101 $\pm$ {\tiny 0.018} (0.063 $\pm$ {\tiny 0.000})	& 0.681 $\pm$ {\tiny 0.109} (\underline{0.993 $\pm$ {\tiny 0.002}})	& 0.109 $\pm$ {\tiny 0.018} (0.070 $\pm$ {\tiny 0.000})	& 0.659 $\pm$ {\tiny 0.131} (\underline{0.993 $\pm$ {\tiny 0.003}})	& 0.582 $\pm$ {\tiny 0.040} (0.481 $\pm$ {\tiny 0.001})	& 0.608 $\pm$ {\tiny 0.041} (0.648 $\pm$ {\tiny 0.001}) \\
	& TERM	& 0.961 $\pm$ {\tiny 0.028} (\underline{0.993 $\pm$ {\tiny 0.003}})	& 0.066 $\pm$ {\tiny 0.003} (0.063 $\pm$ {\tiny 0.000})	& 0.953 $\pm$ {\tiny 0.032} (\underline{0.992 $\pm$ {\tiny 0.002}})	& 0.074 $\pm$ {\tiny 0.003} (0.071 $\pm$ {\tiny 0.001})	& 0.953 $\pm$ {\tiny 0.035} (\underline{0.991 $\pm$ {\tiny 0.006}})	& 0.494 $\pm$ {\tiny 0.011} (0.482 $\pm$ {\tiny 0.001})	& 0.650 $\pm$ {\tiny 0.003} (0.648 $\pm$ {\tiny 0.002}) \\
	& ROS	& 0.918 $\pm$ {\tiny 0.014} (\underline{0.996 $\pm$ {\tiny 0.001}})	& 0.079 $\pm$ {\tiny 0.004} (0.063 $\pm$ {\tiny 0.000})	& 0.891 $\pm$ {\tiny 0.017} (\underline{0.992 $\pm$ {\tiny 0.001}})	& 0.085 $\pm$ {\tiny 0.004} (0.071 $\pm$ {\tiny 0.000})	& 0.905 $\pm$ {\tiny 0.020} (\underline{0.996 $\pm$ {\tiny 0.002}})	& 0.545 $\pm$ {\tiny 0.014} (0.485 $\pm$ {\tiny 0.002})	& 0.680 $\pm$ {\tiny 0.006} (0.653 $\pm$ {\tiny 0.002}) \\
	& RUS	& 0.911 $\pm$ {\tiny 0.008} (\underline{0.995 $\pm$ {\tiny 0.002}})	& 0.081 $\pm$ {\tiny 0.004} (\textcolor{blue}{0.063 $\pm$ {\tiny 0.001}})	& 0.886 $\pm$ {\tiny 0.011} (\underline{0.991 $\pm$ {\tiny 0.001}})	& 0.087 $\pm$ {\tiny 0.003} (\textcolor{blue}{0.071 $\pm$ {\tiny 0.001}})	& 0.894 $\pm$ {\tiny 0.011} (\underline{0.995 $\pm$ {\tiny 0.003}})	& 0.550 $\pm$ {\tiny 0.013} (0.487 $\pm$ {\tiny 0.004})	& 0.681 $\pm$ {\tiny 0.008} (0.654 $\pm$ {\tiny 0.003}) \\
	& DMO+TFCO	& 0.717 $\pm$ {\tiny 0.425} (\underline{0.990 $\pm$ {\tiny 0.005}})	& 0.177 $\pm$ {\tiny 0.278} (0.063 $\pm$ {\tiny 0.000})	& 0.713 $\pm$ {\tiny 0.425} (\underline{0.991 $\pm$ {\tiny 0.001}})	& 0.082 $\pm$ {\tiny 0.050} (0.071 $\pm$ {\tiny 0.000})	& 0.709 $\pm$ {\tiny 0.429} (0.987 $\pm$ {\tiny 0.004})	& 0.584 $\pm$ {\tiny 0.180} (0.481 $\pm$ {\tiny 0.002})	& 0.482 $\pm$ {\tiny 0.255} (0.647 $\pm$ {\tiny 0.002}) \\
\rowcolor{gray}
\multirow{-8}{*}{\cellcolor{white}MURA ($\alpha$=0.99)}	& DMO+ERO (ours)	& 0.989 $\pm$ {\tiny 0.001} (\underline{0.997 $\pm$ {\tiny 0.002}})	& 0.110 $\pm$ {\tiny 0.029} (\textcolor{blue}{0.064 $\pm$ {\tiny 0.001}})	& 0.879 $\pm$ {\tiny 0.054} (\underline{0.993 $\pm$ {\tiny 0.003}})	& 0.108 $\pm$ {\tiny 0.022} (\textcolor{blue}{0.072 $\pm$ {\tiny 0.001}})	& 0.855 $\pm$ {\tiny 0.064} (\underline{0.992 $\pm$ {\tiny 0.004}})	& 0.586 $\pm$ {\tiny 0.058} (\textcolor{blue}{0.489 $\pm$ {\tiny 0.006}})	& \textcolor{red}{0.690 $\pm$ {\tiny 0.023}} (\textcolor{blue}{0.655 $\pm$ {\tiny 0.005}}) \\

\hline
\Xhline{3\arrayrulewidth}
\end{tabular}
}
\label{table:three_split_FROP}
\end{table*}]

\end{document}